\documentclass{article}

\usepackage{microtype}
\usepackage{graphicx}
\usepackage{booktabs} 

\usepackage[hyphens]{url}            
\usepackage{amsmath}
\usepackage{amsfonts}       
\usepackage{nicefrac}       
\usepackage{subcaption}
\usepackage[group-separator={,}]{siunitx}
\usepackage{placeins}

\usepackage{hyperref}

\usepackage[accepted]{icml2019}

\icmltitlerunning{Metropolis-Hastings Generative Adversarial Networks}

\renewcommand{\vec}[1]{{\boldsymbol{\mathbf{#1}}}} 

\newcommand{\R}{\mathbb{R}}

\newcommand{\set}[1]{\mathcal{#1}}

\newcommand{\sample}{\sim}
\newcommand{\given}{|}

\newcommand{\norm}{\mathcal{N}}
\newcommand{\bern}{\textrm{Bern}}

\newcommand{\target}{{p^\star}}
\newcommand{\prop}{q}
\newcommand{\pinit}{{p_0}}

\newcommand{\PG}{{p_G}}
\newcommand{\PD}{{p_D}}
\newcommand{\PR}{{p_{\textrm{data}}}}
\newcommand{\accept}{\alpha}

\newcommand{\setx}{\set{X}}

\newcommand{\exfactor}{1.0}
\newcommand{\pganw}{0.95in}

\newcommand{\chainK}{K}

\begin{document}

\twocolumn[
\icmltitle{Metropolis-Hastings Generative Adversarial Networks}



\icmlsetsymbol{equal}{*}

\begin{icmlauthorlist}
\icmlauthor{Ryan Turner}{uber}
\icmlauthor{Jane Hung}{uber}
\icmlauthor{Eric Frank}{uber}
\icmlauthor{Yunus Saatci}{uber}
\icmlauthor{Jason Yosinski}{uber}
\end{icmlauthorlist}

\icmlaffiliation{uber}{Uber AI Labs}

\icmlcorrespondingauthor{Ryan Turner}{ryan.turner@uber.com}

\icmlkeywords{Metropolis Hastings, generative adversarial networks, Markov chain Monte Carlo}

\vskip 0.3in
]



\printAffiliationsAndNotice{}  

\begin{abstract}
We introduce the Metropolis-Hastings generative adversarial network (MH-GAN), which combines aspects of Markov chain Monte Carlo and GANs.
The MH-GAN draws samples from the distribution implicitly defined by a GAN's discriminator-generator pair, as opposed to standard GANs which draw samples from the distribution defined only by the generator.
It uses the discriminator from GAN training to build a wrapper around the generator for improved sampling.
With a perfect discriminator, this wrapped generator samples from the true distribution on the data exactly even when the generator is imperfect.
We demonstrate the benefits of the improved generator on multiple benchmark datasets, including CIFAR-10 and CelebA, using the DCGAN, WGAN, and progressive GAN\@.
\end{abstract}

\section{Introduction}

Traditionally, density estimation is done with a model that can compute the data likelihood.
Generative adversarial networks (GANs)~\citep{Goodfellow2014} present a radically new way to do density estimation:
They implicitly represent the density of the data via a classifier that distinguishes real from generated data.

\begin{figure*}[tbhp]
    \centering
    \begin{subfigure}[t]{2.85in}
       \centering
       \includegraphics[scale=1.075]{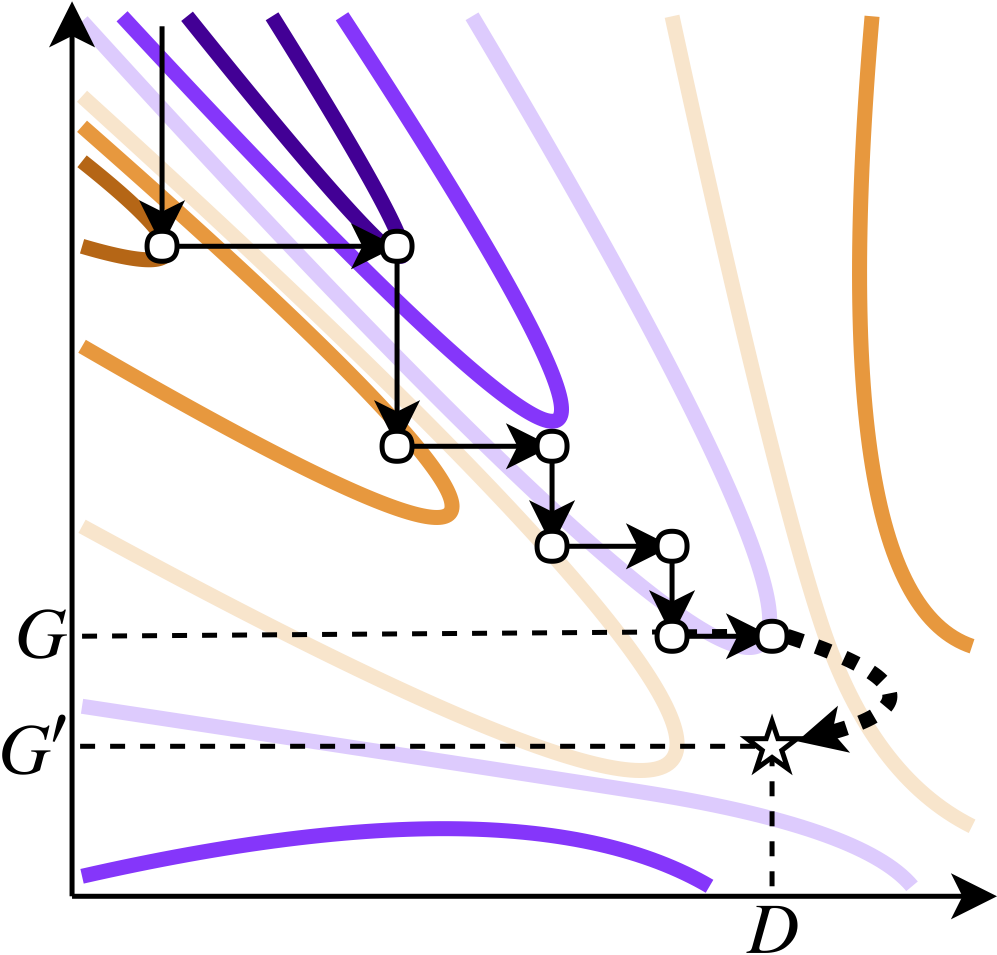}
       \caption{GAN value function}
    \end{subfigure}
    \hfill
    \begin{subfigure}[t]{3.8in}
       \centering
       \includegraphics[scale=1.075]{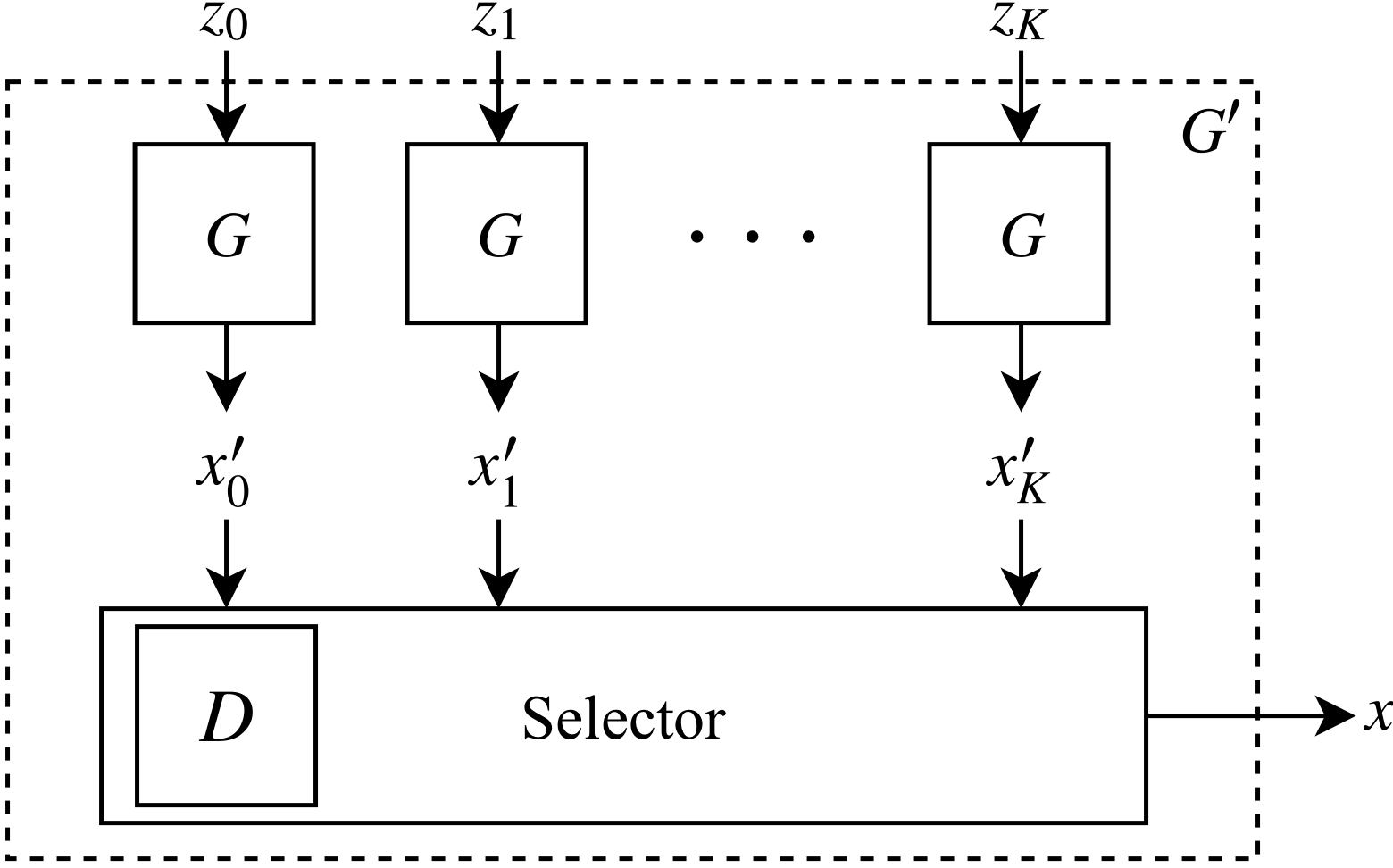}
       \caption{$G'$ wraps $G$}
    \end{subfigure}
    \caption{{\small
    \textbf{(a)} We diagram how training of $D$ and $G$ in GANs performs coordinate descent on the joint minimax value function, shown in the solid black arrow.
    If GAN training produces a perfect $D$ for an imperfect $G$, the MH-GAN wraps $G$ to produce a perfect generator $G'$, as shown in the final dashed arrow.
    The generator $G$ moves vertically towards the orange region while the discriminator $D$ moves horizontally towards the purple.
    \textbf{(b)} We illustrate how the MH-GAN is essentially a selector from multiple draws of $G$.
    In the MH-GAN, the selector is built using a Metropolis-Hastings (MH) acceptance rule from the discriminator scores $D$.
    }}
    \label{fig:intro}
\end{figure*}

GANs iterate between updating a discriminator $D$ and a generator $G$, where $G$ generates new (synthetic) samples of data, and $D$ attempts to distinguish samples of $G$ from the real data.
In the typical setup, $D$ is thrown away at the end of training, and only $G$ is kept for generating new synthetic data points.
In this work, we propose the Metropolis-Hastings GAN (MH-GAN), a GAN that constructs a new generator $G'$ that ``wraps'' $G$ using the information contained in $D$.
This principle is illustrated in Figure~\ref{fig:intro}.\footnote{Code found at:\\ \scriptsize{\url{github.com/uber-research/metropolis-hastings-gans}}}

The MH-GAN uses Markov chain Monte Carlo (MCMC) methods to sample from the distribution implicitly defined by the discriminator $D$ learned for the generator $G$.
This is built upon the notion that the discriminator classifies between the generator $G$ and a data distribution:
\begin{align}
  D(\vec x) = \frac{\PD(\vec x)}{\PD(\vec x) + \PG(\vec x)} \,, \label{eq:define PD}
\end{align}
where $\PG$ is the (intractable) density of samples from the generator $G$, and $\PD$ is the data density \emph{implied} by the discriminator $D$ with respect to $G$.
If GAN training reaches its global optimum, then this discriminator distribution $\PD$ is equal to the data distribution and the generator distribution ($\PD = \PR = \PG$)~\citep{Goodfellow2014}.
Furthermore, if the discriminator $D$ is optimal for a fixed imperfect generator, $G$ then the implied distribution still equals the data distribution ($\PD = \PR \neq \PG$)\@.

We use an MCMC \emph{independence sampler}~\citep{Tierney1994} to sample from $\PD$ by taking multiple samples from $G$\@.
Amazingly, using our algorithm, one can show that given a perfect discriminator $D$ and a decent (but imperfect) generator $G$, one can obtain exact samples from the true data distribution $\PR$.
Standard MCMC implementations require (unnormalized) densities for the target $\PD$ and the proposal $\PG$, which are both unavailable for GANs.
However, the Metropolis-Hastings (MH) algorithm requires only the ratio:
\begin{align}
  \frac{\PD(\vec x)}{\PG(\vec x)} = \frac{D(\vec x)}{1 - D(\vec x)}\,, \label{eq:PD inv}
\end{align}
which we can obtain using only evaluation of $D(\vec x)$.

Sampling from an MH-GAN is more computationally expensive than a standard GAN, but the bigger and more relevant training compute cost remains unchanged.
Thus, the MH-GAN is best suited for applications where sample quality is more important than compute speed at test time.

\nocite{Sugiyama2012}

The outline of this paper is as follows:
Section~\ref{sec:Related Work} reviews diverse areas of relevant prior work.
In Sections~\ref{sec:MCMC Methods} and~\ref{sec:GANs} we explain the necessary background on MCMC methods and GANs.
We explain our methodology of combining these two seemingly disparate areas in Section~\ref{sec:Methods} where we derive the wrapped generator $G'$.
Results on real data (CIFAR-10 and CelebA) and extending common GAN models (DCGAN, WGAN, and progressive GAN) are shown in Section~\ref{sec:Results}.
Section~\ref{sec:conclusions} discusses implications and conclusions.

\section{Related Work}
\label{sec:Related Work}

A few other works combine GANs and MCMC in some way.
\citet{Song2017} use a GAN-like procedure to train a RealNVP~\citep{Dinh2016} MCMC proposal for sampling an externally provided target \smash{$\target$}.
Whereas \citet{Song2017} use GANs to accelerate MCMC, we use MCMC to enhance the samples from a GAN\@.
Similar to \citet{Song2017}, \citet{Kempinska2017} improve proposals in particle filters rather than MCMC\@.
\citet{Song2017} was recently generalized by~\citet{Neklyudov2018}.

\subsection{Discriminator Rejection Sampling}
A concurrent work with similar aims from~\citet{Azadi2018} proposes discriminator rejection sampling (DRS) for GANs, which performs rejection sampling on the outputs of $G$ by using the probabilities given by $D$.
While conceptually appealing at first, DRS suffers from two major shortcomings in practice.
First, it is necessary to find an upper-bound on $D$ over all possible samples in order to obtain a valid proposal distribution for rejection sampling.
Because this is not possible, one must instead rely on estimating this bound by drawing many pilot samples.
Secondly, even if one were to find a good bound, the acceptance rate would become very low due to the high-dimensionality of the sampling space.
This leads~\citet{Azadi2018} to use an extra $\gamma$ heuristic to shift the logit $D$ scores, making the model sample from a distribution different from $\PR$ even when $D$ is perfect.
We use MCMC instead, which was invented precisely as a replacement for rejection sampling in higher dimensions.
We further improve the robustness of MCMC via use of a \emph{calibrator} on the discriminator to get more accurate probabilities for computing acceptance.

\section{Background and Notation}
\label{sec:Background}

In this section, we briefly review the notation and equations with MCMC and GANs.

\subsection{MCMC Methods}
\label{sec:MCMC Methods}

MCMC methods attempt to draw a chain of samples $\vec x_{1:\chainK} \in \setx^\chainK$ that marginally come from a target distribution $\target$.
We refer to the initial distribution as $\pinit$ and the proposal for the independence sampler as $\vec x' \sample \prop(\vec x' \given \vec x_k)=\prop(\vec x')$.
The proposal $\vec x' \in \setx$ is accepted with probability
\begin{align}
  \accept(\vec x', \vec x_k) = \min\left(1, \frac{\target(\vec x')\prop(\vec x_k)}{\target(\vec x_k)\prop(\vec x')}\right) \in [0,1]\,. \label{eq:alpha def}
\end{align}
If $\vec x'$ is accepted, $\vec x_{k+1} = \vec x'$, otherwise $\vec x_{k+1} = \vec x_k$.
Note that when estimating the distribution $\target$, one must include the duplicates that are a result of rejections in $\vec x'$.

\paragraph{Independent samples}
Many evaluation metrics assume perfectly iid samples.
Although MCMC methods are typically used to produce correlated samples, we can produce iid samples by using one chain per sample:
Each chain samples $\vec x_0 \sample \pinit$ and then does $\chainK$ MH iterations to get $\vec x_\chainK$ as the output of the chain, which is the output of $G'$.
Using multiple chains is also better for GPU parallelization.

\paragraph{Detailed balance}
The detailed balance condition implies that if $\vec x_k \sample \target$ exactly then $\vec x_{k+1} \sample \target$ exactly as well.
Even if $\vec x_k$ is not exactly distributed according to $\target$, the Kullback-Leibler (KL) divergence between the implied density it is drawn from and $\target$ always decreases as $k$ increases~\citep{Murray2008}.
We use detailed balance to motivate our approach to MH-GAN initialization.

\subsection{GANs}
\label{sec:GANs}

GANs implicitly model the data $\vec x$ via a synthetic data generator \smash{$G \in \R^{d} \rightarrow \setx$}:
\begin{align}
  \vec x = G(\vec z)\,, \quad \vec z \sample \norm(\vec 0, \vec I_{d})\,.
\end{align}
This implies a (intractable) distribution on the data $\vec x \sample \PG$.
We refer to the unknown true distribution on the data $\vec x$ as $\PR$.
The discriminator $D \in \setx \rightarrow [0,1]$ is a soft classifier predicting if a data point is real as opposed to being sampled from $\PG$\@.

If $D$ converges optimally for a fixed $G$, then \smash{$D = \PR/(\PR + \PG)$}, and if both $D$ and $G$ converge then $\PG = \PR$~\citep{Goodfellow2014}.
GAN training forms a game between $D$ and $G$.
In practice $D$ is often better at estimating the density ratio than G is at generating high-fidelity samples~\citep{Shibuya2017}.
This motivates wrapping an imperfect $G$ to obtain an improved $G'$ by using the density ratio information contained in $D$.

\begin{figure*}[htbp]
    \centering
    \includegraphics[width=1.0\linewidth]{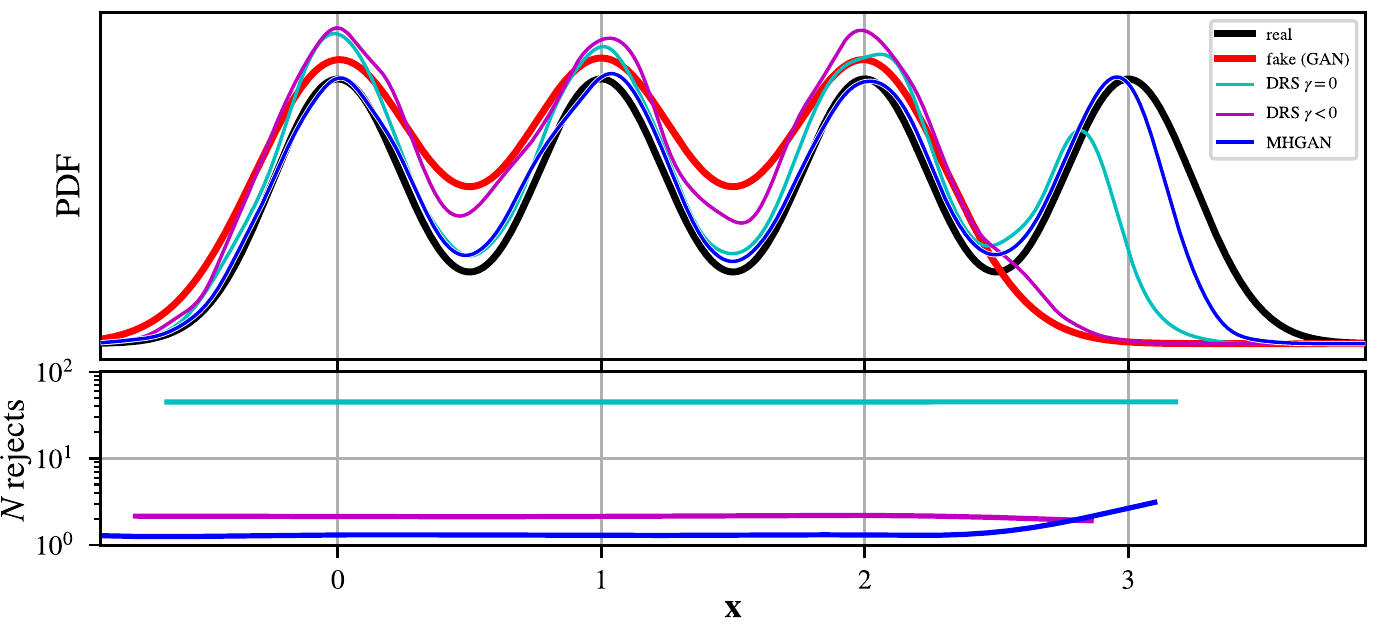}
    \caption{{\small
    Illustration comparing the MH-GAN setup with the formulation of DRS on a univariate example.
    This figure uses a $\PR$ of four Gaussian mixtures while $\PG$ is missing one of the mixtures.
    The top row shows the resulting density of samples, while the bottom row shows the typical number of rejects before accepting a sample at that $\vec x$ value.
    The MH-GAN recovers the true density except in the far right tail where there is an exponentially small chance of getting a sample from the proposal $\PG$.
    DRS with $\gamma=0$ shift should also be able to recover the density exactly, but it has an even larger error in the right tail.
    These errors arise because DRS must approximate the max $D$ score and use only \num{10000} pilot samples to do so, as in \citet{Azadi2018}.
    Additionally, due to the large maximum $D$, it needs a large number of draws before a single accept.
    DRS with $\gamma$ shift is much more sample efficient, but completely misses the right mode as the setup invalidates the rejection sampling equations.
    The MH-GAN is more adaptive in that it quickly accepts samples for the areas $\PG$ models well; more MCMC rejections occur before accepting a sample in the right poorly modeled mode.
    In all cases the MH-GAN is more efficient than DRS without $\gamma$ shift.
    Presumably, this effect becomes greater in high dimensions.
    }}
    \label{fig:univariate_example}
\end{figure*}

\section{Methods}
\label{sec:Methods}

In this section we show how to sample from the distribution $\PD$ implied by the discriminator $D$.
We apply~\eqref{eq:PD inv} and~\eqref{eq:alpha def} for a target of $\target=\PD$ and proposal $\prop=\PG$:
\begin{align}
  \frac{\PD}{\PG} &= \frac{1}{D^{-1} - 1} \\
  \implies
  \accept(\vec x', \vec x_k) &= \min\left(1, \frac{D(\vec x_k)^{-1} - 1}{D(\vec x')^{-1} - 1}\right)\,. \label{eq:alpha from D}
\end{align}
The ratio $\PD/\PG$ is computed entirely from the discriminator scores $D$.
If $D$ is perfect, $\PD = \PR$, so the sampler will marginally sample from $\PR$.
The use of~\eqref{eq:alpha from D} is further illustrated in Algorithm~\ref{alg:mhgan}.

A toy one-dimensional example with just such a perfect discriminator is shown in Figure~\ref{fig:univariate_example}.
In this example the MH-GAN is able to correctly reconstruct a missing mode in the generating distribution from the tail of a faulty generator.

\paragraph{Calibration}
The probabilities for $D$ must not merely provide a good AUC score, but must also be well \emph{calibrated}.
In other words, if one were to warp the probabilities of the perfect discriminator in~\eqref{eq:define PD} it may still suffice for standard GAN training, but it will not work in the MCMC procedure defined in~\eqref{eq:alpha from D}, as it will result in erroneous density ratios.

We can demonstrate the miscalibration of $D$ using the statistic of~\citet{Dawid1997} on held out samples \smash{$\vec x_{1:N}$} and real/fake labels \smash{$y_{1:N} \in \{0,1\}^N$}.
If $D$ is well calibrated, i.e., $y$ is indistinguishable from a \smash{$y \sample \bern(D(\vec x))$}, then
\begin{align}
  Z = \frac{\sum_{i=1}^N y_i - D(\vec x_i)}{\sqrt{\sum_{i=1}^N D(\vec x_i) (1 - D(\vec x_i))}} \!\!\implies\!\! Z \sample \norm(0,1)\,. \label{eq:calib score}
\end{align}
That is, we expect the $Z$ diagnostic to be a Gaussian in large $N$ for any well-calibrated classifier.
This means that for large values of $Z$, such as when $|Z| > 2$, we reject the hypothesis that $D$ is well-calibrated.

\paragraph{Correcting Calibration}
While~\eqref{eq:calib score} may tell us a classifier is poorly calibrated, we also need to be able to fix it.
Furthermore, some GANs (like WGAN) require calibration because their discriminator only outputs a score and not a probability.
To correct an uncalibrated classifier, denoted $\tilde{D} \in \setx \rightarrow \R$, we use a \emph{held out} calibration set (e.g., 10\% of the training data) and either logistic, isotonic, or beta~\citep{Kull2017} regression to warp the output of $\tilde{D}$.
The held out calibration set contains an equal number of positive and negative examples, which in the case of GANs is an even mix of real samples and fake samples from $G$.
After $\tilde{D}$ is learned, we train a probabilistic classifier $C \in \R \rightarrow [0,1]$ to map $\tilde{D}(\vec x_i)$ to $y_i$ using the calibration set.
The calibrated classifier is built via $D(\vec x_i) = C(\tilde{D}(\vec x_i))$.

\paragraph{Initialization}
We also avoid the burn-in issues that usually plague MCMC methods.
Recall that via the detailed balance property~\citep[Ch.~1]{Gilks1996}, if the marginal distribution of a Markov chain state $\vec x \in \setx$ at time step $k$ matches the target $\PD$ (\smash{$\vec x_k \sample \PD$}), then the marginal at time step $k+1$ will also follow $\PD$ (\smash{$\vec x_{k+1} \sample \PD$})\@.
In most MCMC applications it is not possible to get an initial sample from the target distribution (\smash{$\vec x_0 \sample \PD$})\@.

However, for MH-GAN, we have access to real data from the target distribution.
By initializing the chain at a sample of real data (the correct distribution), we apply the detailed balance property and avoid burn-in.
If no generated sample is accepted by the end of the chain, we restart sampling from a synthetic sample to ensure the initial real sample is never output.
To make restarts rare, we set $\chainK$ large (often 640)\@.

Using a restart after an MCMC chain of only rejects has a theoretical potential for bias.
However, MCMC in practice often uses chain diagnostics as a stopping criterion, which suffers the same bias potential~\citep{Cowles1999}.
Alternatively, we could never restart and always report the state after $\chainK$ samples, which will occasionally include the initial real sample.
This might be a better approach in certain statistical problems, where we care more about eliminating any potential source of bias, than in image generation.

\paragraph{Perfect Discriminator}
The assumption of a perfect $D$ may be weakened for two reasons:
(A)~Because we recalibrate the discriminator, the actual probabilities can be incorrect as long as the decision boundary between real and fake is correct.
(B)~Because the discriminator is only ever evaluated at samples from $G$ or the initial real sample $\vec x_0$, $D$ only needs to be accurate on the manifold of samples from the generator $\PG$ and the real data $\PR$.

\begin{figure*}[tbp]
    \centering
    \includegraphics[width=1.0\linewidth]{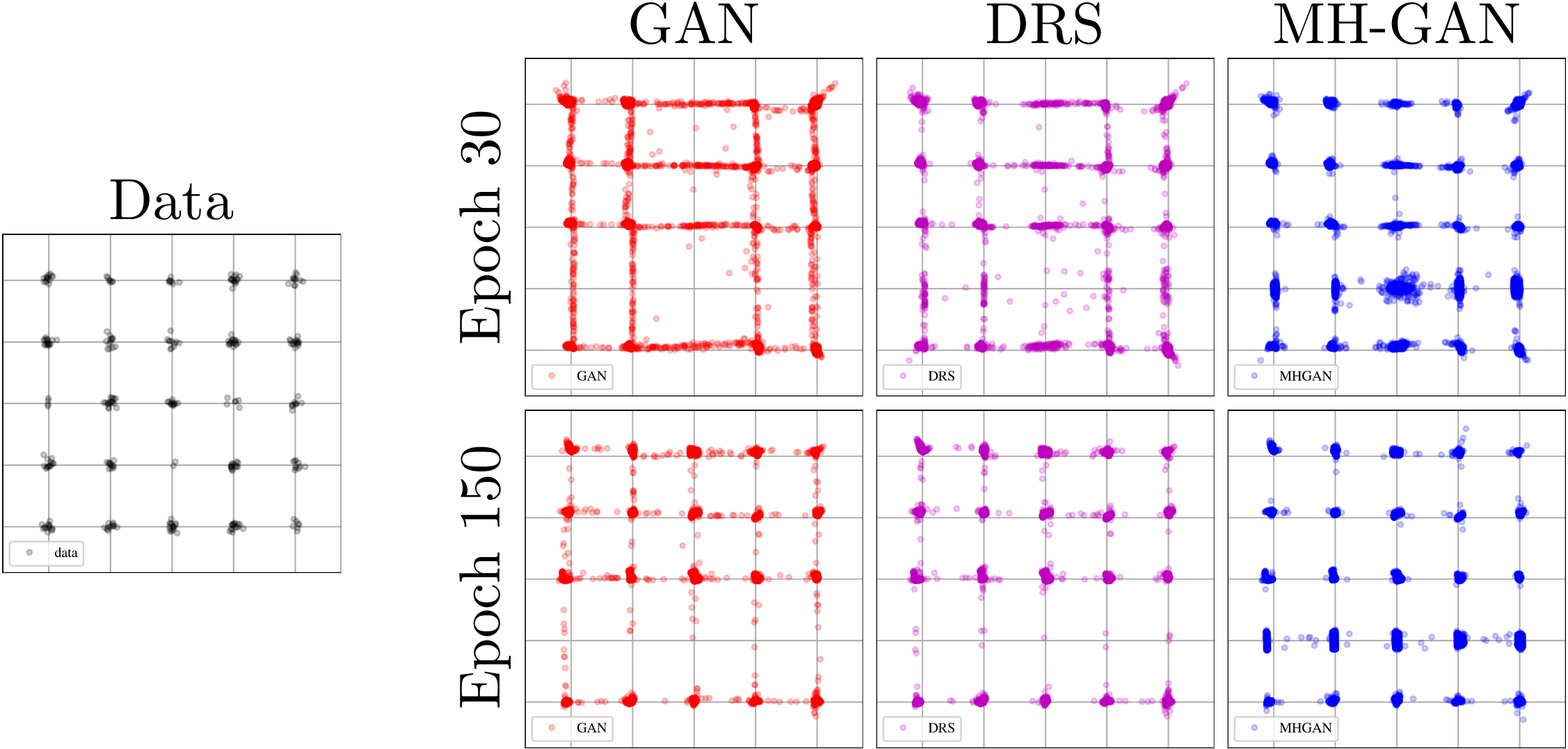}
    \caption{{\small
    The 25 Gaussians example.
    We show the state of the generators at epoch 30 (when MH-GAN begins showing large gains) on the top row and epoch 150 (the final epoch) on the bottom row.
    The MH-GAN corrects areas of mis-assigned mass in the original GAN\@.
    DRS appears visually closer to the original GAN than the data, whereas the MH-GAN appears closer to the actual data.
    }}
    \label{fig:mog_example}
\end{figure*}

\begin{algorithm}[tb]
   \caption{MH-GAN}
   \label{alg:mhgan}
\begin{algorithmic}
   \STATE {\bfseries Input:} generator $G$, calibrated disc.~$D$, real samples
   \STATE Assign random real sample $\vec x_0$ to $\vec x$
   \FOR{$k=1$ {\bfseries to} $\chainK$}
   \STATE Draw $\vec x'$ from $G$
   \STATE Draw $U$ from $\textrm{Uniform}(0,1)$
   \IF{\smash{$U \leq (D(\vec x)^{-1} - 1) / (D(\vec x')^{-1} - 1)$}}
   \STATE $\vec x \leftarrow \vec x'$
   \ENDIF
   \ENDFOR
   \STATE If $\vec x$ is still real sample $\vec x_0$ restart with draw from $G$ as $\vec x_0$
   \STATE {\bfseries Output:} sample $\vec x$ from $G'$
\end{algorithmic}
\end{algorithm}

\section{Results}
\label{sec:Results}

We first show an illustrative synthetic mixture model example followed by real data with images.

\subsection{Mixture of 25 Gaussians}

We consider the $5 \times 5$ grid of two-dimensional Gaussians used in~\citet{Azadi2018}, which has become a popular toy example in the GAN literature~\citep{Dumoulin2016}.
The means are arranged on the grid $\mu \in \{{-2},{-1},0,1,2\}$ and use a standard deviation of $\sigma = 0.05$.

\paragraph{Experimental setup}
Following~\citet{Azadi2018}, we use four fully connected layers with ReLU activations for both the generator and discriminator.
The final output layer of the discriminator is a sigmoid, and no nonlinearity is applied to the final generator layer.
All hidden layers have size 100, with a latent \smash{$\vec z \in \R^2$}.
We used \num{64000} standardized training points and generated \num{10000} points in test.

\paragraph{Visual results}
In Figure~\ref{fig:mog_example}, we show the original data along with samples generated by the GAN\@.
We also show samples enhanced via the MH-GAN (with calibration) and with DRS\@.
The standard GAN creates spurious links along the grid lines between modes and misses some modes along the bottom row.
DRS is able to reduce some of the spurious links but not fill in the missing modes.
The MH-GAN further reduces the spurious links and recovers these under-estimated modes.

\begin{figure*}[htbp]
    \centering
    \begin{subfigure}[b]{0.32\textwidth}
       \centering
       \includegraphics[width=2.2in]{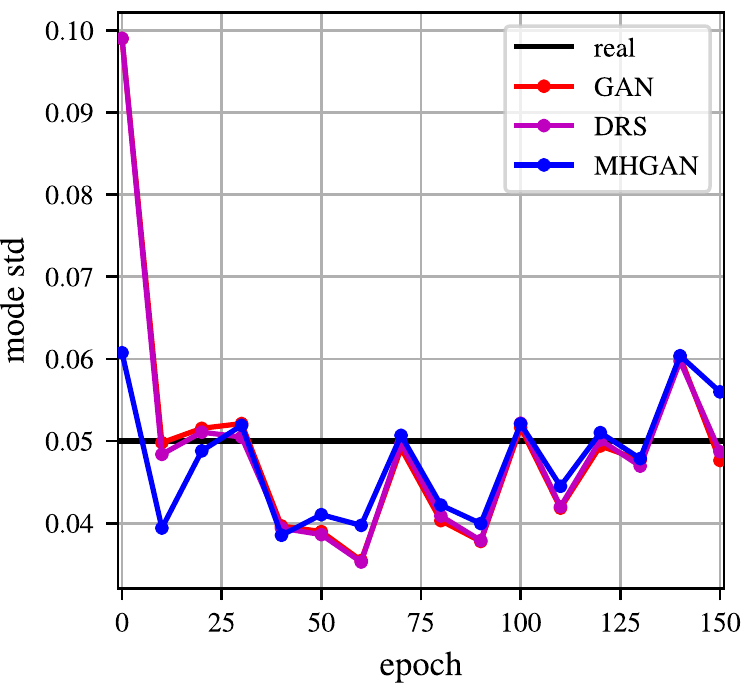}
       \caption{mode std.~dev.}
       \label{fig:std}
    \end{subfigure}
    \begin{subfigure}[b]{0.32\textwidth}
       \centering
       \includegraphics[width=2.2in]{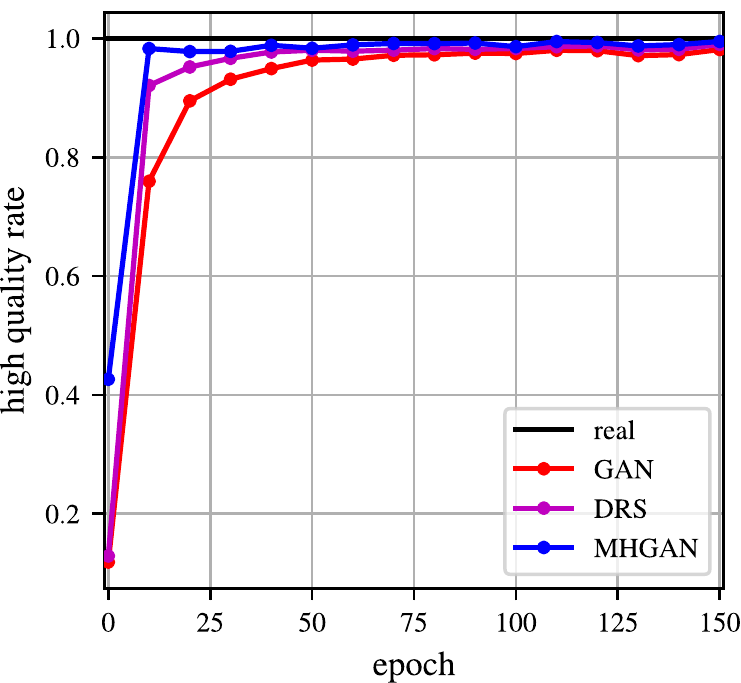}
       \caption{high quality rate}
       \label{fig:hqr}
    \end{subfigure}
    \begin{subfigure}[b]{0.32\textwidth}
       \centering
       \includegraphics[width=2.2in]{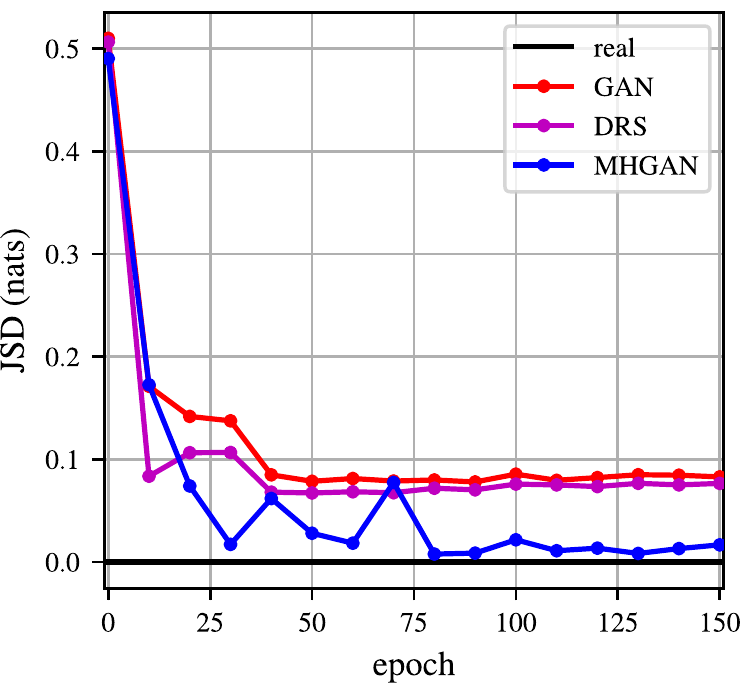}
       \caption{Jensen-Shannon divergence}
       \label{fig:jsd}
    \end{subfigure}
    \caption{{\small
    Results of the MH-GAN experiments on the mixture of 25 Gaussians example.
     \textbf{(a)} On the left, we show the standard deviation of samples within a single mode.
    The black lines represent values for the true distribution.
     \textbf{(b)} In the center, we show the high quality rate (samples near a real mode) across different GAN setups.
     \textbf{(c)} On the right, we show the Jensen-Shannon divergence (JSD) between the distribution on the nearest mode vs a uniform, which is the generating distribution on mixture components.
    The MH-GAN shows, on average, a $5 \times$ improvement in JSD over DRS\@.
    We considered adding error bars to these plots via a bootstrap analysis, but the error bars are too small to be visible.
    }}
    \label{fig:mog_metrics}
\end{figure*}

\paragraph{Quantitative results}
These results are made more quantitative in Figure~\ref{fig:mog_metrics}, where we follow some of the metrics for the example from~\citet{Azadi2018}.
We consider the standard deviations within each mode in Figure~\ref{fig:std} and the rate of ``high quality'' samples in Figure~\ref{fig:hqr}.
A sample is assigned to a mode if its $L_2$ distance is within four standard deviations ($\leq 4 \sigma = 0.2$) of its mean.
Samples within four standard deviations of any mixture component are considered ``high quality''.
The within standard deviation plot~(Figure~\ref{fig:std}) shows a slight improvement for MH-GAN, and the high quality sample rate~(Figure~\ref{fig:hqr}) approaches 100\% faster for the MH-GAN than the GAN or DRS\@.

To test the spread of the distribution, we inspect the categorical distribution of the closest mode.
Far away (non-high quality) samples are assigned to a 26th unassigned category.
This categorical distribution should be uniform over the 25 real modes for a perfect generator.
To assess generator quality, we look at the Jensen-Shannon divergence (JSD) between the sample mode distribution and a uniform distribution.
This is a much more stringent test of appropriate spread of probability mass than checking if a single sample is produced near a mode (as in~\citet{Azadi2018})\@.

\begin{figure*}[htbp]
    \centering
    \begin{subfigure}[b]{0.32\textwidth}
       \centering
       \includegraphics[width=2.2in]{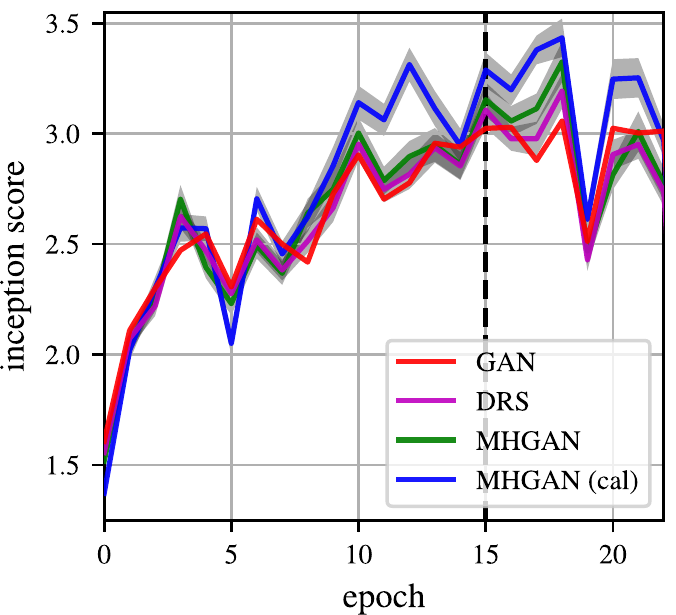}
       \caption{performance by epoch}
       \label{fig:incep_by_epoch}
    \end{subfigure}
    \begin{subfigure}[b]{0.32\textwidth}
       \centering
       \includegraphics[width=2.2in]{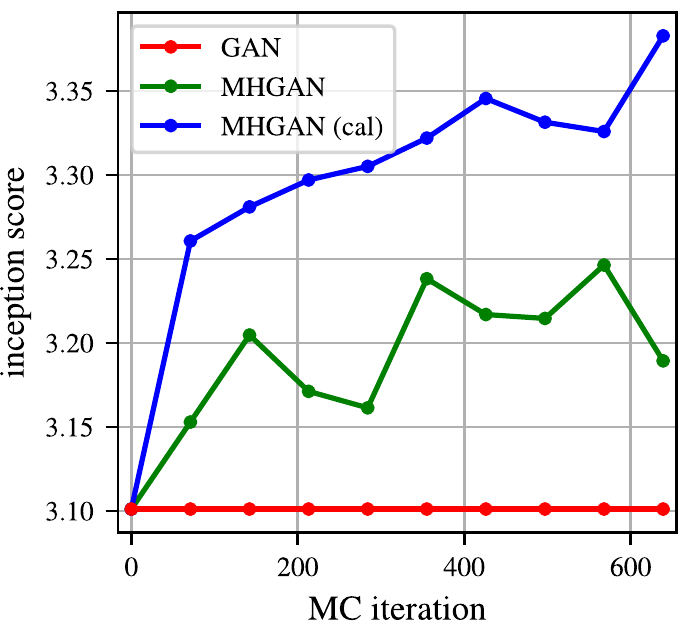}
       \caption{performance by MCMC iteration}
       \label{fig:incep_by_iter}
    \end{subfigure}
    \begin{subfigure}[b]{0.32\textwidth}
       \centering
       \includegraphics[width=2.2in]{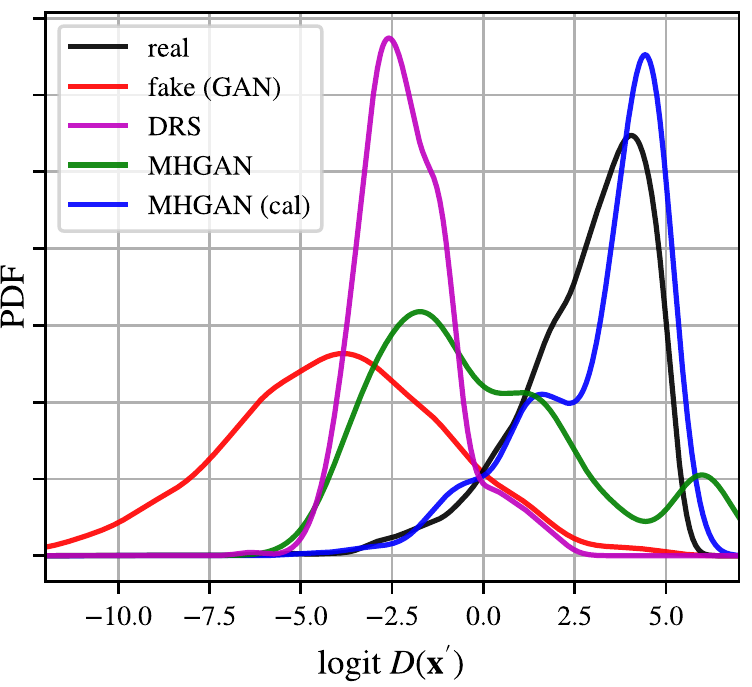}
       \caption{epoch 13 scores}
       \label{fig:score_dist_overlap}
    \end{subfigure}
    \caption{{\small
    Results of the MH-GAN experiments on CIFAR-10 using the DCGAN\@.
     \textbf{(a)} On the left, we show the Inception score vs.~training epoch of the DCGAN with $k=640$ MH iterations.
    MH-GAN denotes using the raw discriminator scores and MH-GAN (cal) for the calibrated scores.
    The error bars on MH-GAN performance (in gray) are computed using a t-test on the variation per batch across 80 splits of the Inception score.
     \textbf{(b)} In the center we show the Inception score vs.~number of MCMC iterations $k$ for the GAN at epoch 15.
     \textbf{(c)} On the right, we show the scores at epoch 13 where there is some overlap between the scores of fake and real images.
    When there is overlap, the MH-GAN corrects the $\PG$ distribution to have scores looking similar to the real data.
    DRS fails to fully shift the distribution because 1)~it does not use calibration and 2)~its ``$\gamma$ shift'' setup violates the validity of rejection sampling.
    }}
    \label{fig:inception}
\end{figure*}

In Figure~\ref{fig:jsd}, we see that the MH-GAN improves the JSD over DRS by $5 \times$ on average, meaning it achieves a much more balanced spread across modes.
DRS fails to make gains after epoch 30.
Using the principled approach of the MH-GAN along with calibrated probabilities ensures a correct spread of probability mass.

\subsection{Real Data}

For real data experiments we considered the CelebA~\citep{Liu2015} and CIFAR-10~\citep{Torralba2008} data sets modeled using the DCGAN~\citep{Radford2015} and WGAN~\citep{Arjovsky2017, Gulrajani2017}.
To evaluate the generator $G'$, we plot Inception scores~\citep{Salimans2016} per epoch in Figure~\ref{fig:incep_by_epoch} after $k=640$ MCMC iterations.
Figure~\ref{fig:incep_by_iter} shows Inception score per MCMC iteration:
most gains are made in the first $k=100$ iterations, but gains continue to $k=400$.
This shows that the MH-GAN allows a tunable trade-off between sample quality and computation cost.

\begin{figure*}
    \centering
    \begin{subfigure}[b]{0.49\textwidth}
       \centering
       \includegraphics[width=3.2in]{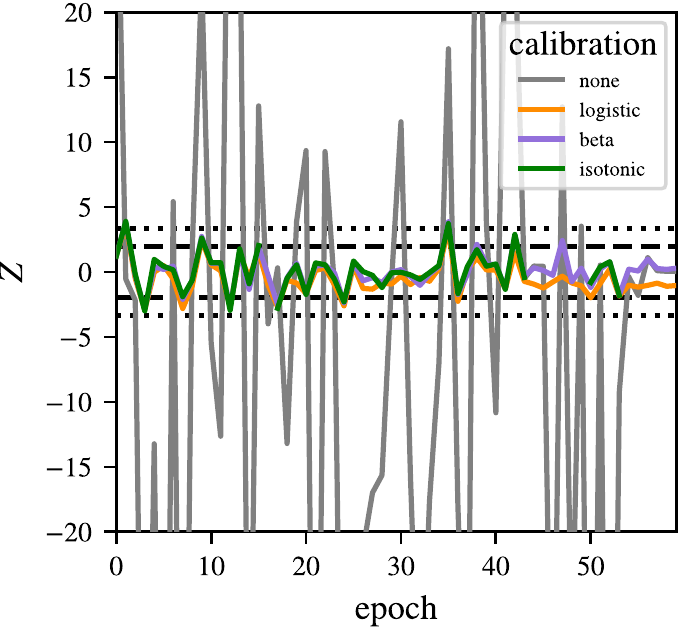}
       \caption{CIFAR-10}
       \label{fig:calibration cifar}
    \end{subfigure}
    \begin{subfigure}[b]{0.49\textwidth}
       \centering
       \includegraphics[width=3.2in]{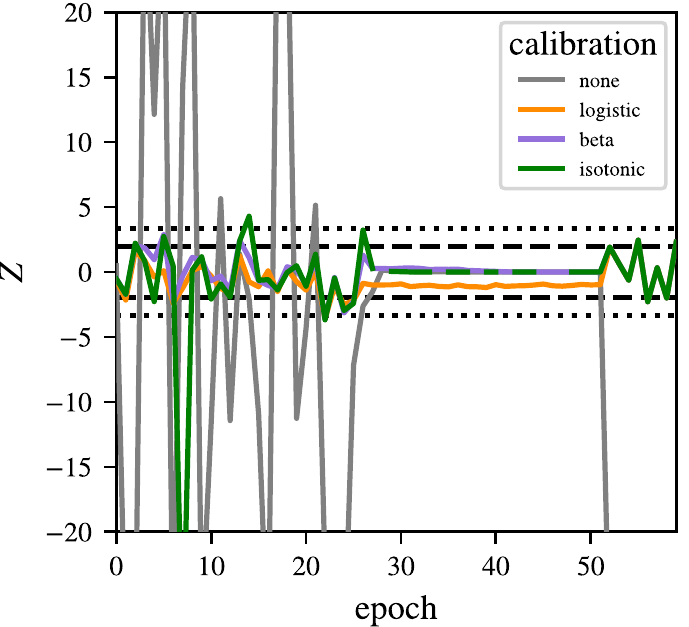}
       \caption{CelebA}
       \label{fig:calibration CelebA}
    \end{subfigure}
    \caption{{\small
    We show the calibration statistic $Z$~\eqref{eq:calib score} for the discriminator on held out data for the DCGAN\@.
    The results for CIFAR-10 are shown on the left \textbf{(a)}, and CelebA on the right \textbf{(b)}.
    The raw discriminator is clearly miscalibrated being far outside the region expected by chance (dashed black), and after multiple comparison correction (dotted black)\@.
    All the calibration methods give roughly equivalent results.
    CelebA has a period of training instability during epochs 30--50 which gives trivially calibrated classifiers.
    }}
    \label{fig:calibration}
\end{figure*}

\begin{table*}[htbp]
\begin{center}
  \caption{{\small
    Results showing Inception score improvements from MH-GAN on DCGAN and WGAN at epoch 60.
    Like Figure~\ref{fig:incep_by_epoch}, the error bars and p-values are computed using a paired t-test across Inception score batches (higher is better)\@.
    All results except for DCGAN on CelebA are significant at \smash{$p < 10^{-4}$}.
    WGAN does not learn a typical GAN discriminator that outputs a probability, so calibration is actually required in this case.
    }}
    \label{tbl:inception}
\vspace{1mm}
{\small
\begin{tabular}{|l|l|r|l|r||l|r|l|r|}
\toprule
~                 & \multicolumn{4}{c||}{DCGAN}                                & \multicolumn{4}{c|}{WGAN} \\
\toprule
~            & {CIFAR-10}                &       p   & CelebA                  &       p   & CIFAR-10                 &        p  & CelebA                 &       p  \\
\midrule
GAN          &        2.8789             &        -- &      2.3317             &        -- &       3.0734             &        -- &     2.7876             &       -- \\
DRS          &        2.977(77)          &    0.0131 &      2.511(50)          & $<$0.0001 &               ~          &         ~ &             ~          &        ~ \\
DRS (cal)    &        3.073(80)          & $<$0.0001 &      2.869(67)          & $<$0.0001 &       3.137(64)          &    0.0497 &     2.861(66)          &   0.0277 \\
MH-GAN       &        3.113(69)          & $<$0.0001 &      2.682(50)          & $<$0.0001 &               ~          &         ~ &             ~          &        ~ \\
MH-GAN (cal) &        \textbf{3.379}(66) & $<$0.0001 &      \textbf{3.106}(64) & $<$0.0001 &       \textbf{3.305}(83) & $<$0.0001 &     \textbf{2.889}(89) &   0.0266 \\
\bottomrule
\end{tabular}
}
\end{center}
\end{table*}

In Table~\ref{tbl:inception}, we summarize performance (Inception score) across all experiments, running MCMC to $k=640$ iterations in all cases.
Behavior is qualitatively similar to that in Figure~\ref{fig:incep_by_epoch}.
While DRS improves on a direct GAN, MH-GAN improves Inception score more in every case.
Calibration helps in every case;
and we found a slight advantage for isotonic regression over other calibration methods.
Results are computed at epoch 60, and as in Figure~\ref{fig:incep_by_epoch}, error bars and p-values are computed using a paired t-test across Inception score batches.
All results are significantly better than the baseline GAN at \smash{$p < 0.05$}.

\paragraph{Score distribution}
In Figure~\ref{fig:score_dist_overlap}, we show what $G'$ does to the distribution on discriminator scores.
MCMC shifts the distribution of the fakes to match the distribution on true images.
We also observed that the MH acceptance rate is primarily determined by the overlap of the distributions on $D$ scores between real and fake samples.
If the AUC of $D$ is less than 0.90 we see acceptance rates over 20\%; but when the AUC of $D$ is 0.95, acceptance rates drop to 10\%.

\paragraph{Calibration results}
Figure~\ref{fig:calibration} shows the results per epoch for both CIFAR-10 and CelebA\@.
It shows that the raw discriminator is highly miscalibrated, but can be fixed with any of the calibration methods.
The $Z$ statistic for the raw discriminator $D$ (DCGAN on CIFAR-10) varies from $-77.57$ to $48.98$ in the first 60 epochs; even after Bonferroni correction at $N \!\!=\!\! 60$, we expect $|Z| < 3.35$ with 95\% confidence for a calibrated classifier.
The calibrated discriminator varies from $-2.91$ to $3.60$, showing almost perfect calibration.
Accordingly, it is unsurprising that the calibrated discriminator significantly boosts performance in the MH-GAN\@.

\paragraph{Visual results}
We show example images from the CIFAR-10 and CelebA setups in the Appendix~\ref{sec:SI} (Figures~\ref{fig:cifar_samples}--\ref{fig:celeba_samples})\@.
The selectors (such as MH-GAN) result in a wider spread of probability mass across background colors.
For CIFAR-10, it enhances modes with animal-like outlines and vehicles.

\subsection{Progressive GAN}

To further illustrate the power of the MH-GAN approach we consider the progressive GAN (PGAN)~\citep{Karras2017}, which recently produced shockingly realistic images.
We applied the MH-GAN to a PGAN using the same setup as with DCGAN, at $k=800$.
We used the pre-trained network of~\citet{Karras2017} on CelebA-HQ (1024$\times$1024)\@.
Large batches of samples are in Appendix~\ref{sec:SI} (Figures~\ref{fig:PGAN samples}--\ref{fig:MHGAN 64x})\@.

In Table~\ref{tbl:pgan}, we use the PGAN as our base GAN and generate random samples from the base, as well as from the addition of DRS and MH-GAN selectors.
The different selectors (DRS and MH-GAN) are run on the same batches of images, so the same images may appear for both generators.
Although the PGAN sometimes produces near photorealistic images, it also produces many flawed nightmare like images.
To assess image quality, five human labelers manually labeled images as warped or acceptable.
Table~\ref{tbl:pgan} shows that MH-GAN selects significantly fewer warped images.

Both DRS and MH-GAN show an ability to select just the realistic images.
The MH-GAN samples are nearly perfect, while DRS still has many flawed samples.

\section{Conclusions}
\label{sec:conclusions}

We have shown how to incorporate the knowledge in the discriminator $D$ into an improved generator $G'$.
Our method is based on the premise that $D$ is better at density ratio estimation than $G$ is at sampling data, which may be a harder task.
The principled MCMC setup selects among samples from $G$ to correct biases in $G$.
This is the only method in the literature which has the property that given a perfect $D$ one can recover $G$ such that $\PG = \PR$.

We have shown the raw discriminators in GANs and DRS are poorly calibrated.
To our knowledge, this is the first work to evaluate the discriminator in this way and to rigorously show the poor calibration of the discriminator.
Because the MH-GAN algorithm may be used to wrap any other GAN, there are countless possible use cases.

\fboxsep=1.5pt  
\newcommand{\unreal}[1]{\fcolorbox{white}{red}{#1}}

\begin{table*}[htbp]
\vspace{-3mm}  
\begin{center}
  \caption{{\small
    We show 16 \emph{random samples} from the PGAN, calibrated (improved) DRS, and MH-GAN, from the same sequence of $G$ samples.
    There are 5 cases where the PGAN produces bad warpings (red) while the MH-GAN does not, and 0 cases where the MH-GAN does and the PGAN does not;
    for DRS, there are 7 where only DRS is warped, and 1 where only MH-GAN is warped.
    Even with 16 samples, the MH-GAN is better under a one-sided pairwise trinomial test~\citep{Coakley1996} at $p=0.017$ vs DRS and $p=0.013$ vs PGAN\@.
    }}
  \label{tbl:pgan}
\setlength\tabcolsep{1pt} 
\begin{tabular}{cc|cc|cc}
\multicolumn{2}{c|}{PGAN (base)} & \multicolumn{2}{c|}{PGAN with DRS (cal)} & \multicolumn{2}{c}{PGAN with MH-GAN (cal)} \\
\toprule
\unreal{\includegraphics[width=\pganw]{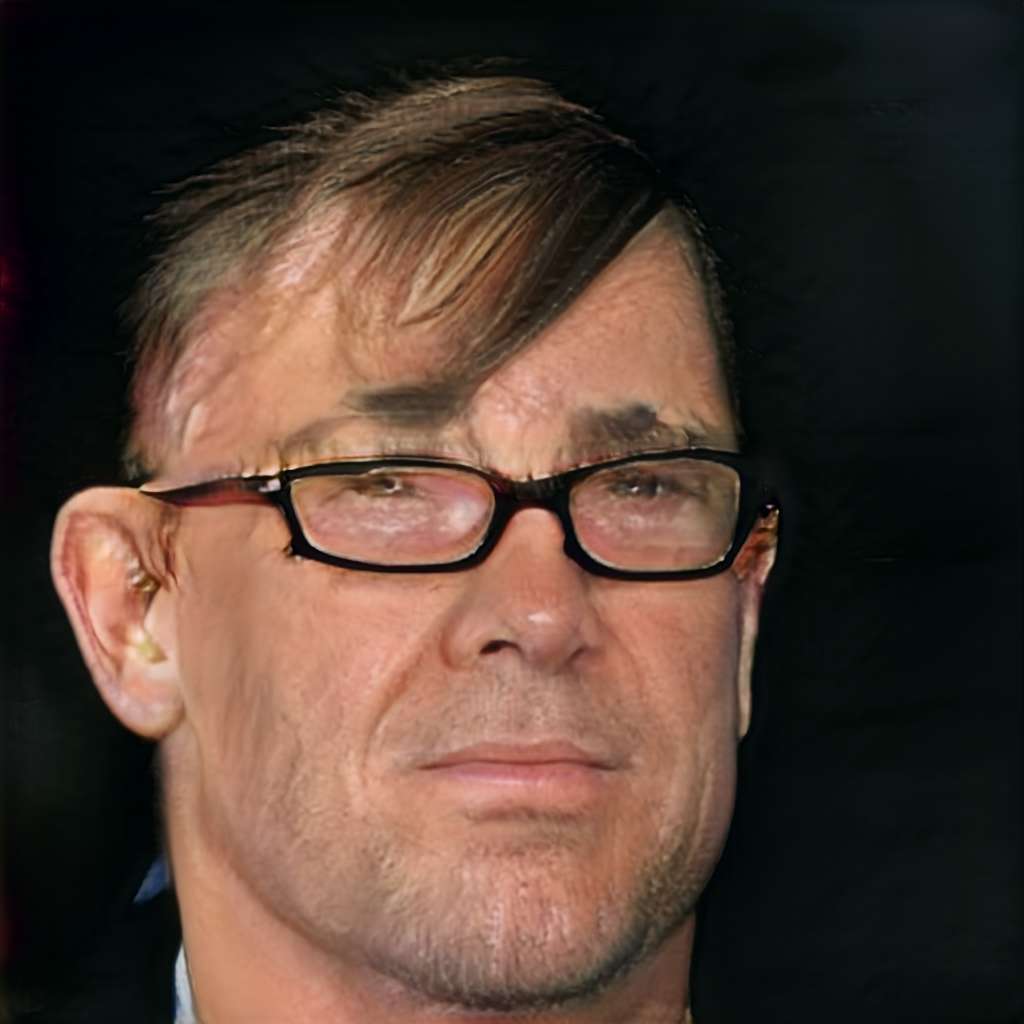}} &
\includegraphics[width=\pganw]{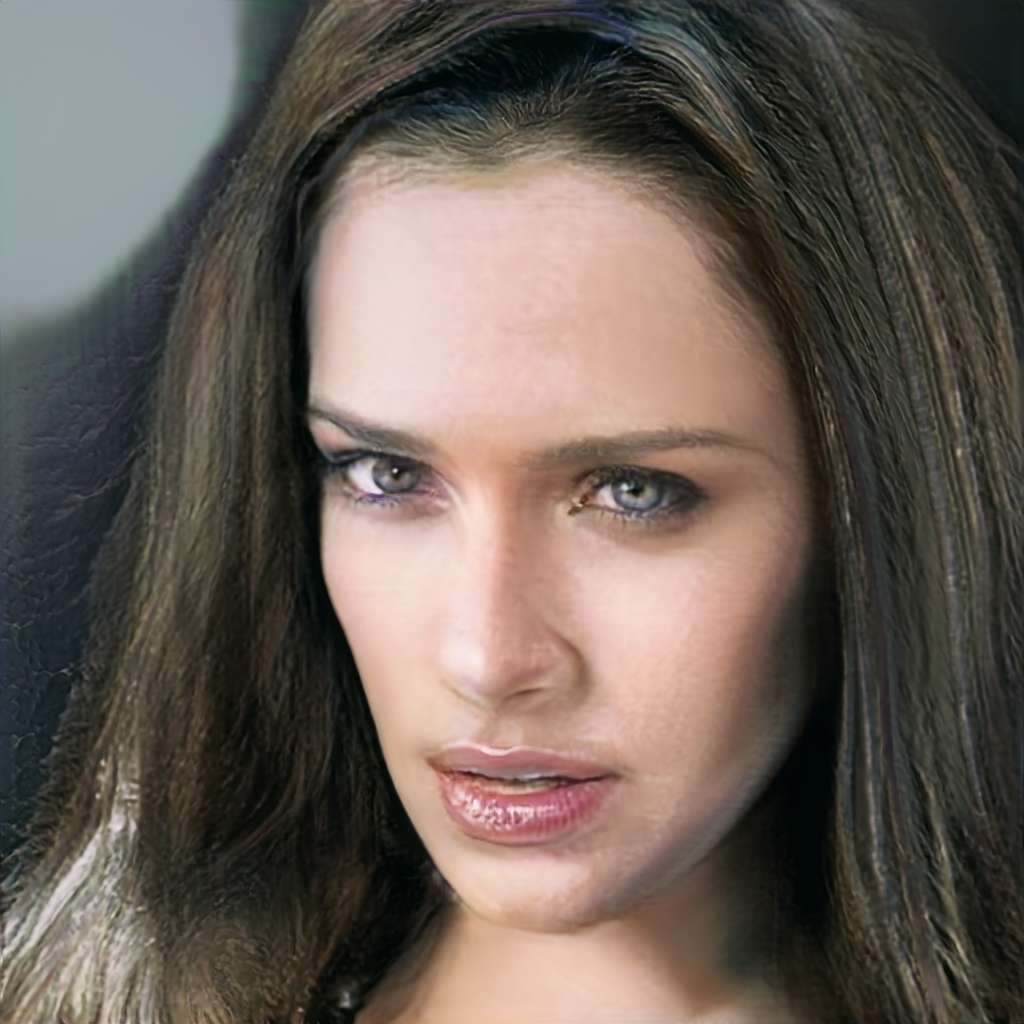} &
\unreal{\includegraphics[width=\pganw]{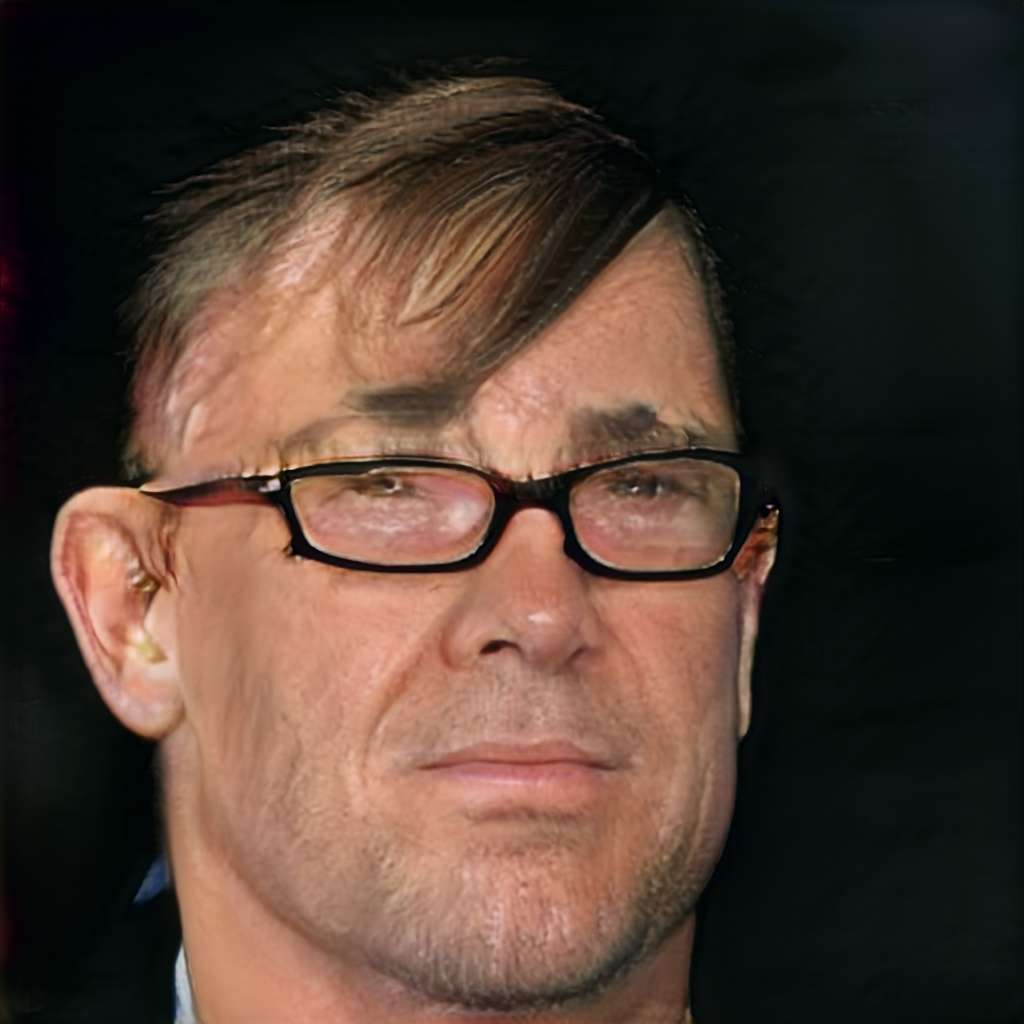}} &
\includegraphics[width=\pganw]{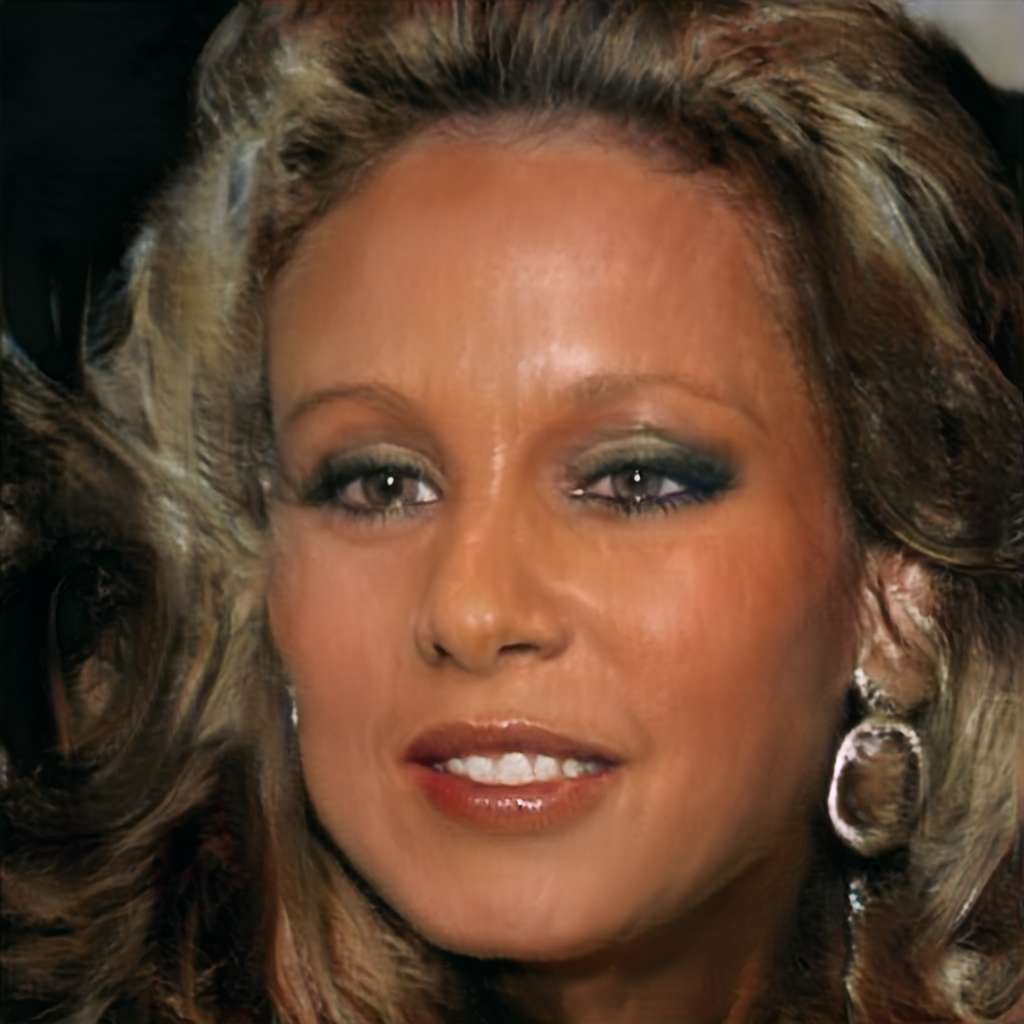} &
\includegraphics[width=\pganw]{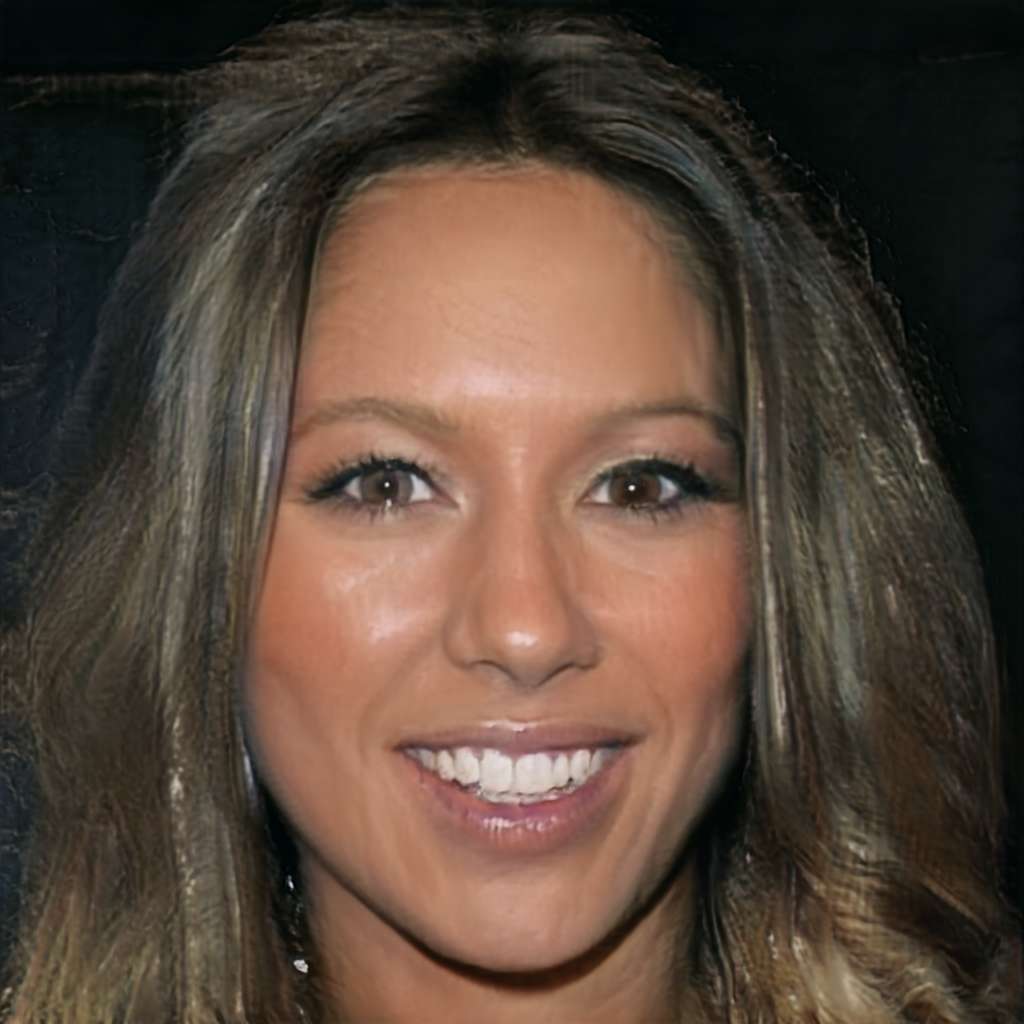} &
\includegraphics[width=\pganw]{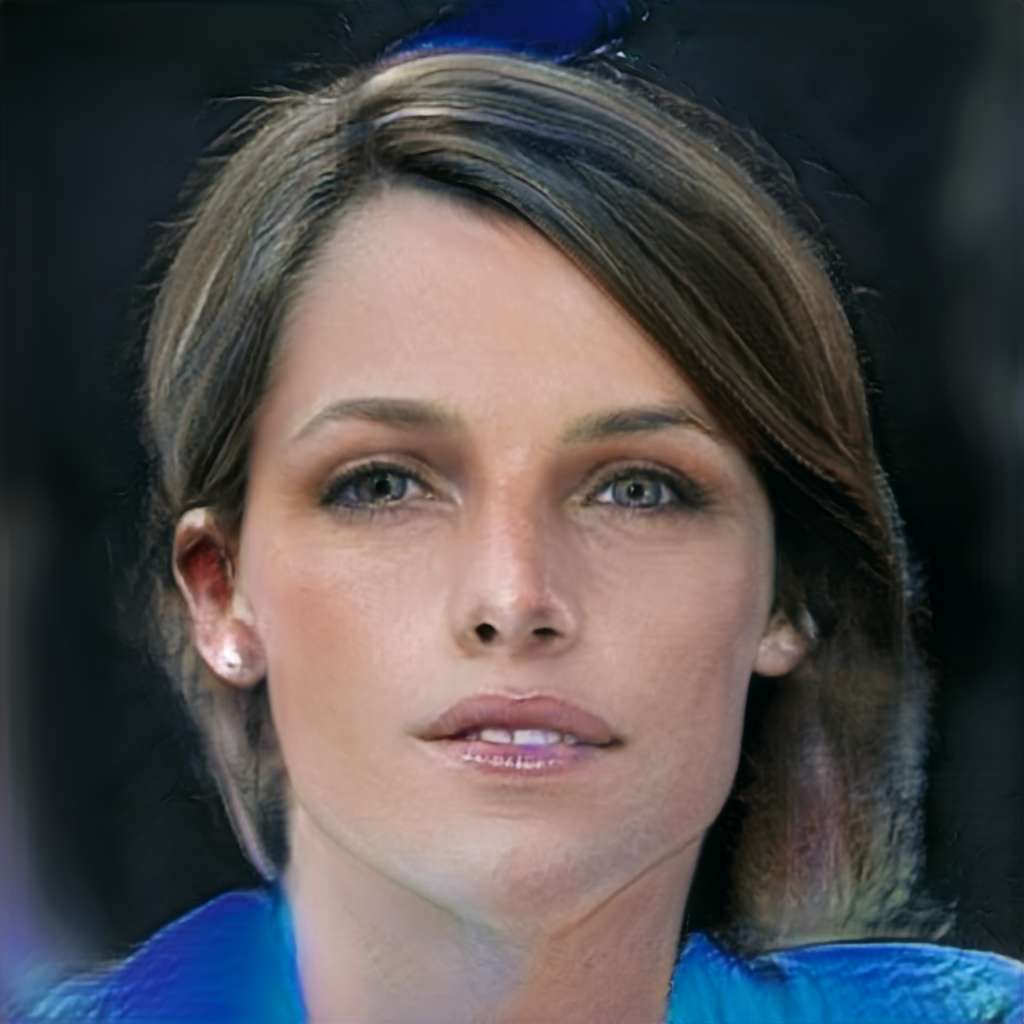} \\
\unreal{\includegraphics[width=\pganw]{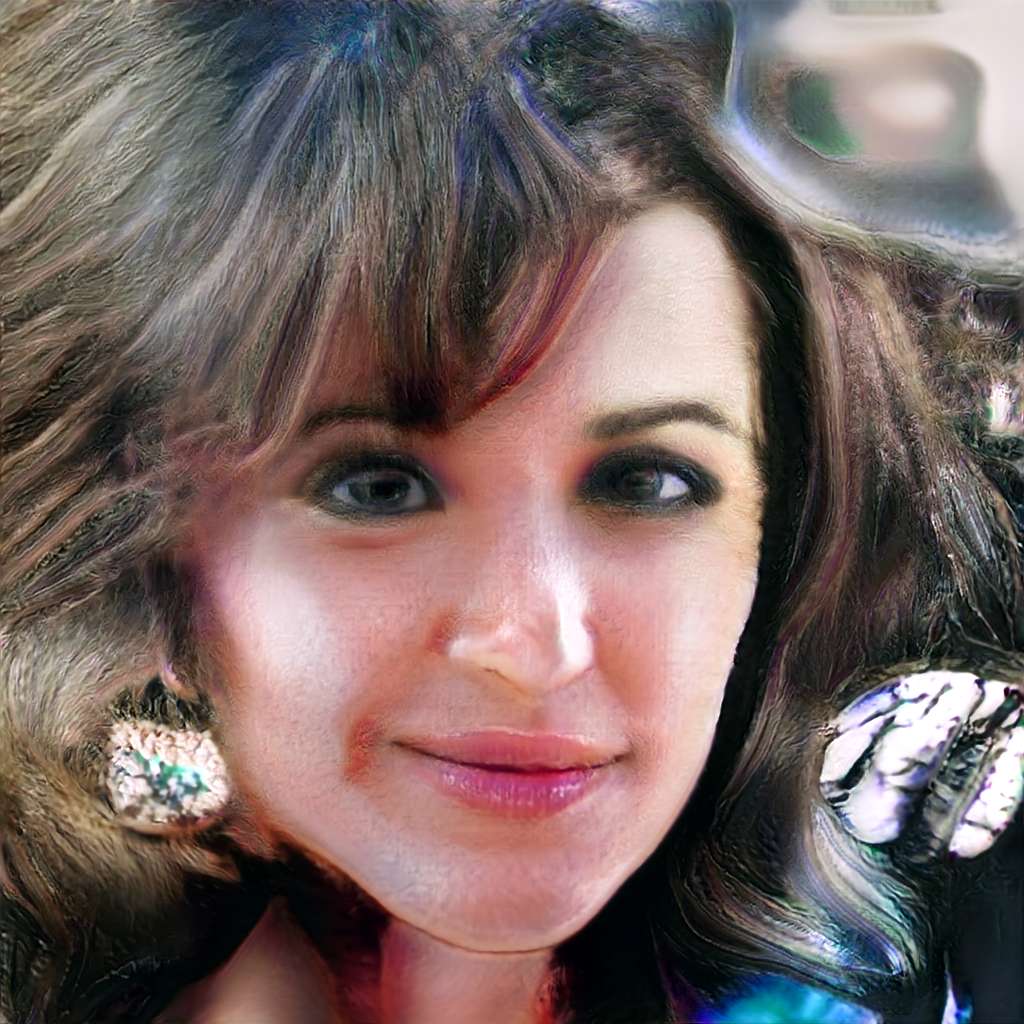}} &
\unreal{\includegraphics[width=\pganw]{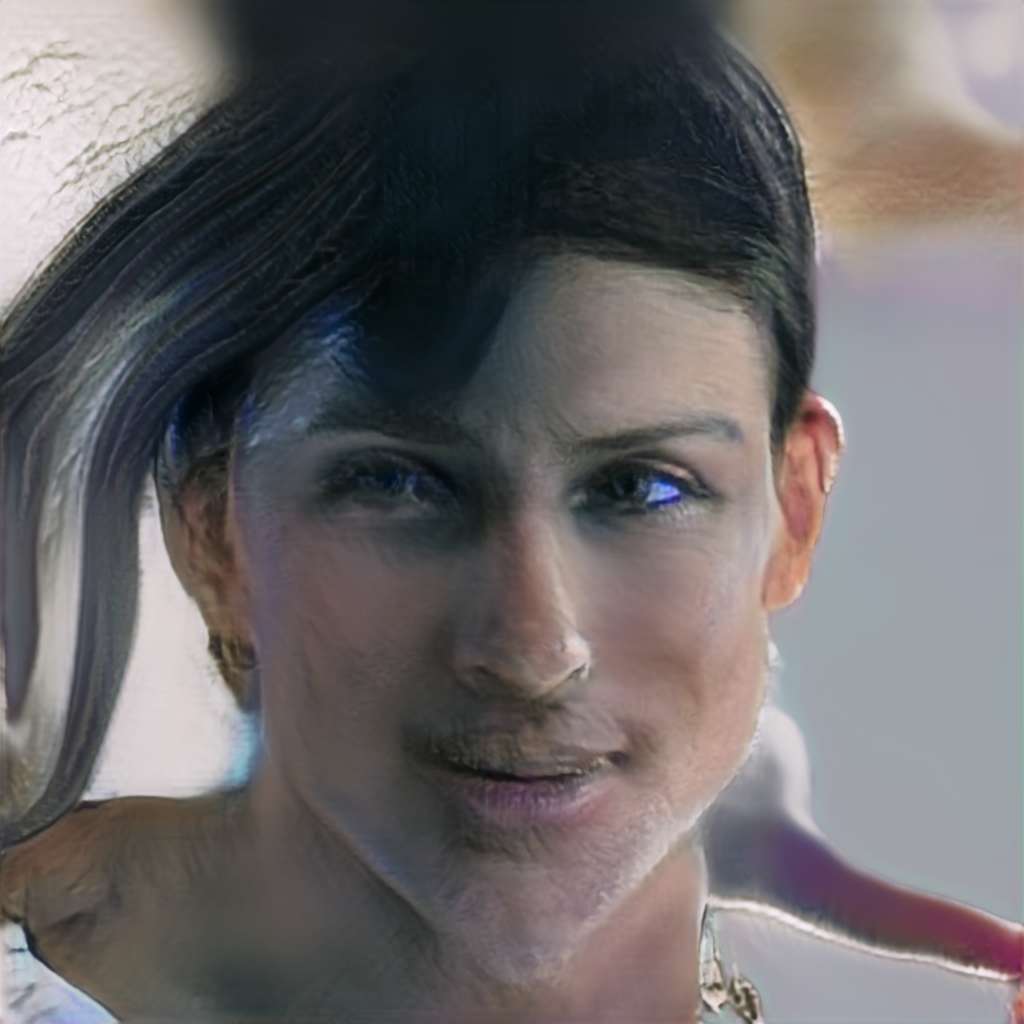}} &
\includegraphics[width=\pganw]{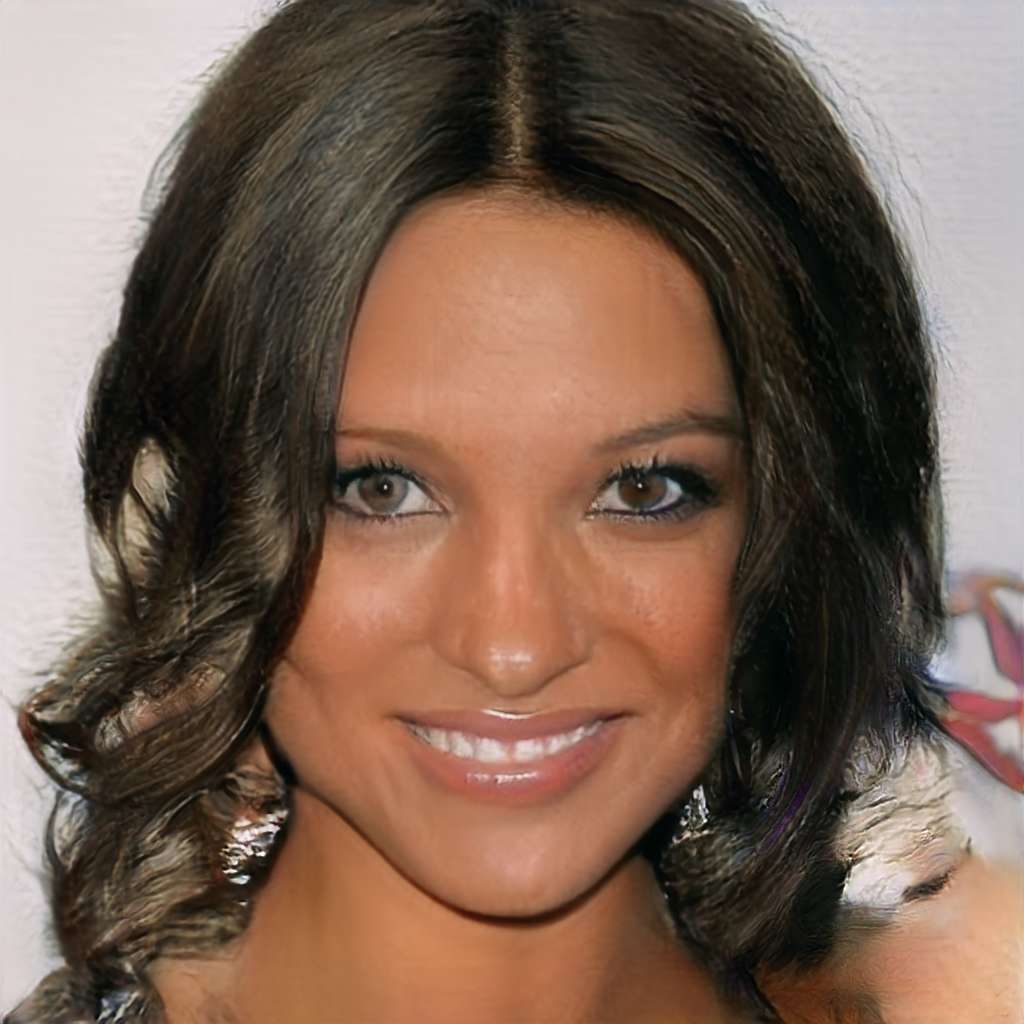} &
\unreal{\includegraphics[width=\pganw]{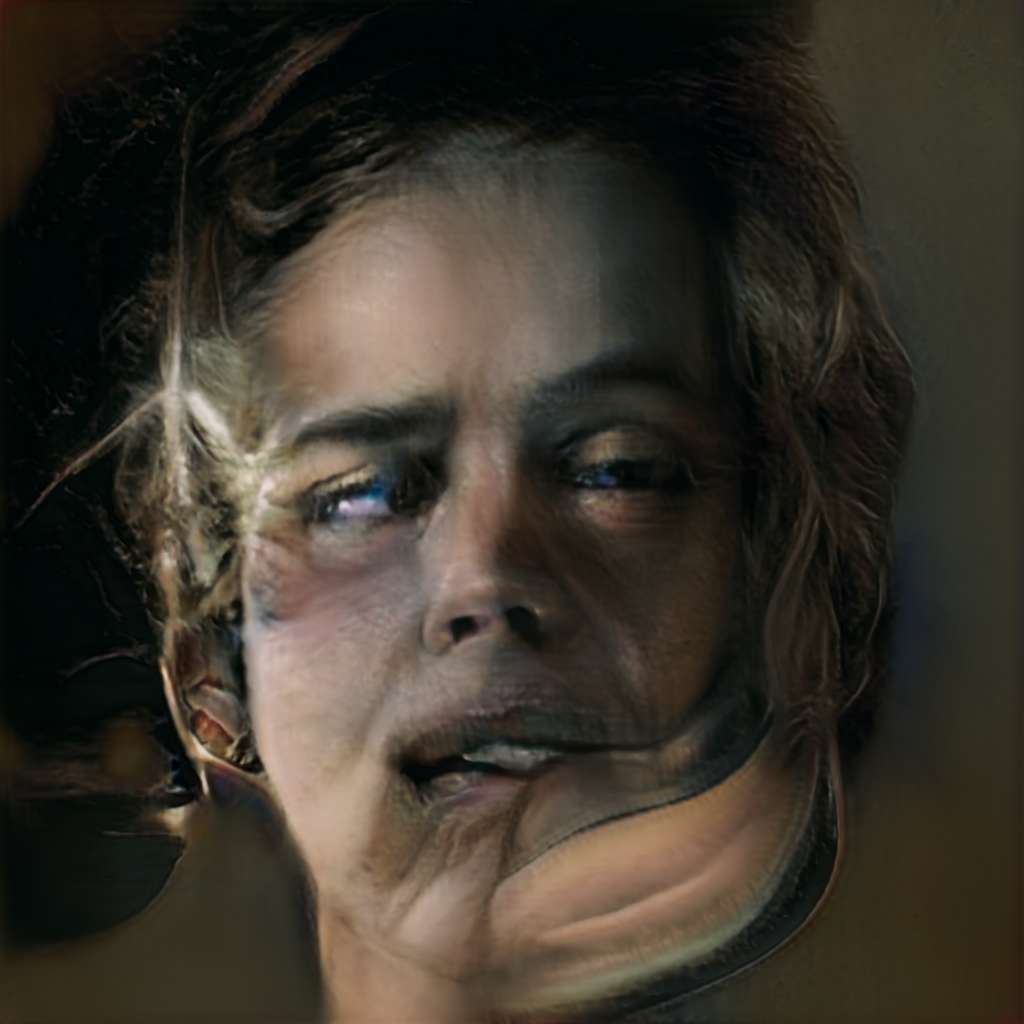}} &
\includegraphics[width=\pganw]{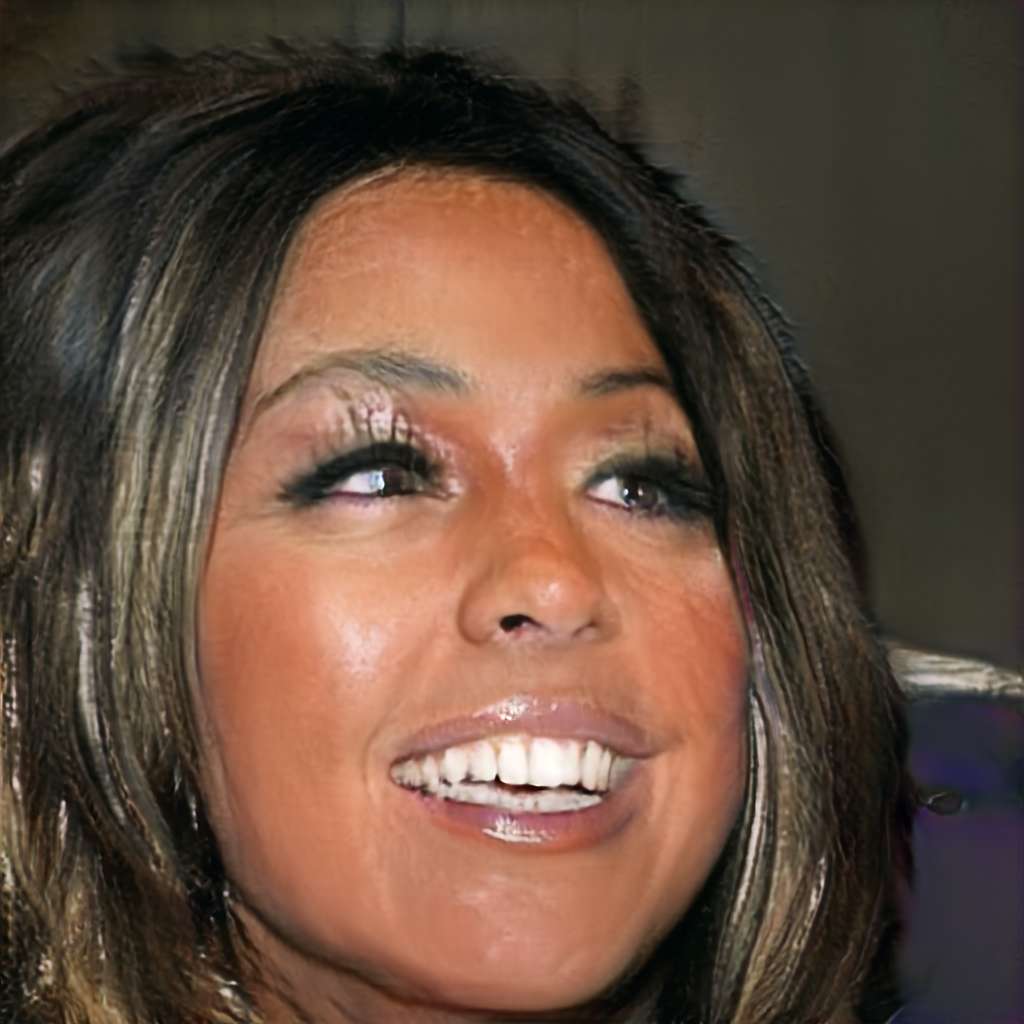} &
\includegraphics[width=\pganw]{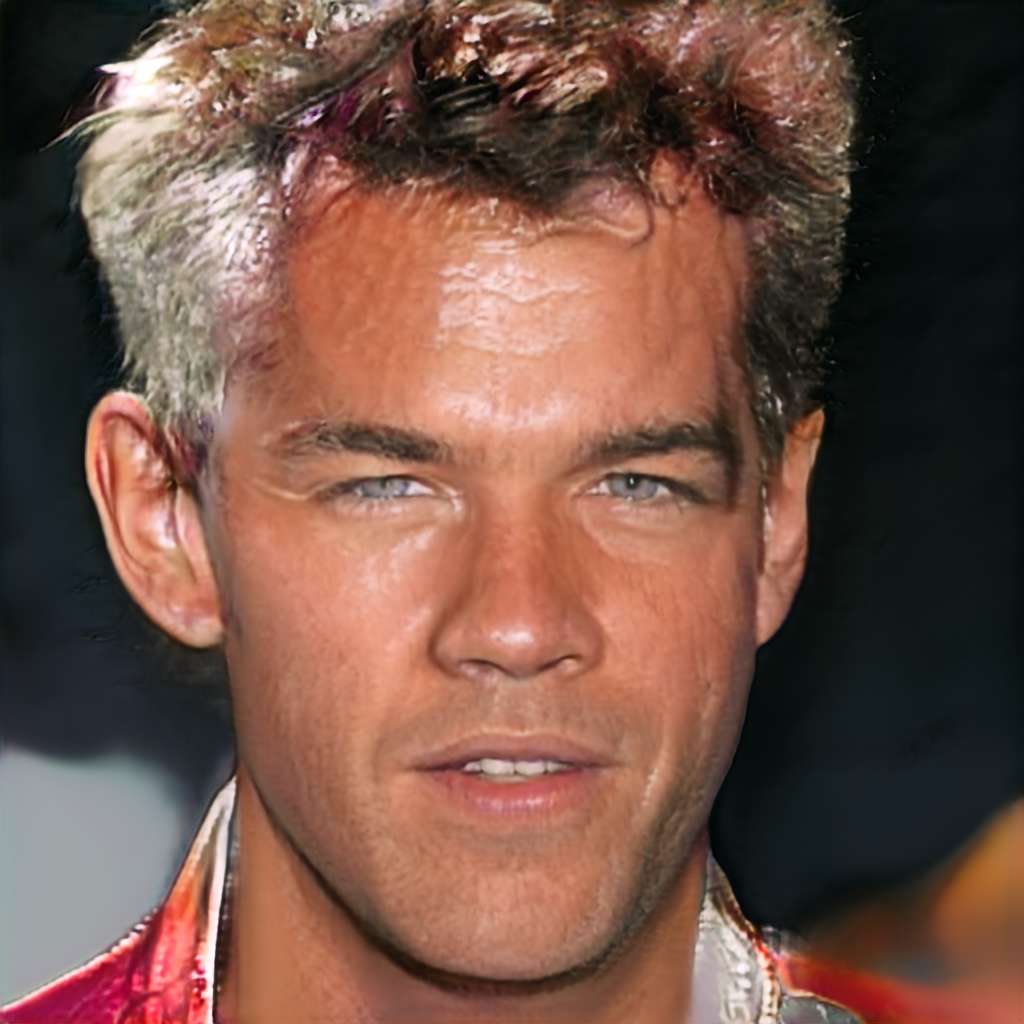} \\
\includegraphics[width=\pganw]{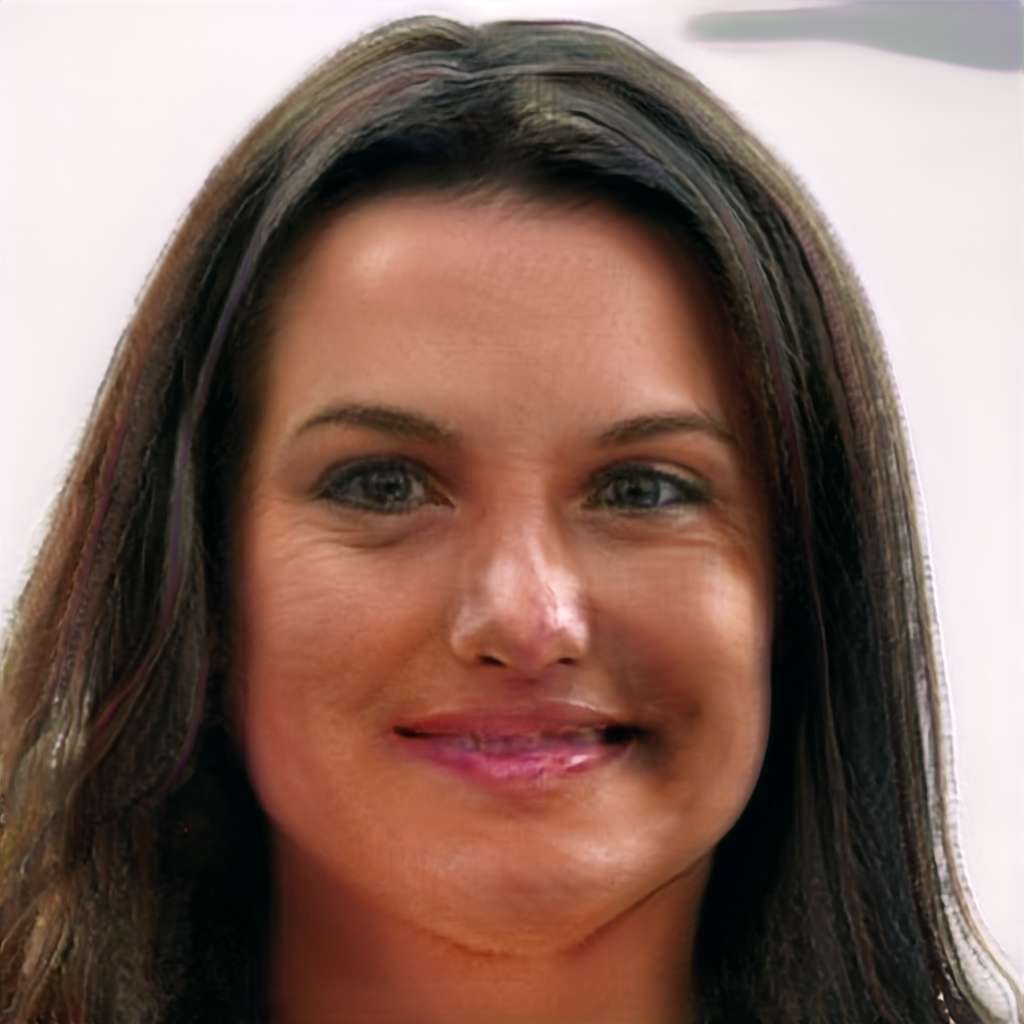} &
\unreal{\includegraphics[width=\pganw]{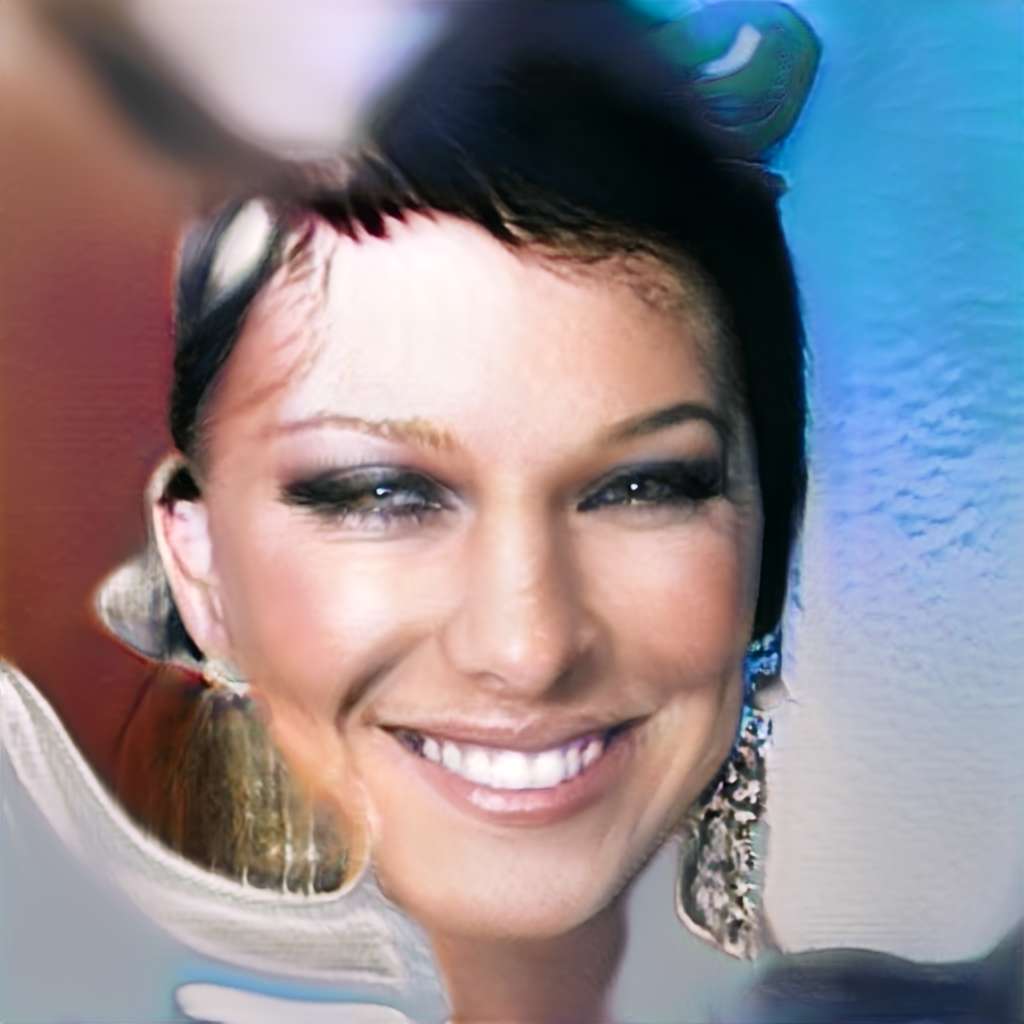}} &
\unreal{\includegraphics[width=\pganw]{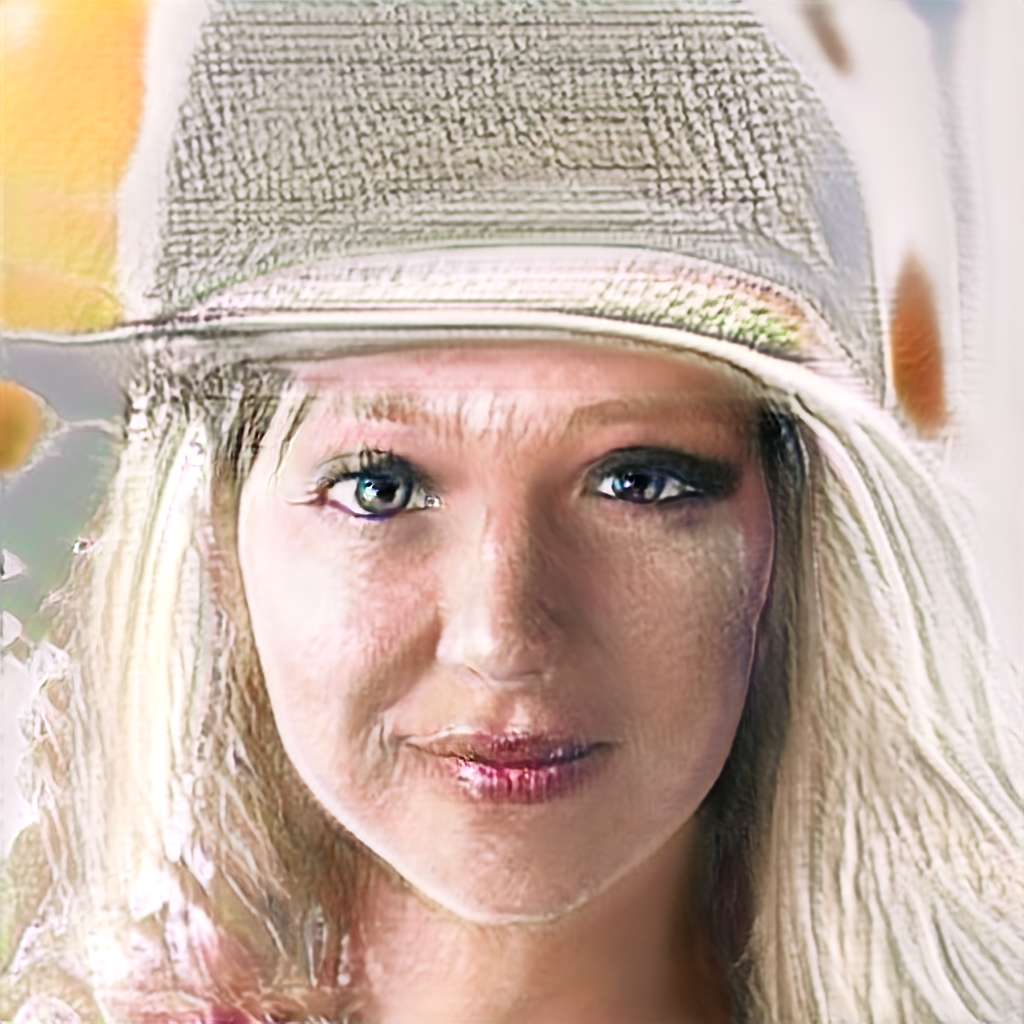}} &
\includegraphics[width=\pganw]{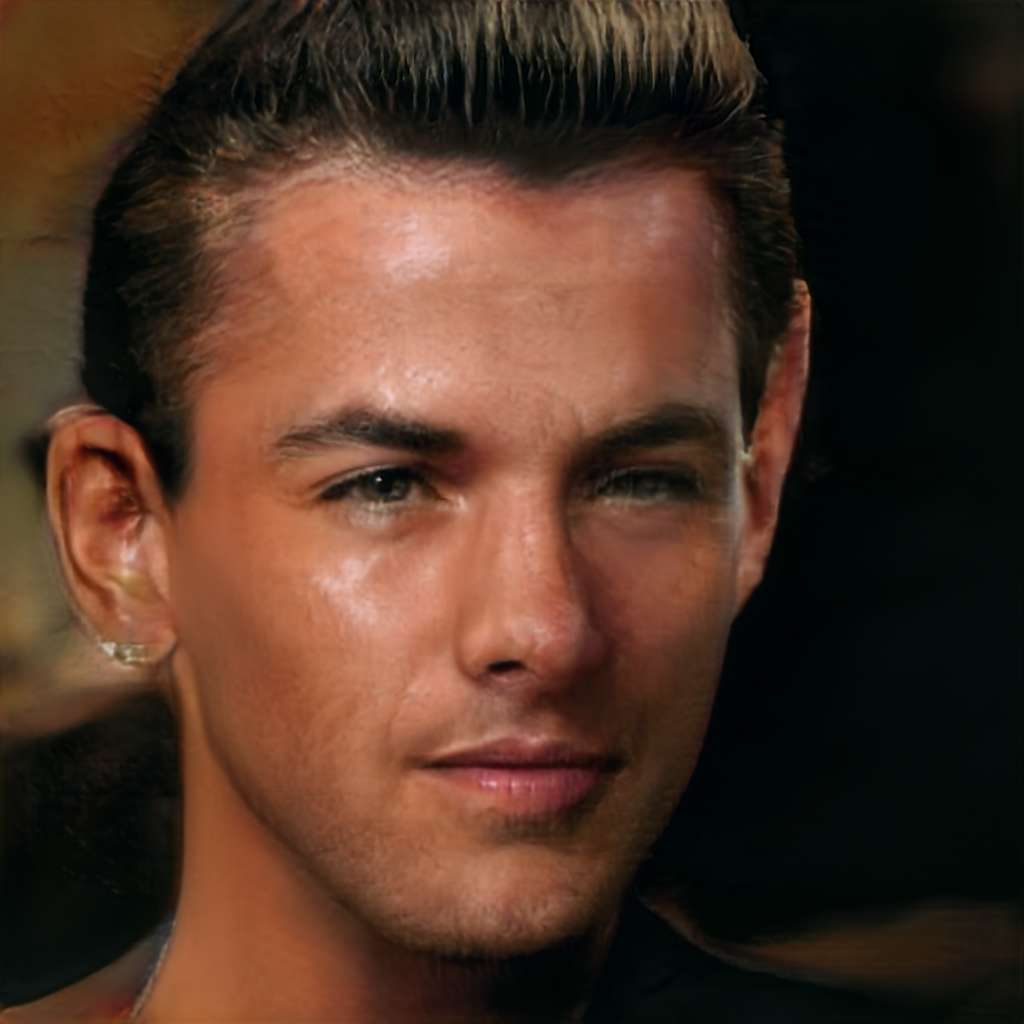} &
\includegraphics[width=\pganw]{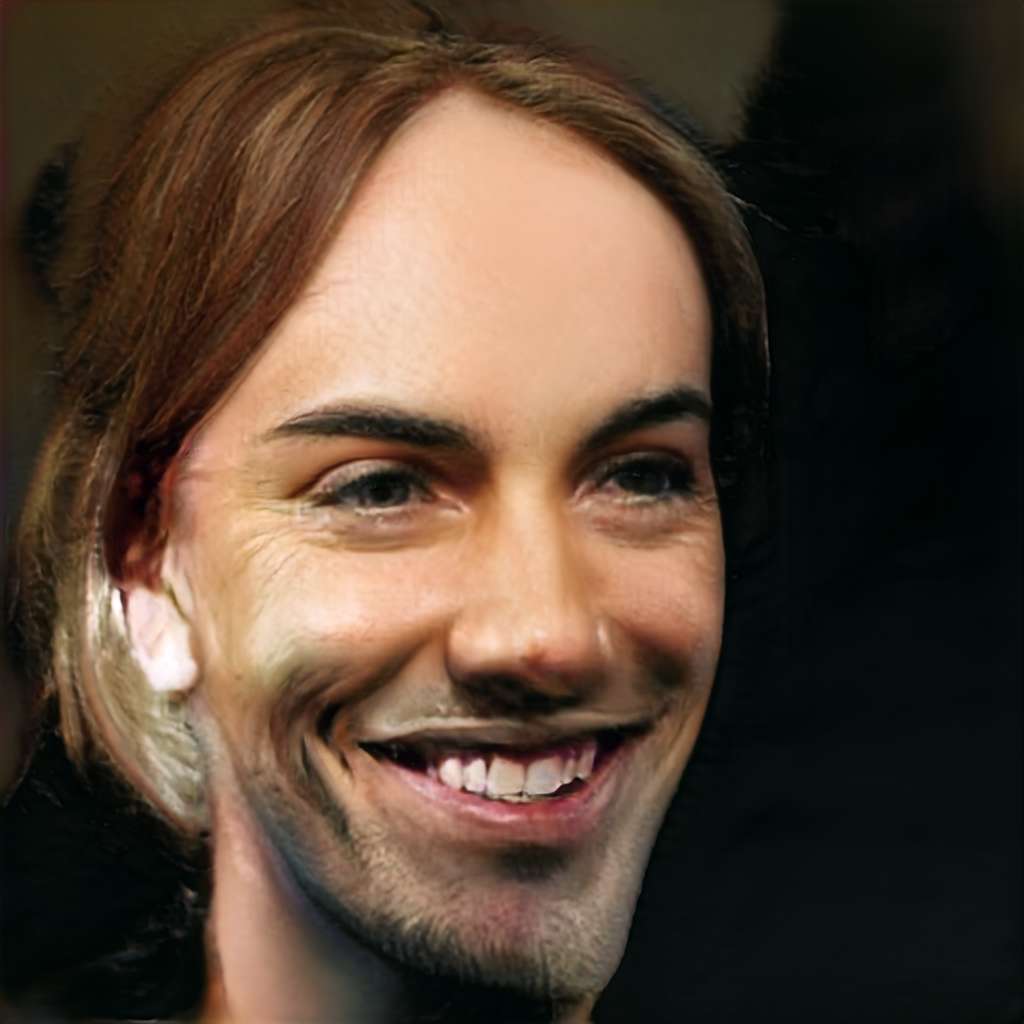} &
\includegraphics[width=\pganw]{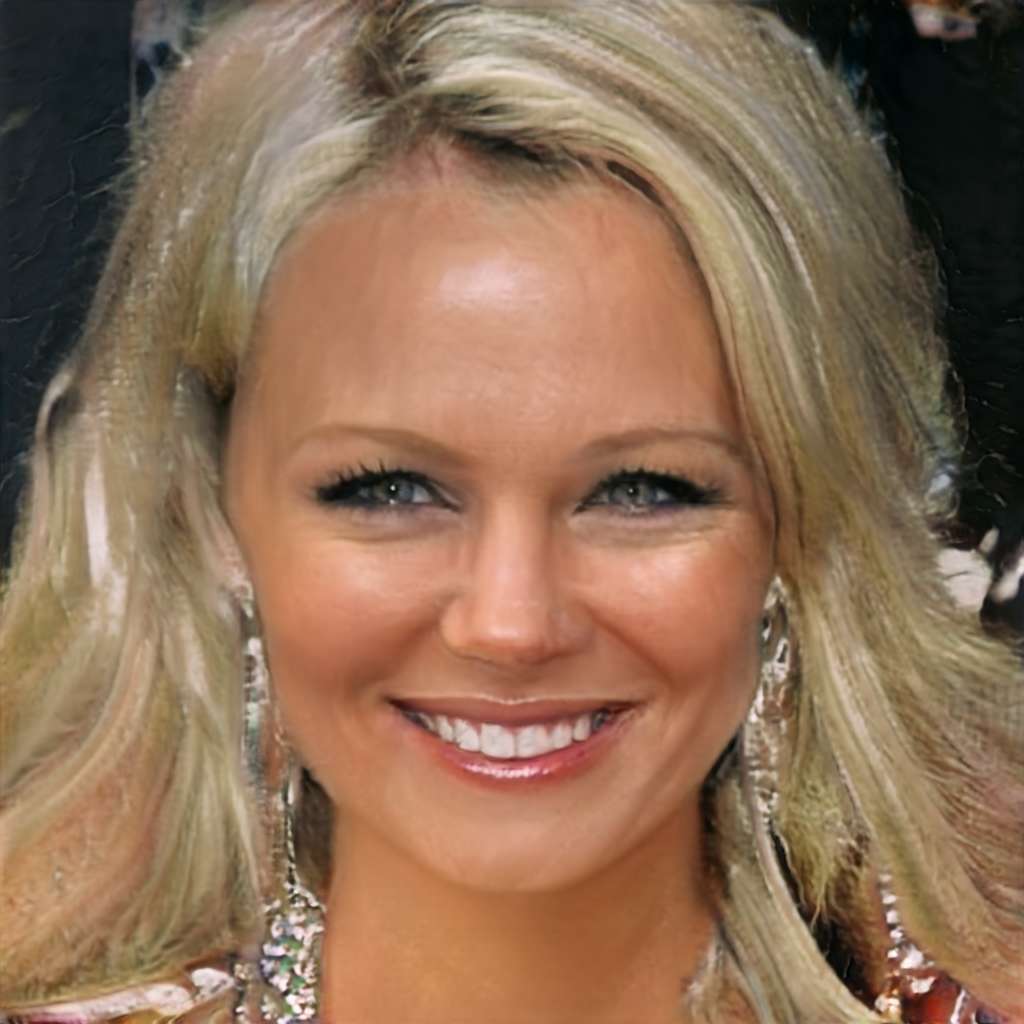} \\
\includegraphics[width=\pganw]{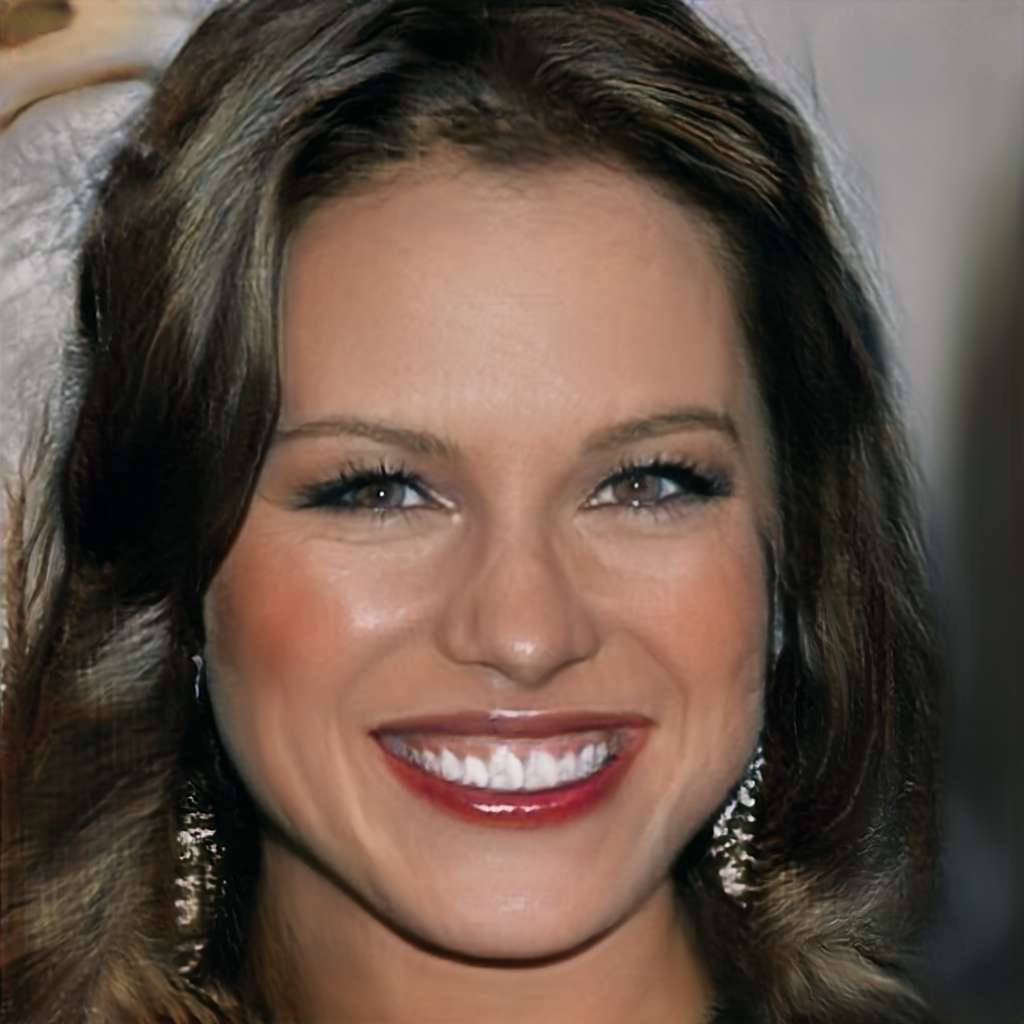} &
\includegraphics[width=\pganw]{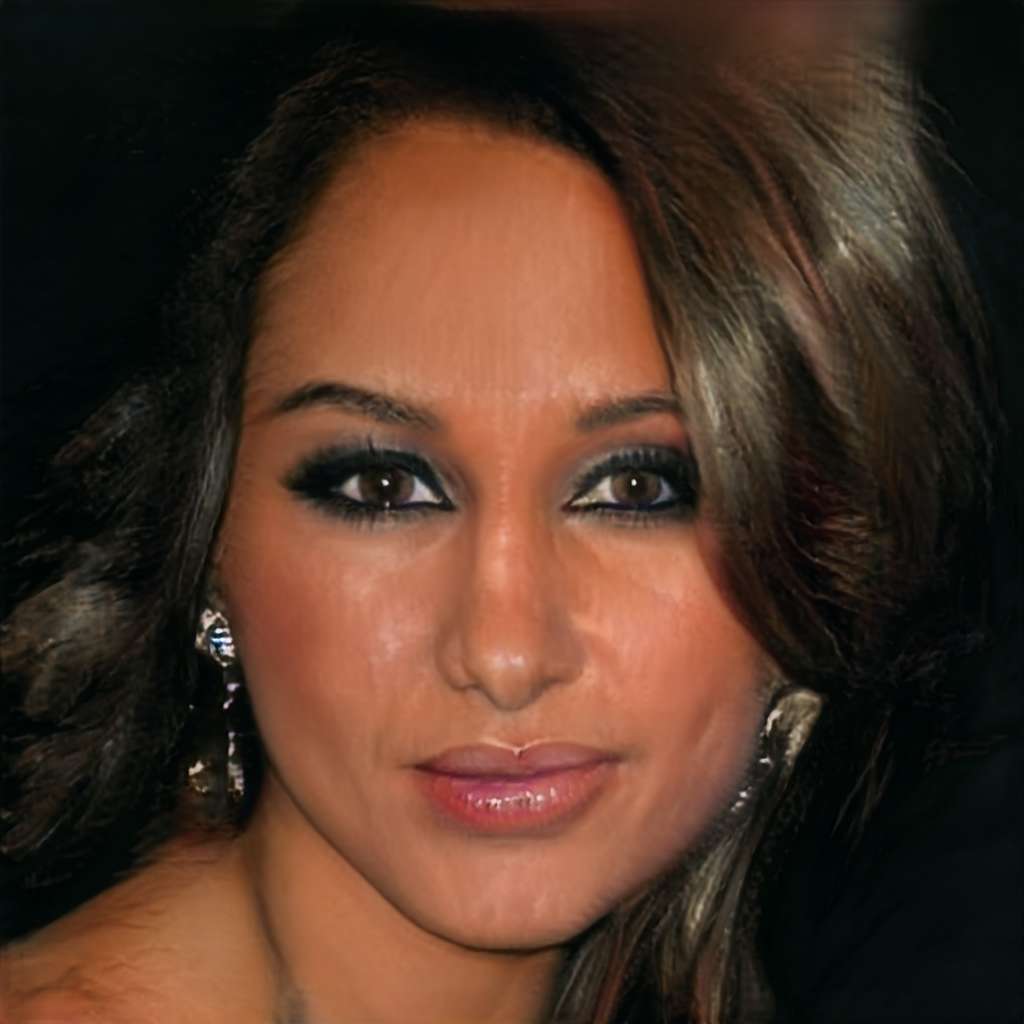} &
\includegraphics[width=\pganw]{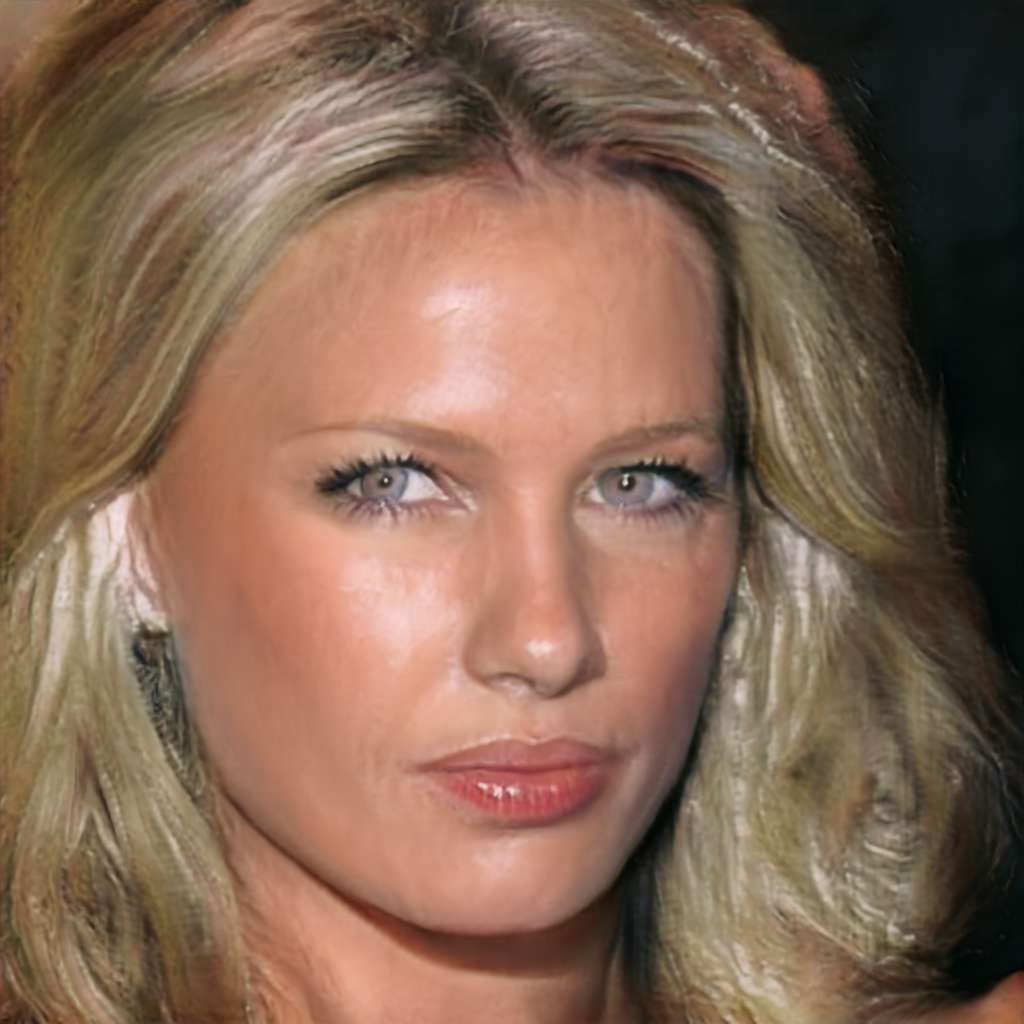} &
\unreal{\includegraphics[width=\pganw]{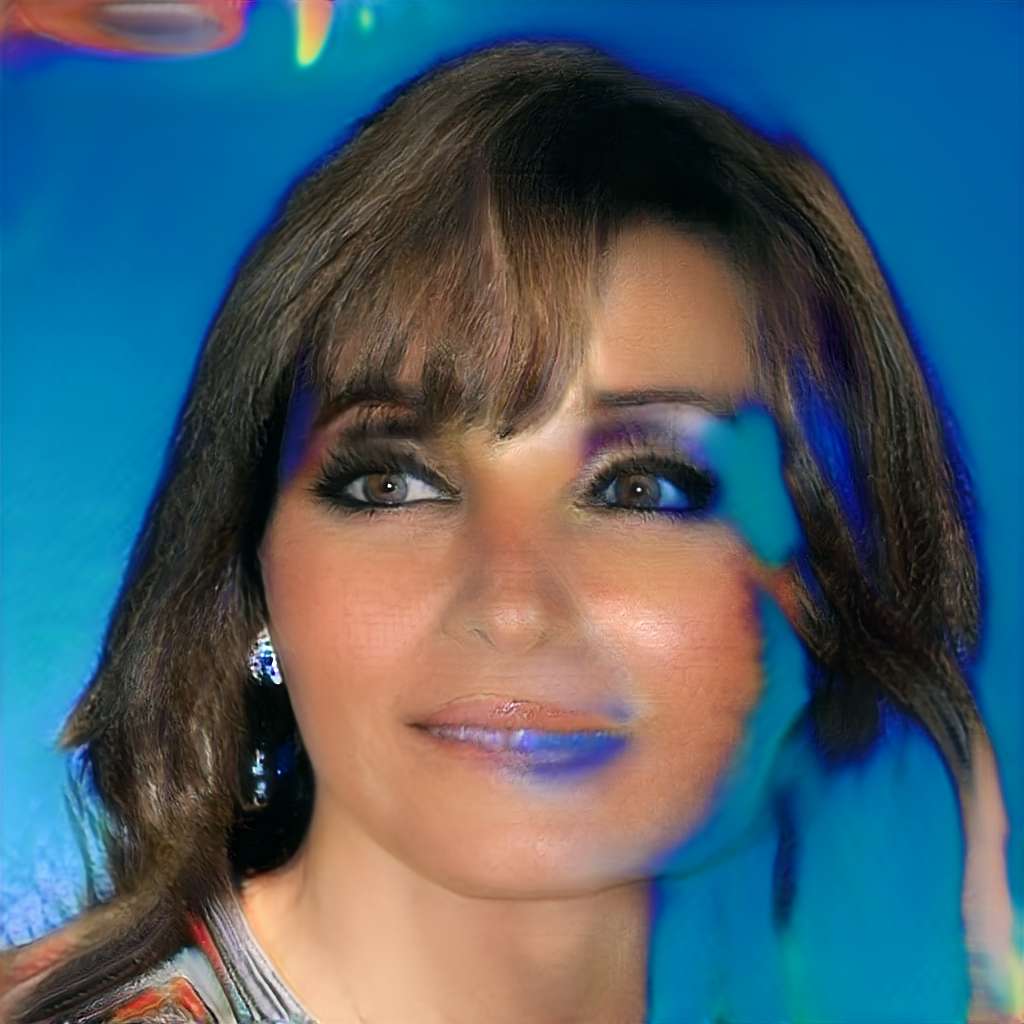}} &
\includegraphics[width=\pganw]{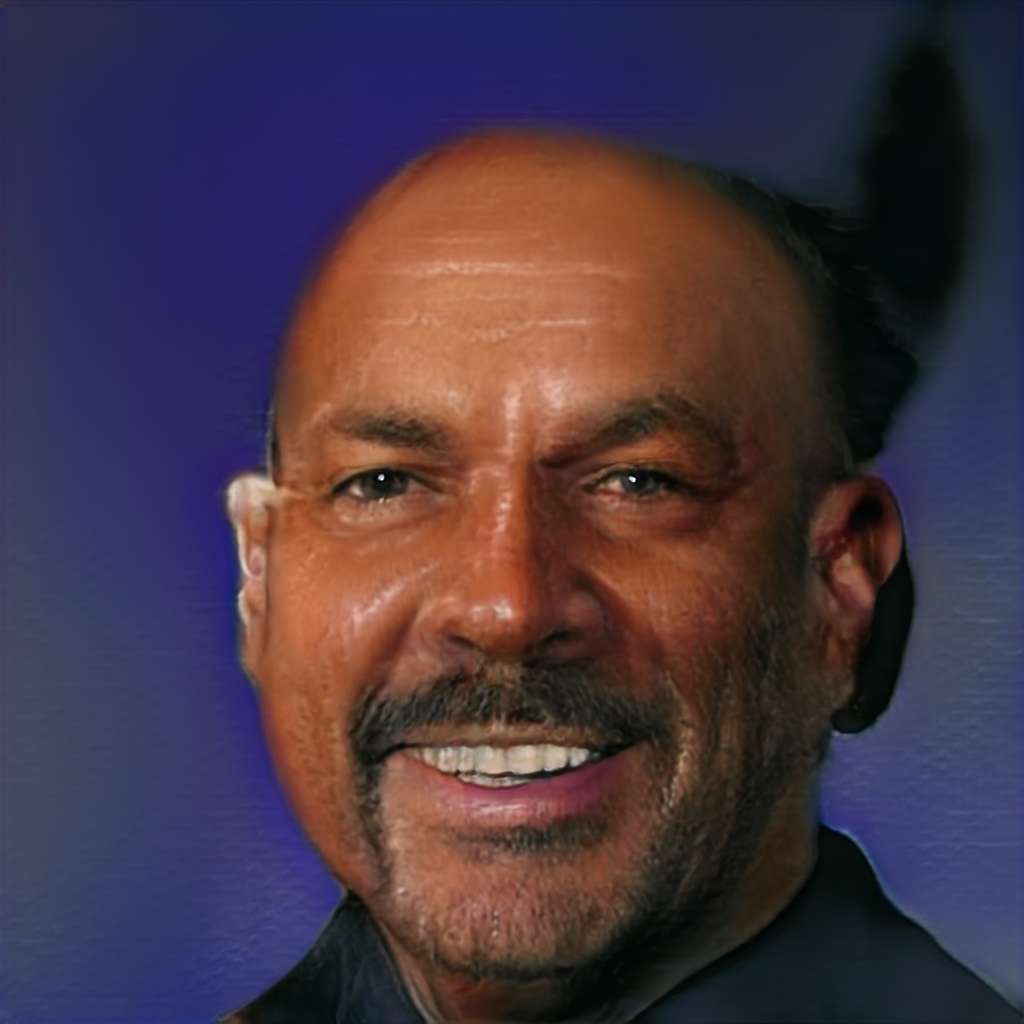} &
\includegraphics[width=\pganw]{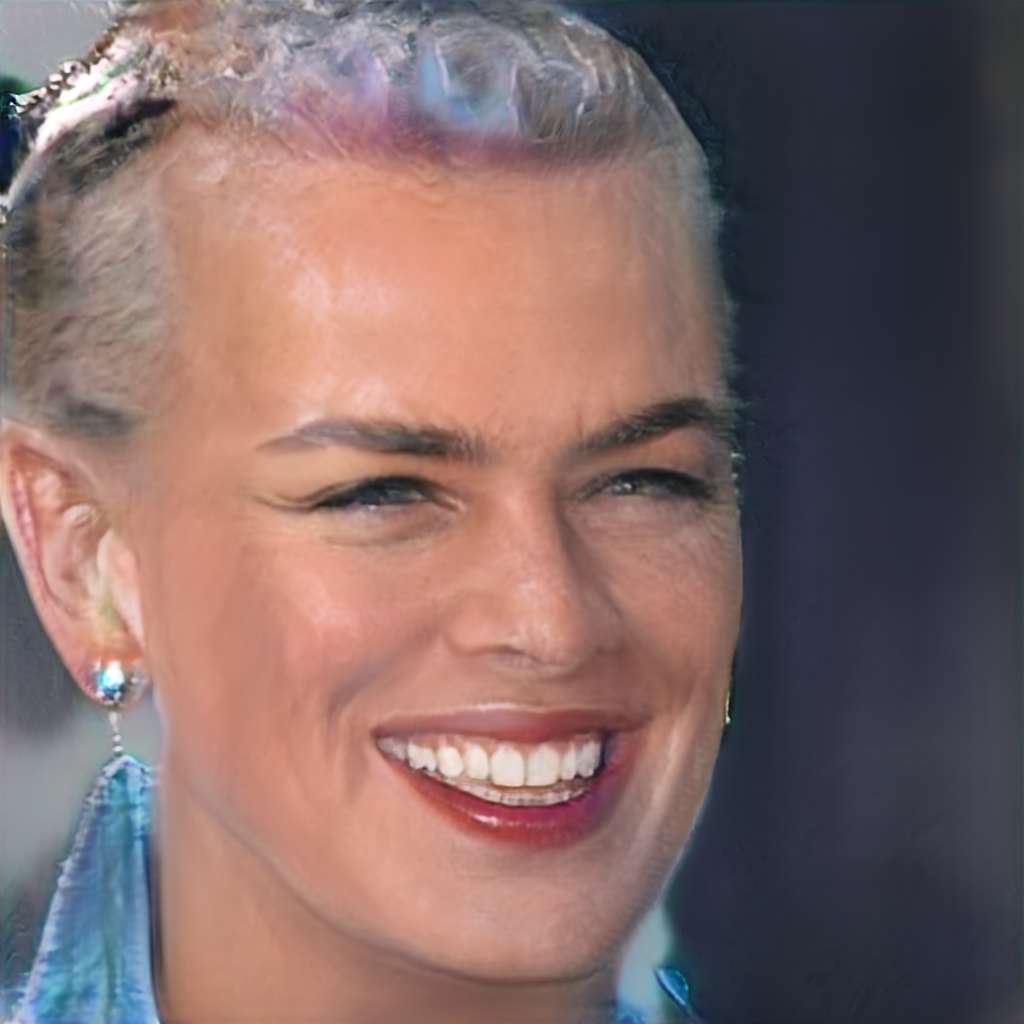} \\
\includegraphics[width=\pganw]{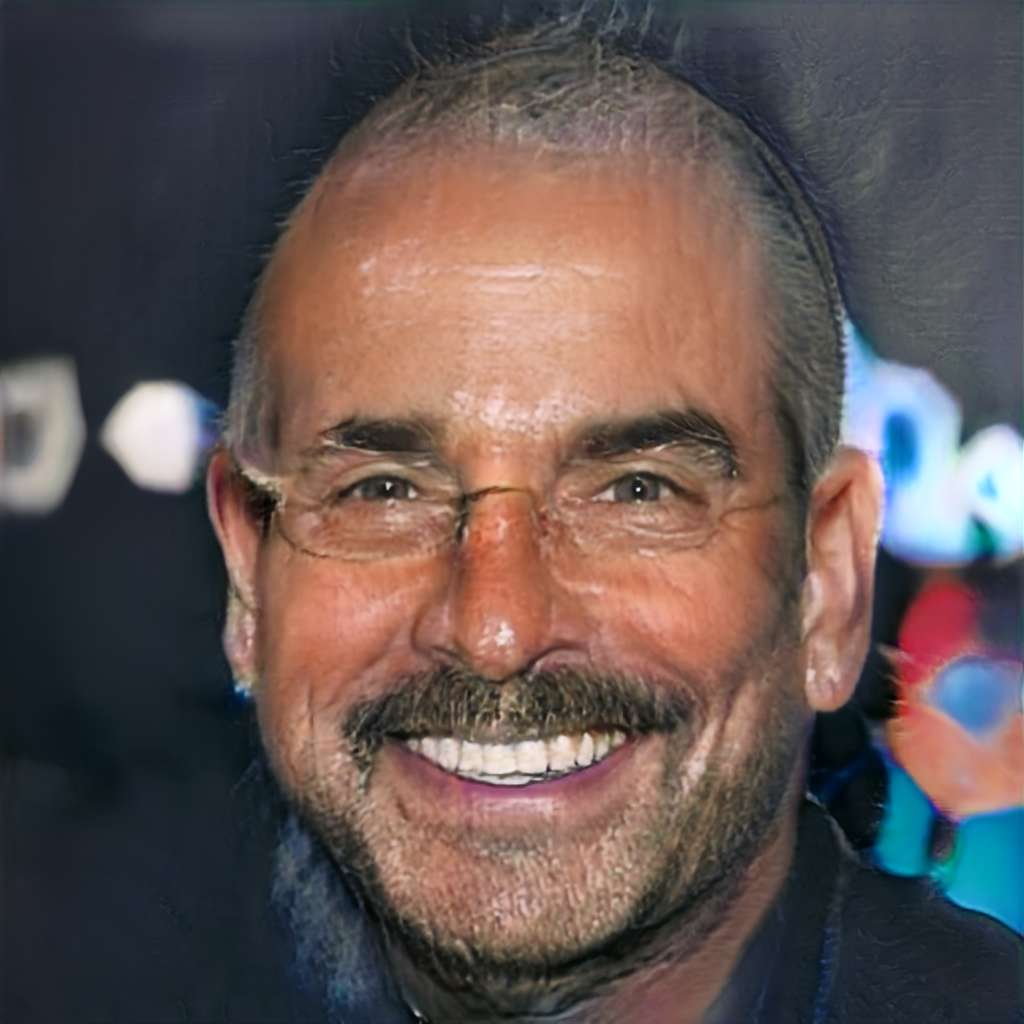} &
\includegraphics[width=\pganw]{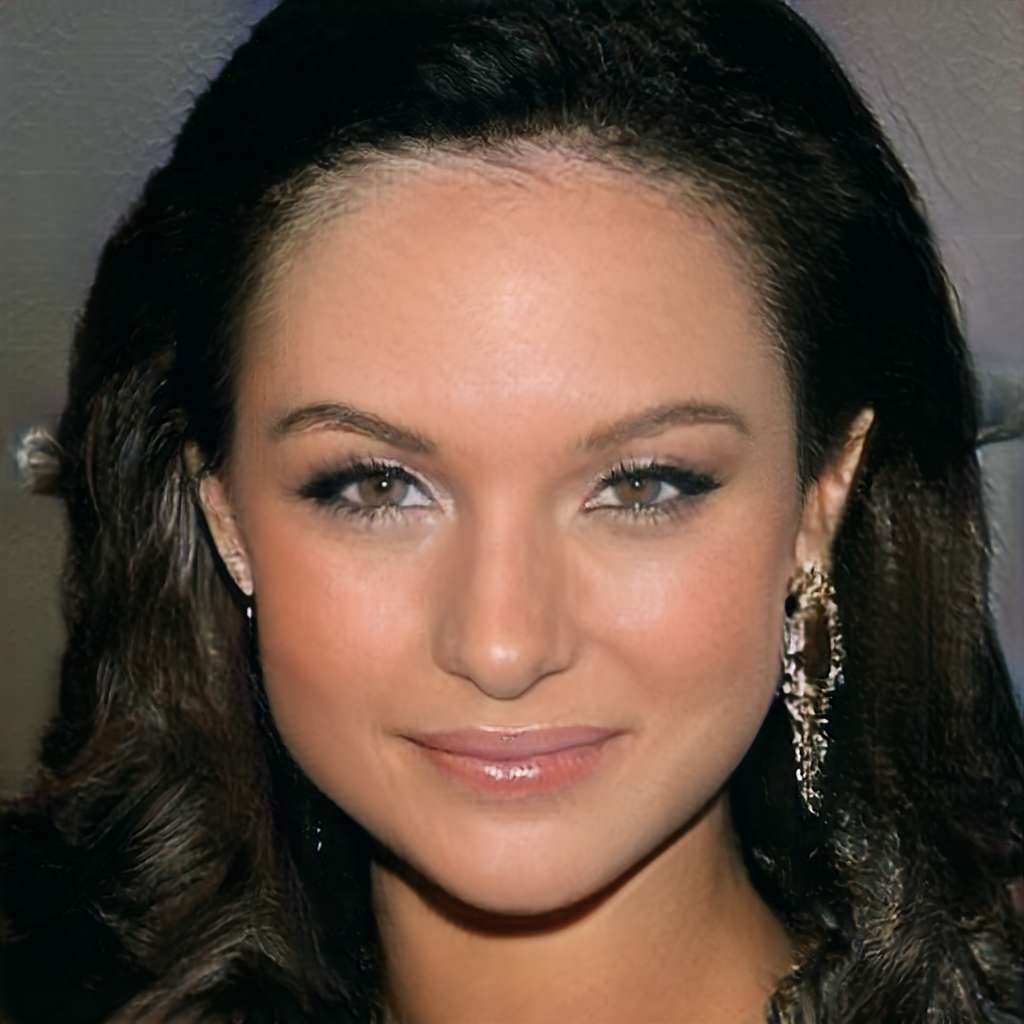} &
\includegraphics[width=\pganw]{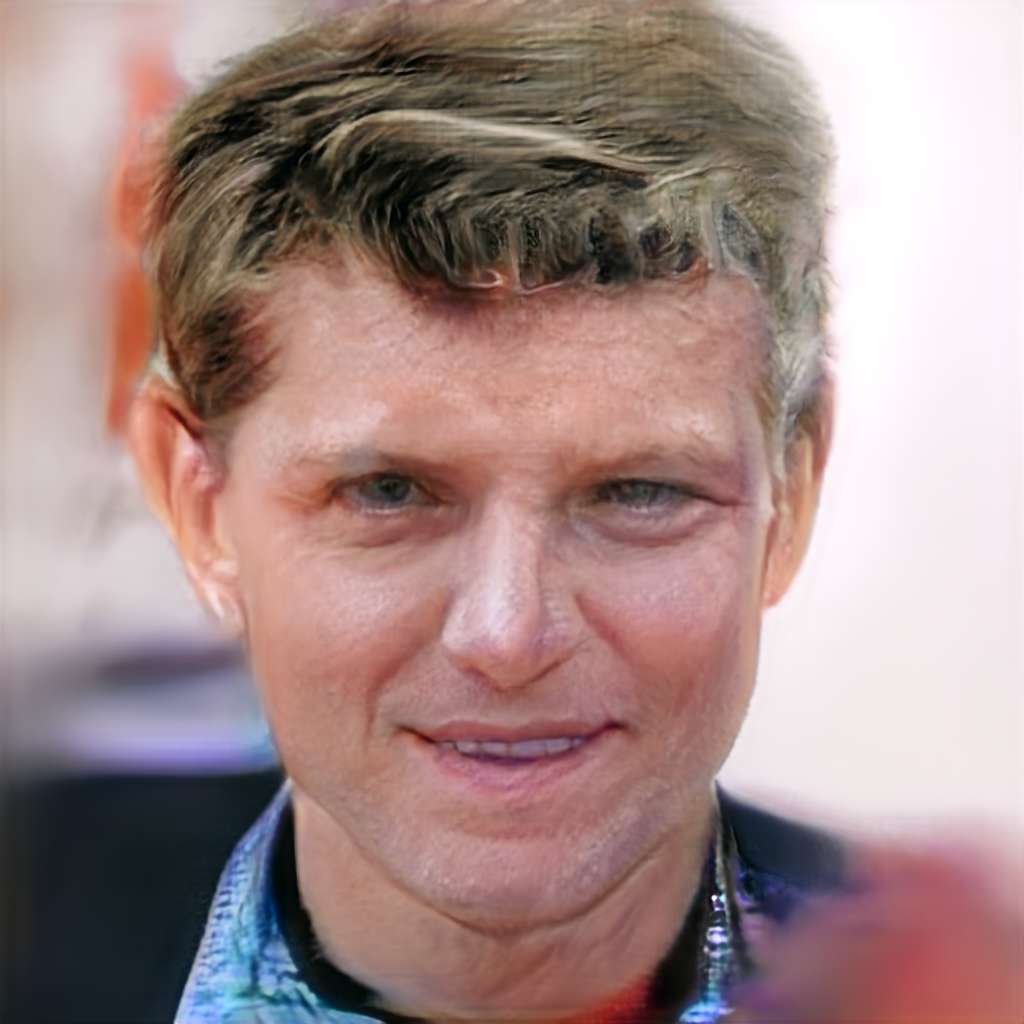} &
\unreal{\includegraphics[width=\pganw]{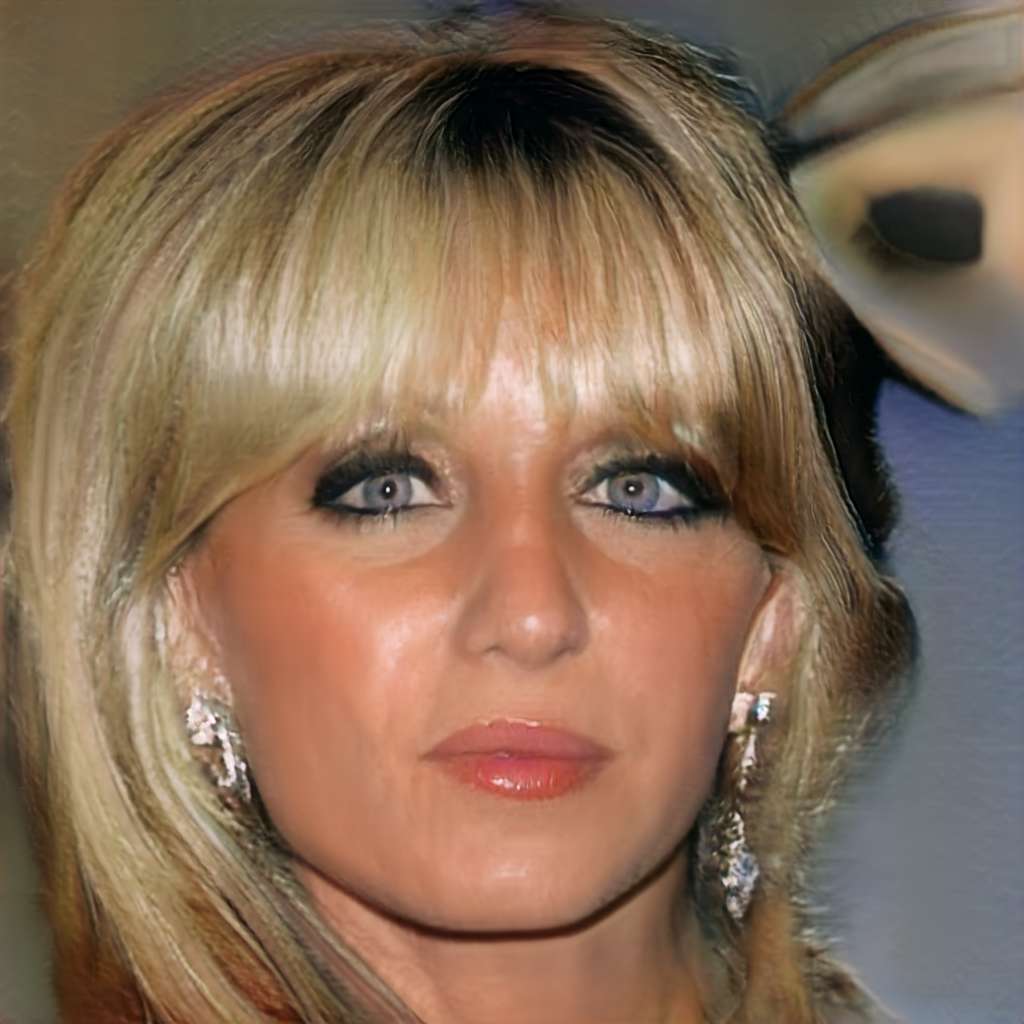}} &
\includegraphics[width=\pganw]{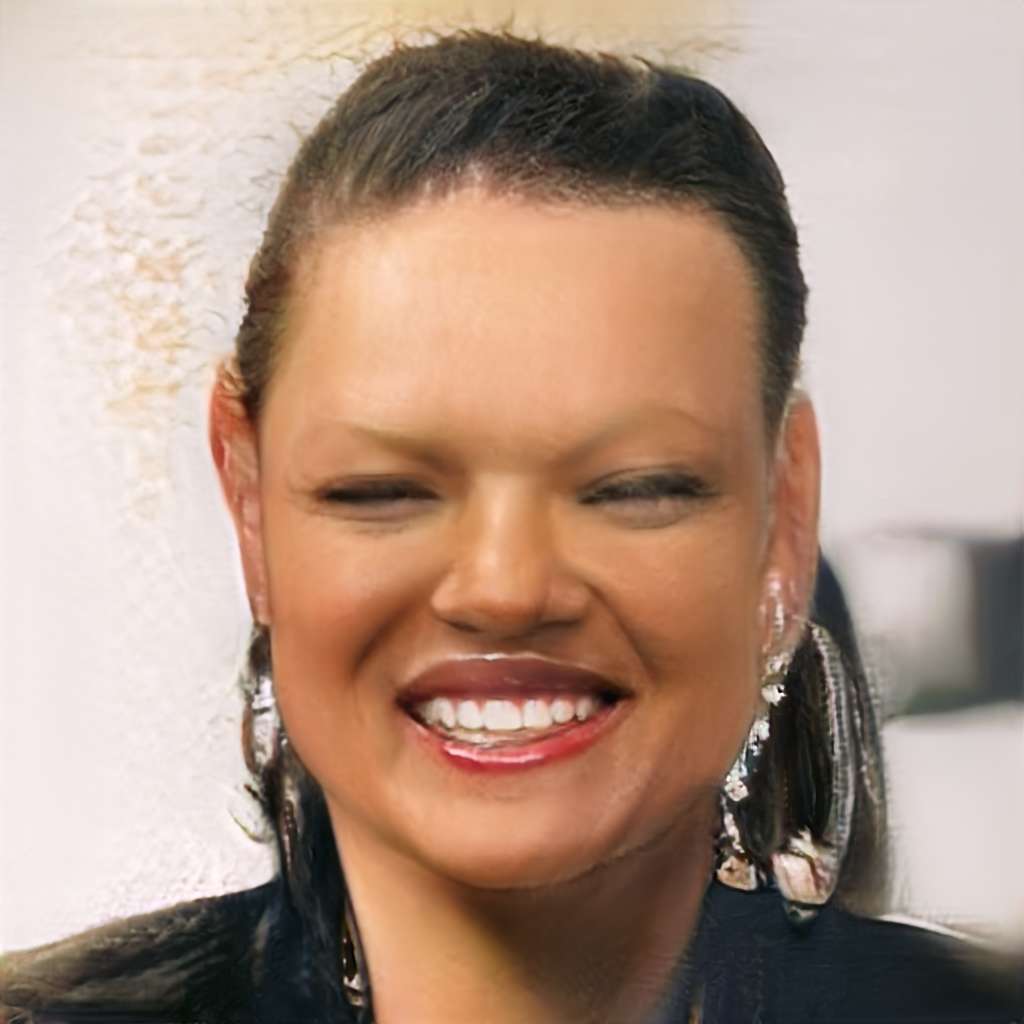} &
\includegraphics[width=\pganw]{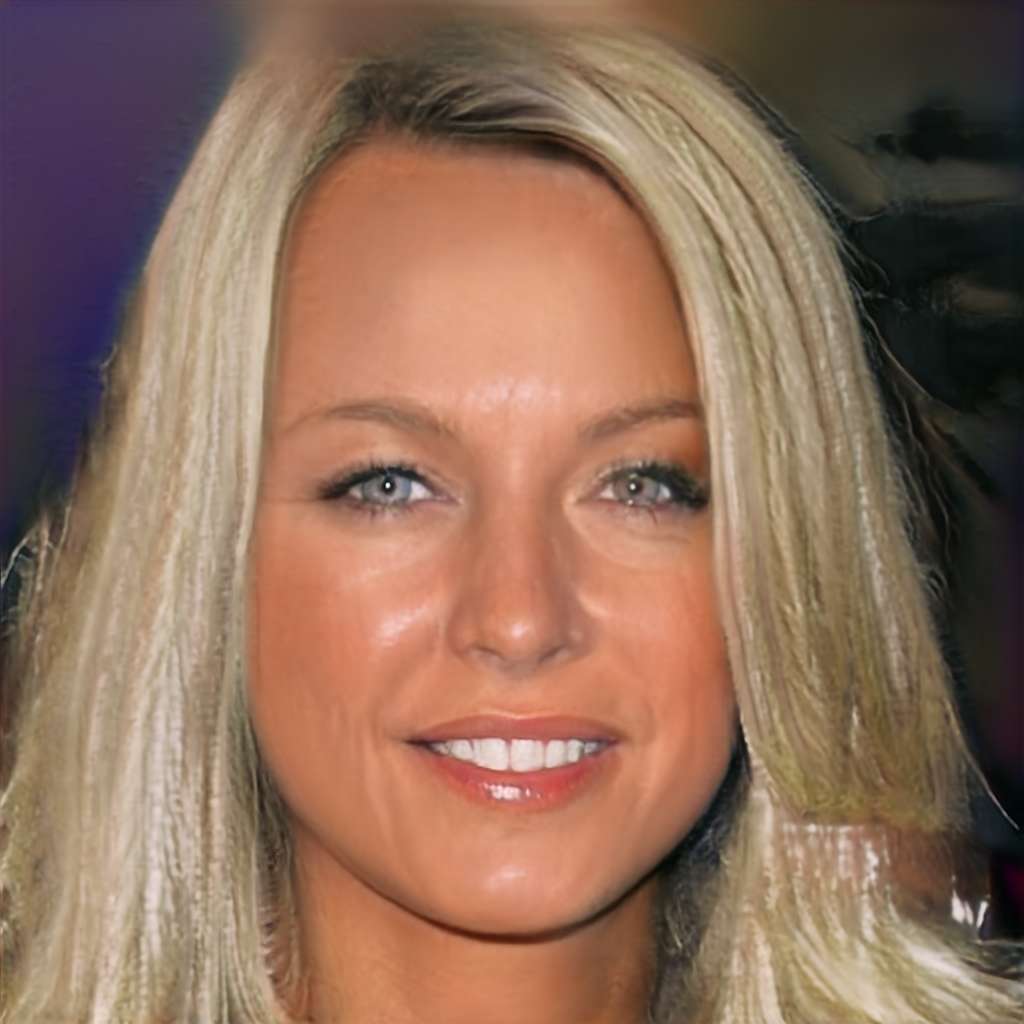} \\
\includegraphics[width=\pganw]{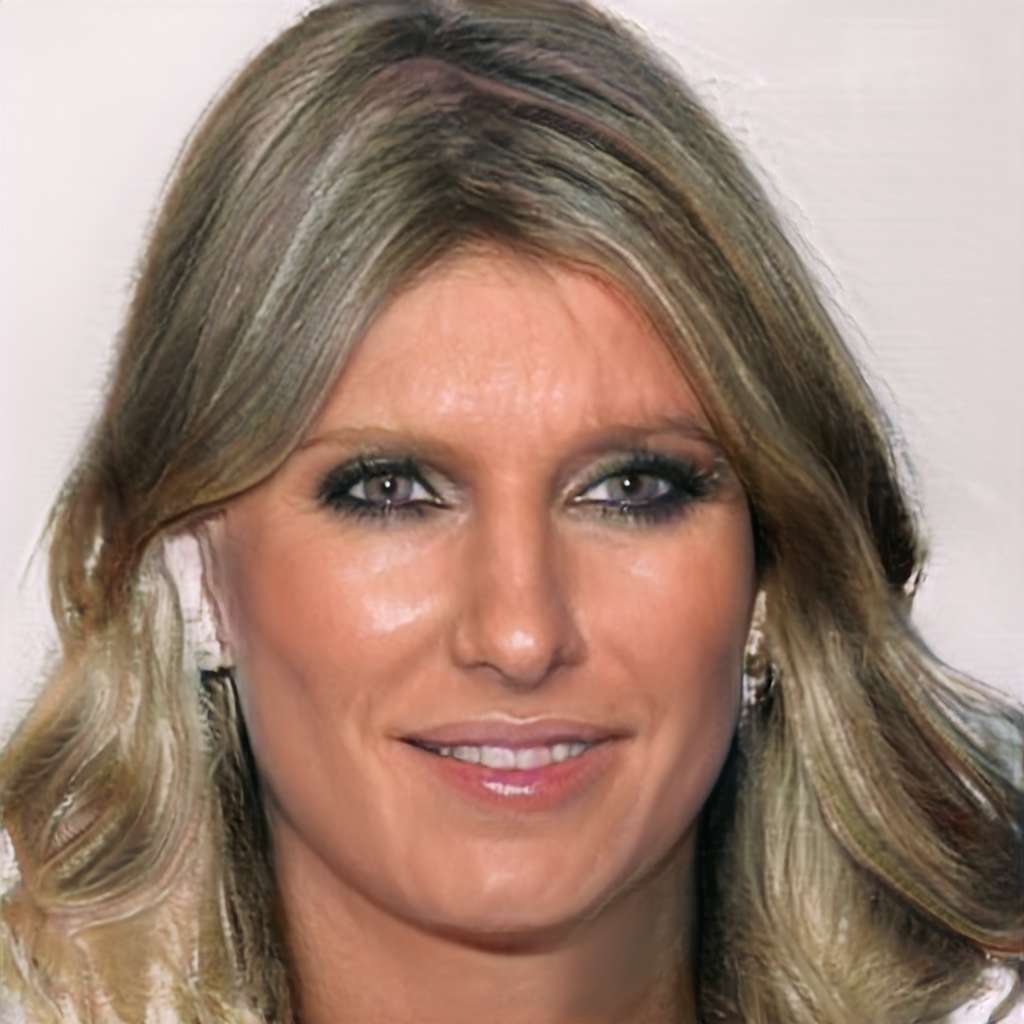} &
\includegraphics[width=\pganw]{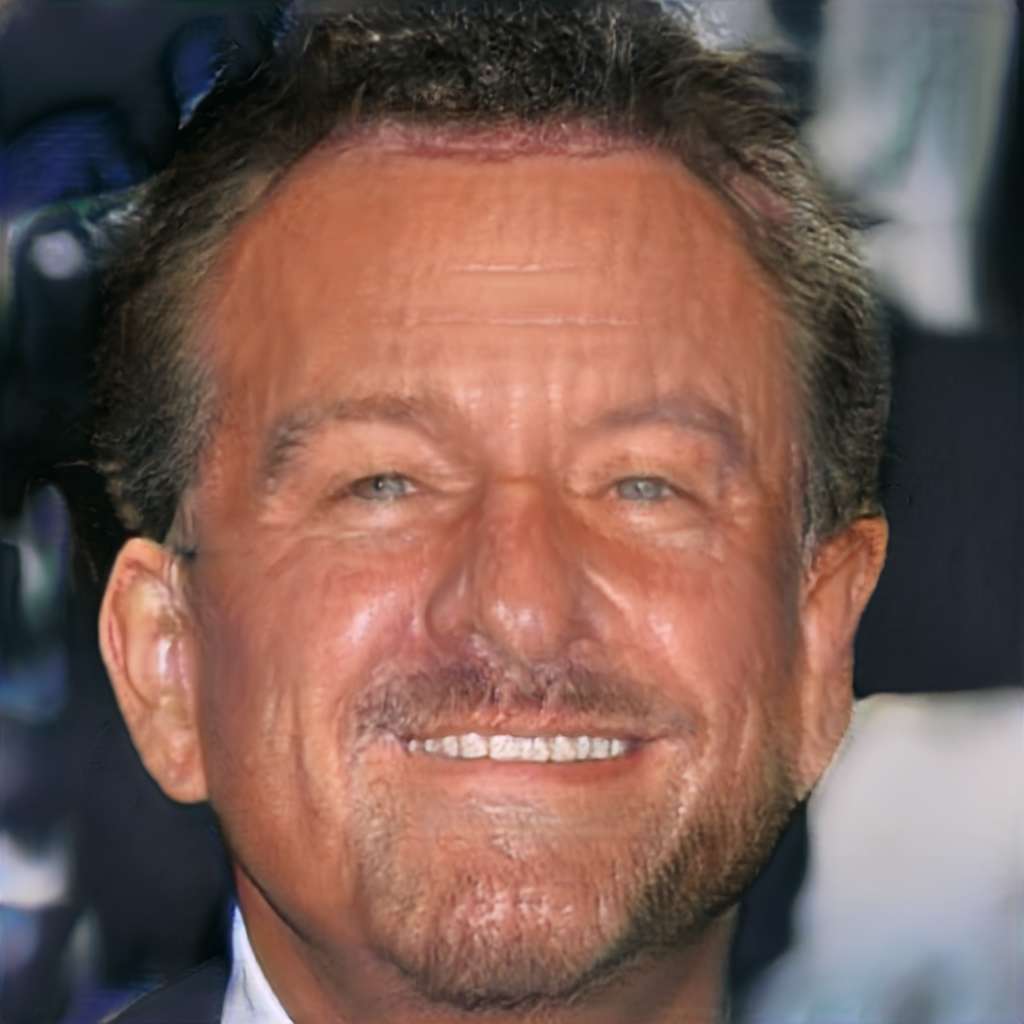} &
\unreal{\includegraphics[width=\pganw]{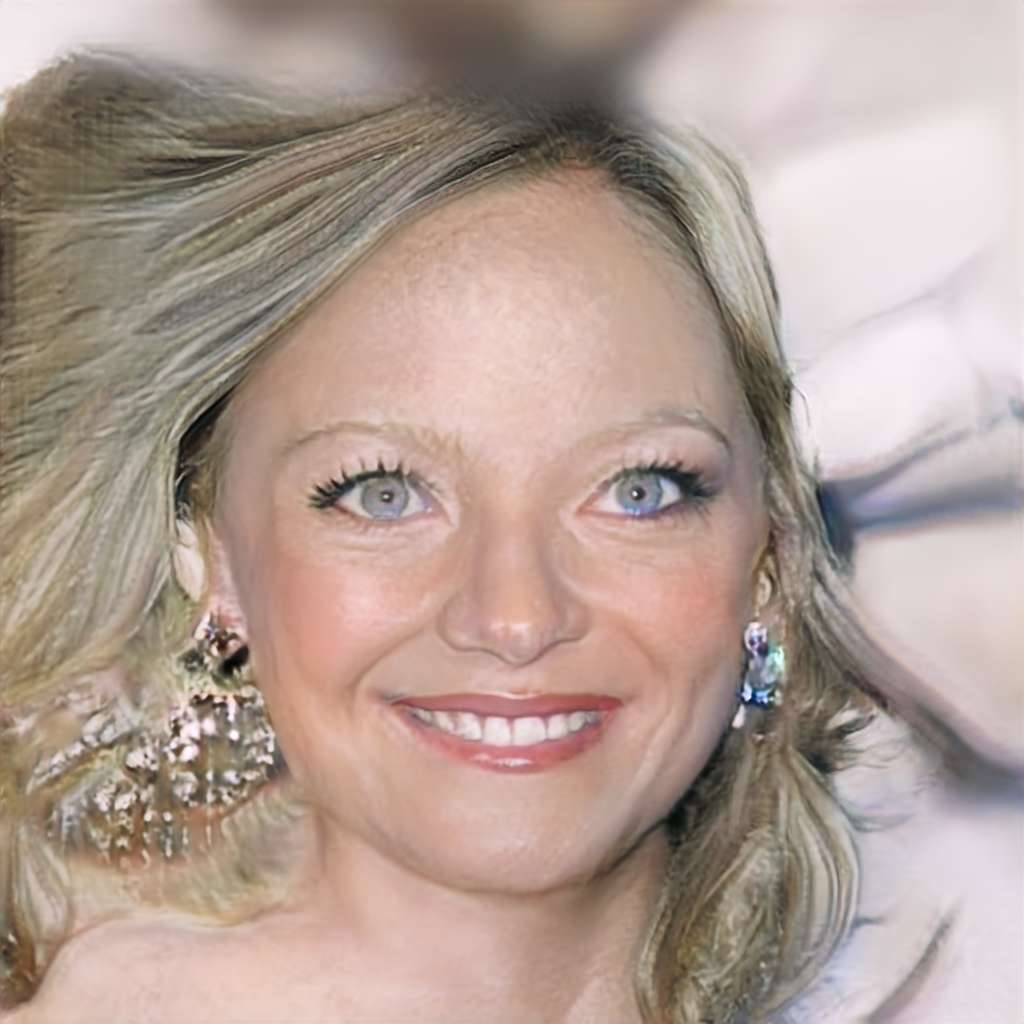}} &
\unreal{\includegraphics[width=\pganw]{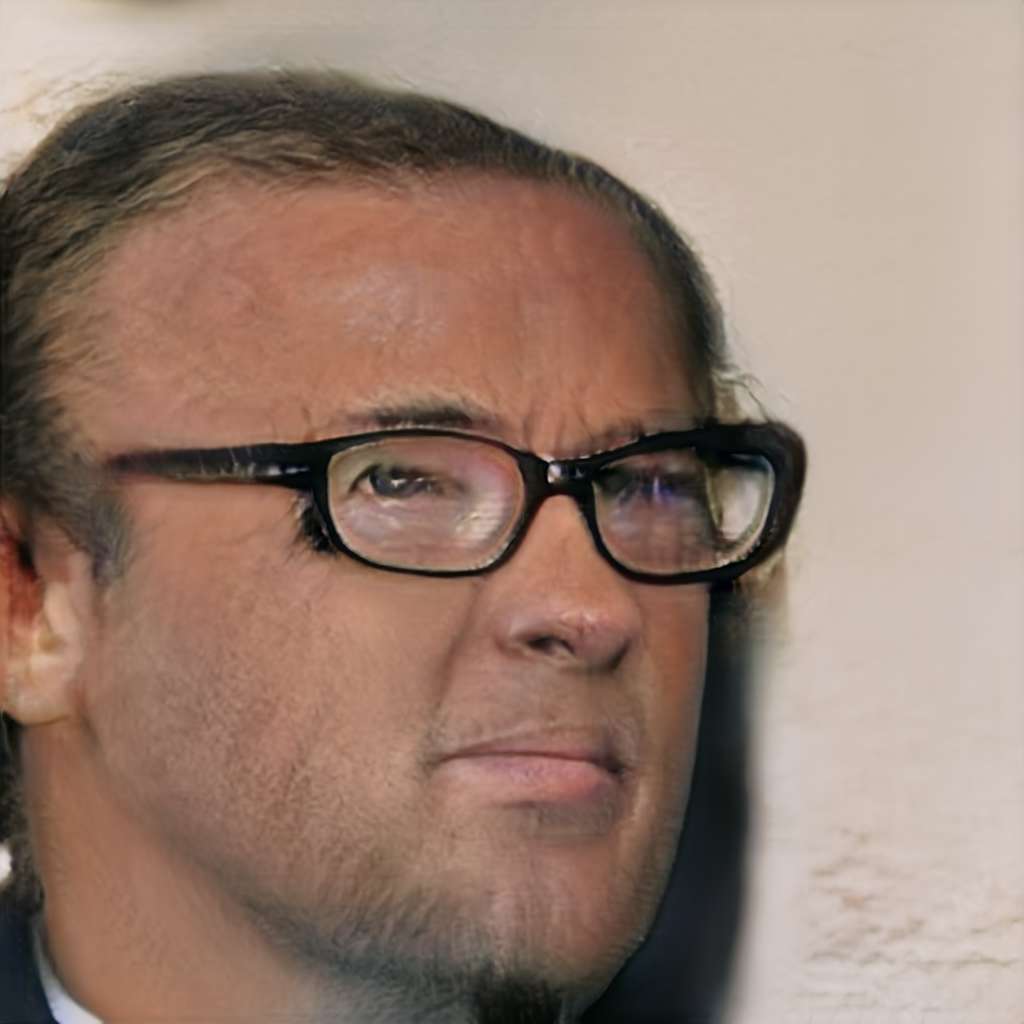}} &
\includegraphics[width=\pganw]{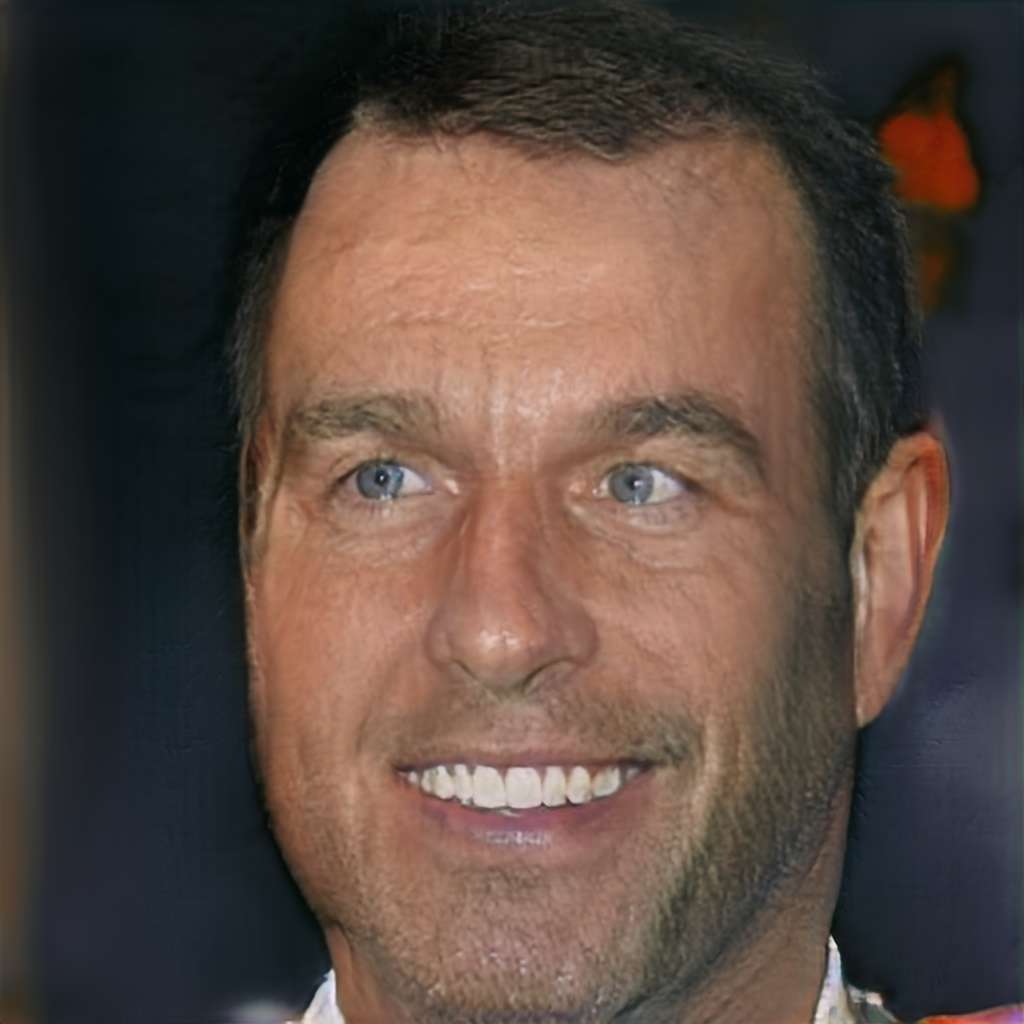} &
\includegraphics[width=\pganw]{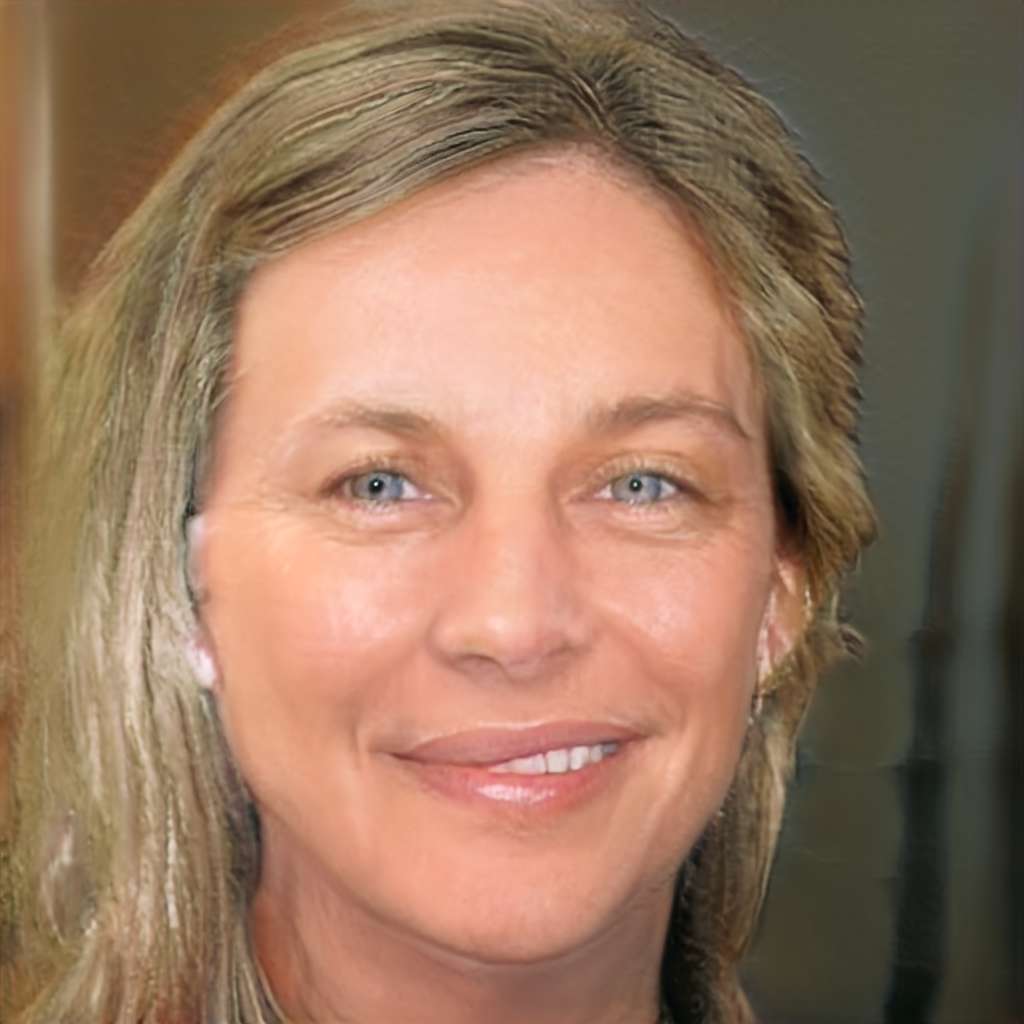} \\
\includegraphics[width=\pganw]{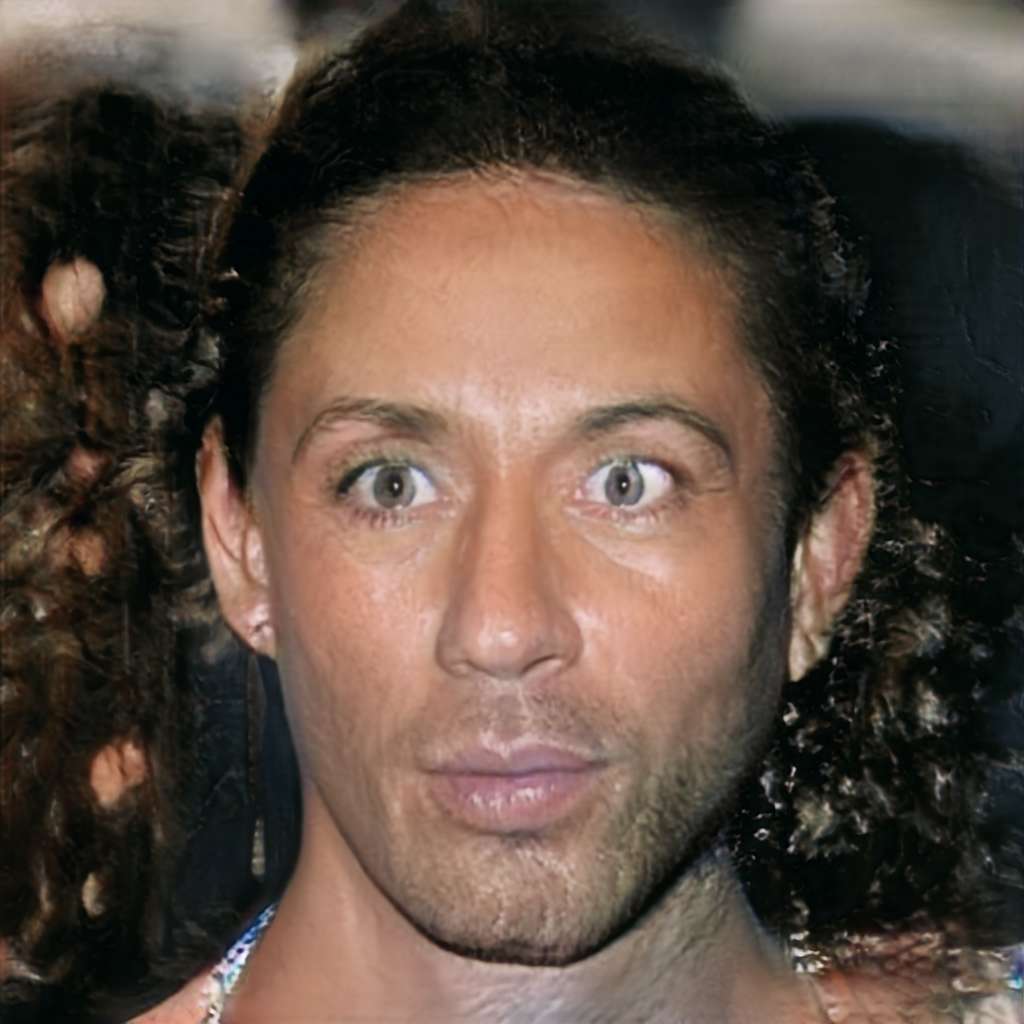} &
\includegraphics[width=\pganw]{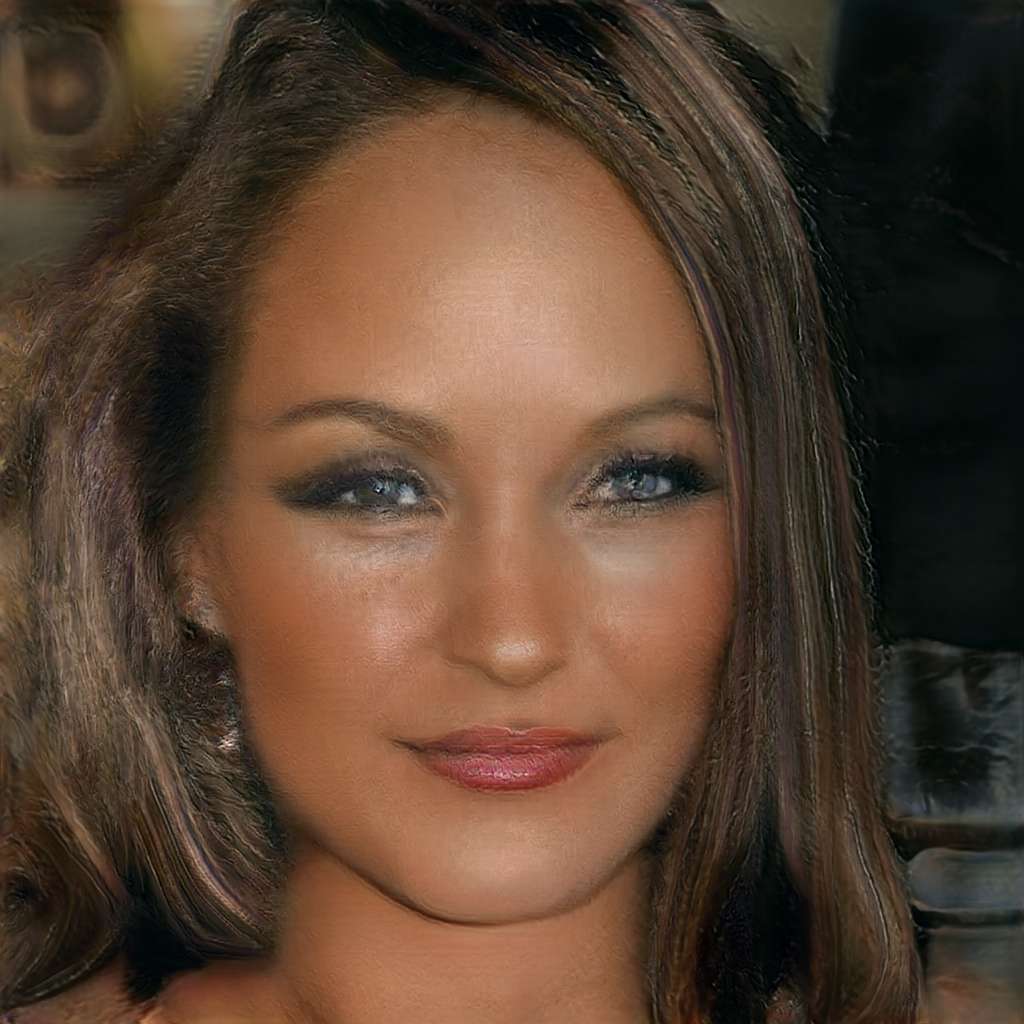} &
\includegraphics[width=\pganw]{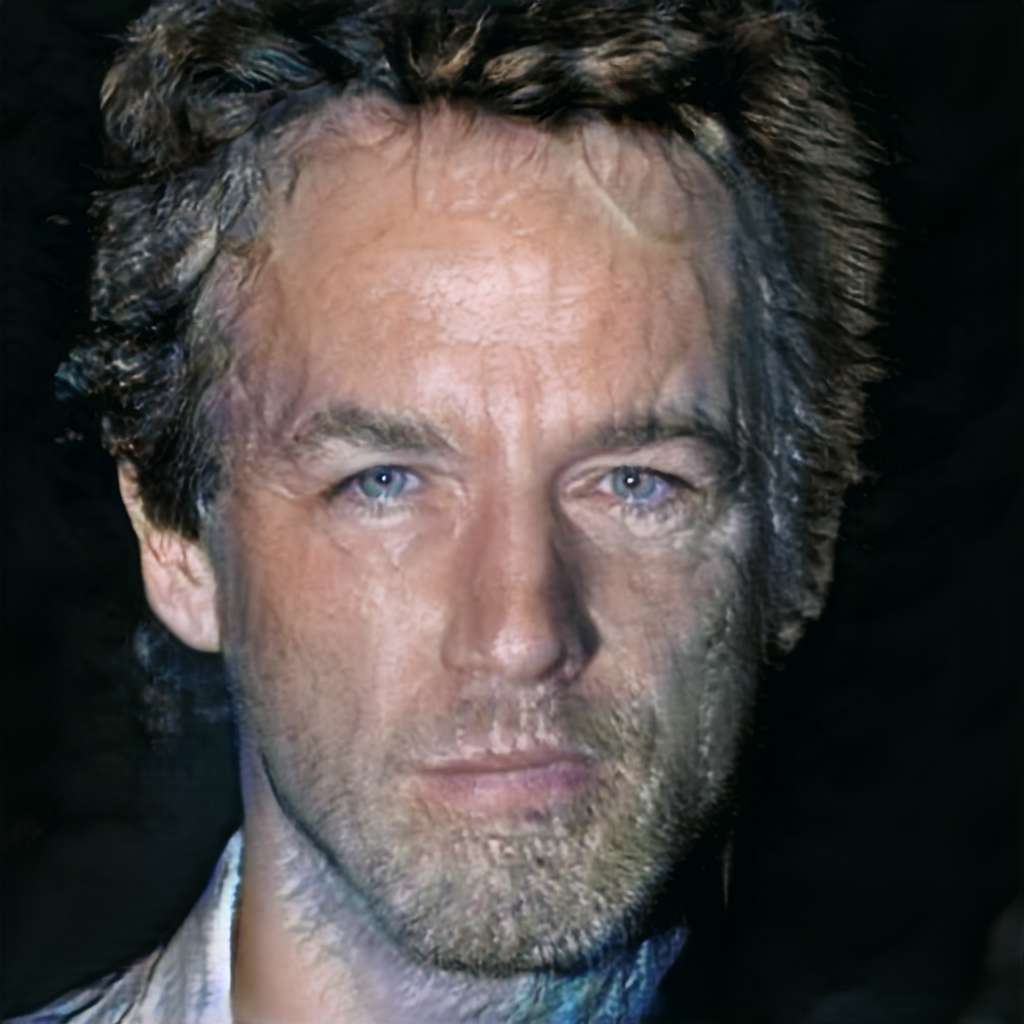} &
\includegraphics[width=\pganw]{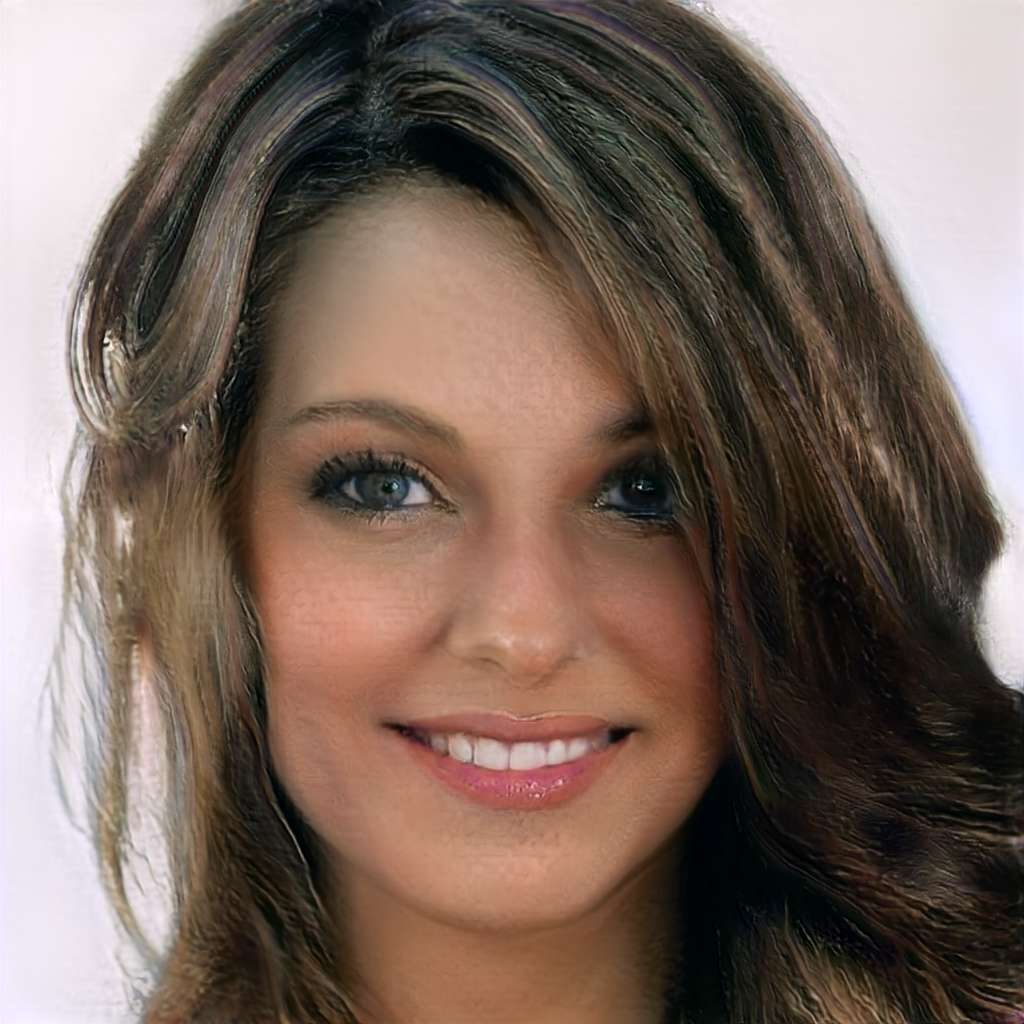} &
\includegraphics[width=\pganw]{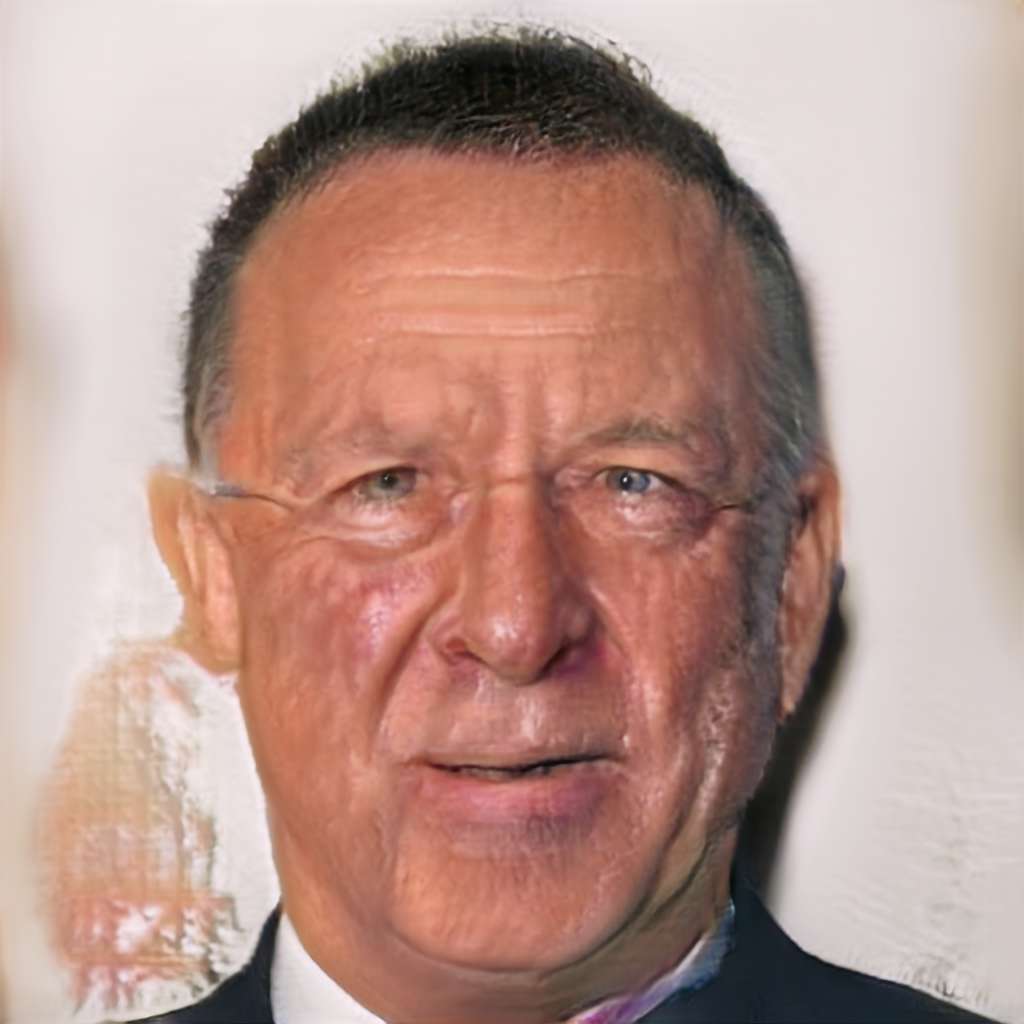} &
\includegraphics[width=\pganw]{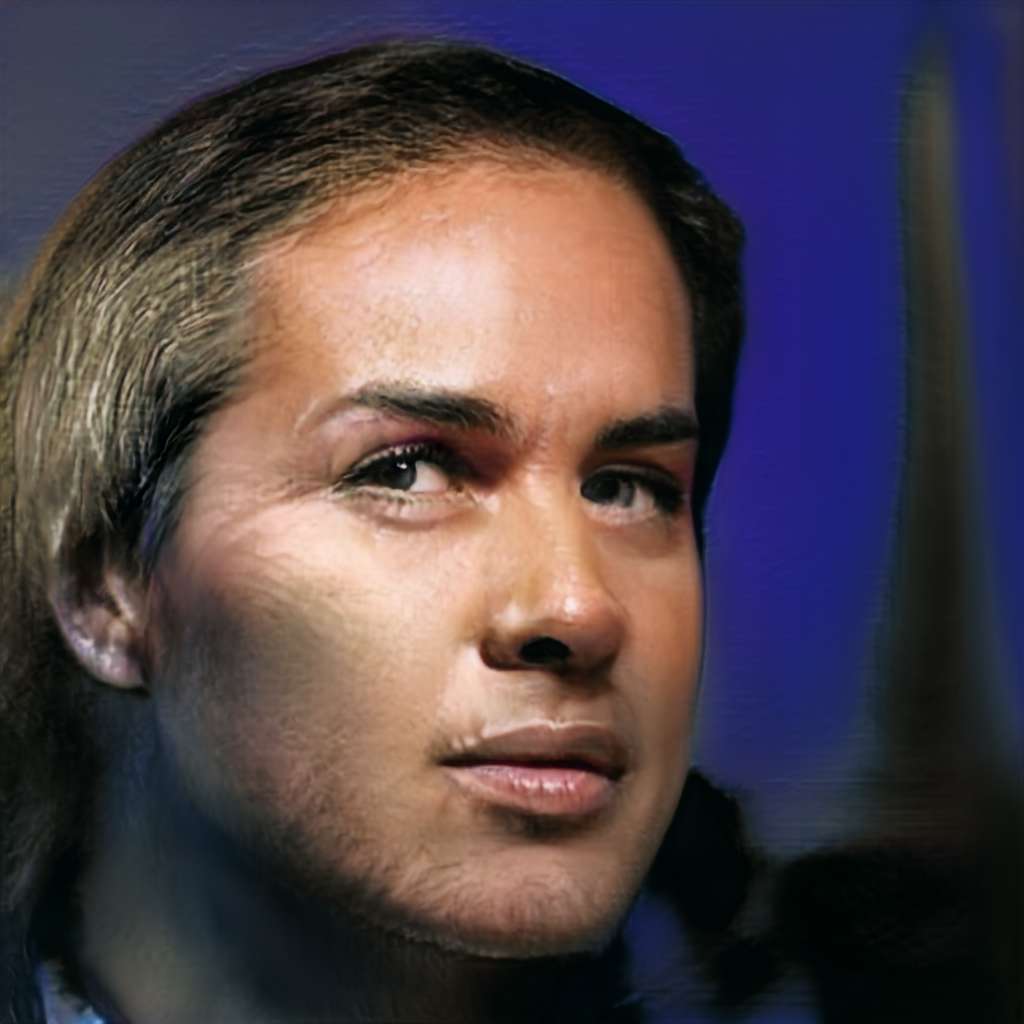} \\
\unreal{\includegraphics[width=\pganw]{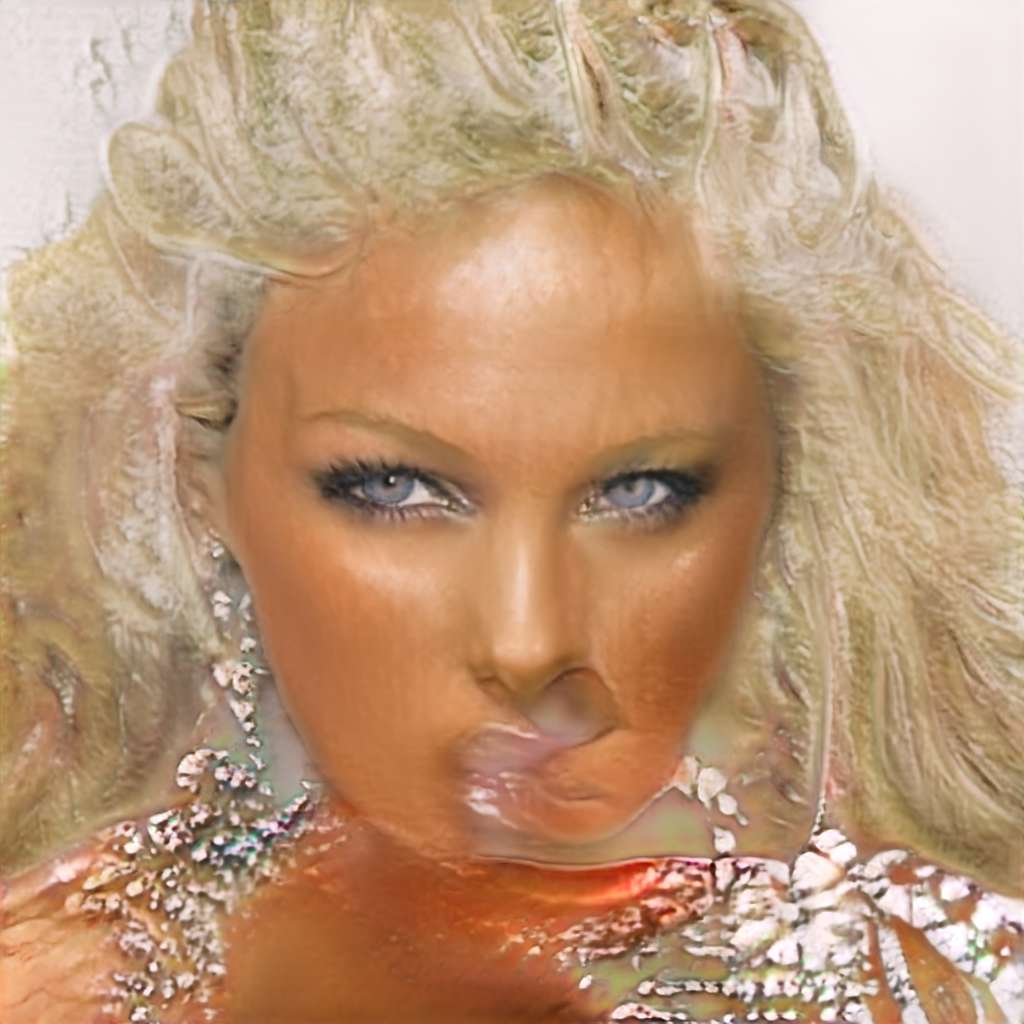}} &
\unreal{\includegraphics[width=\pganw]{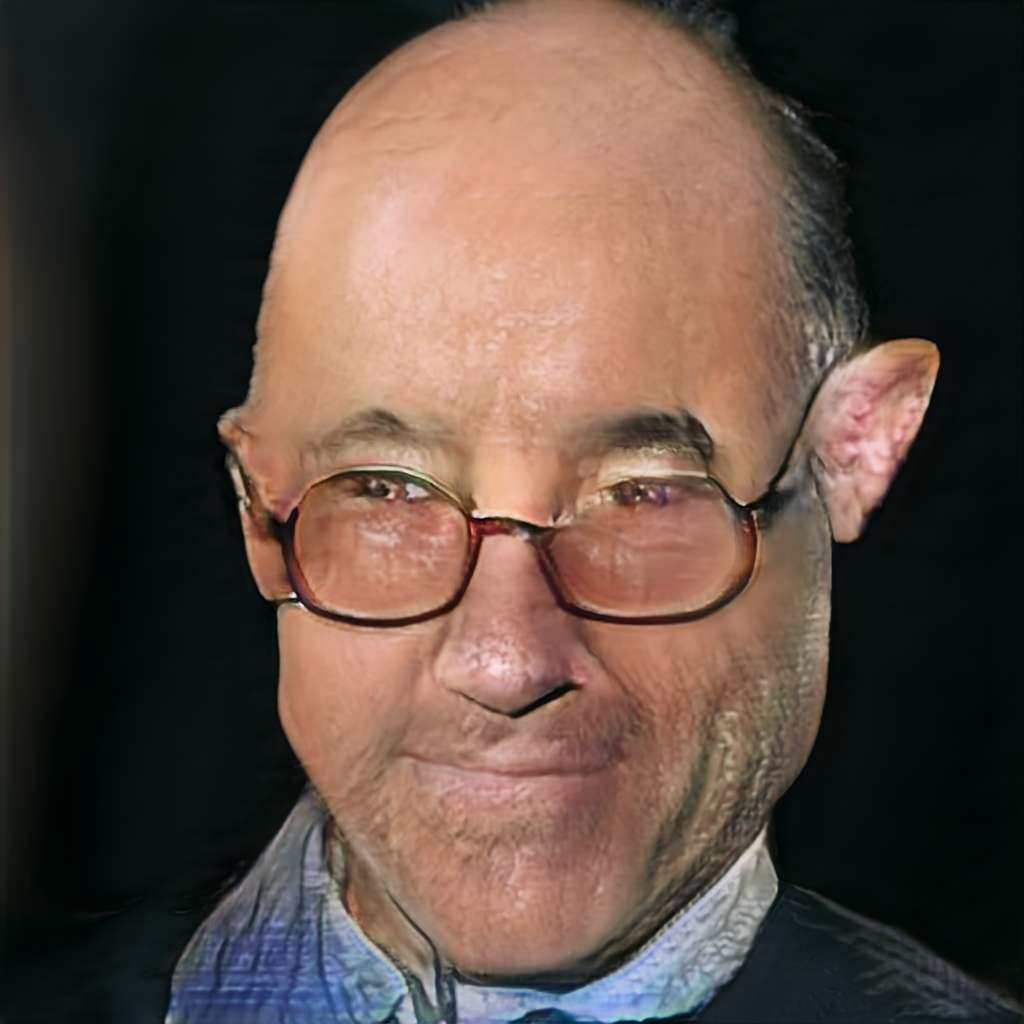}} &
\includegraphics[width=\pganw]{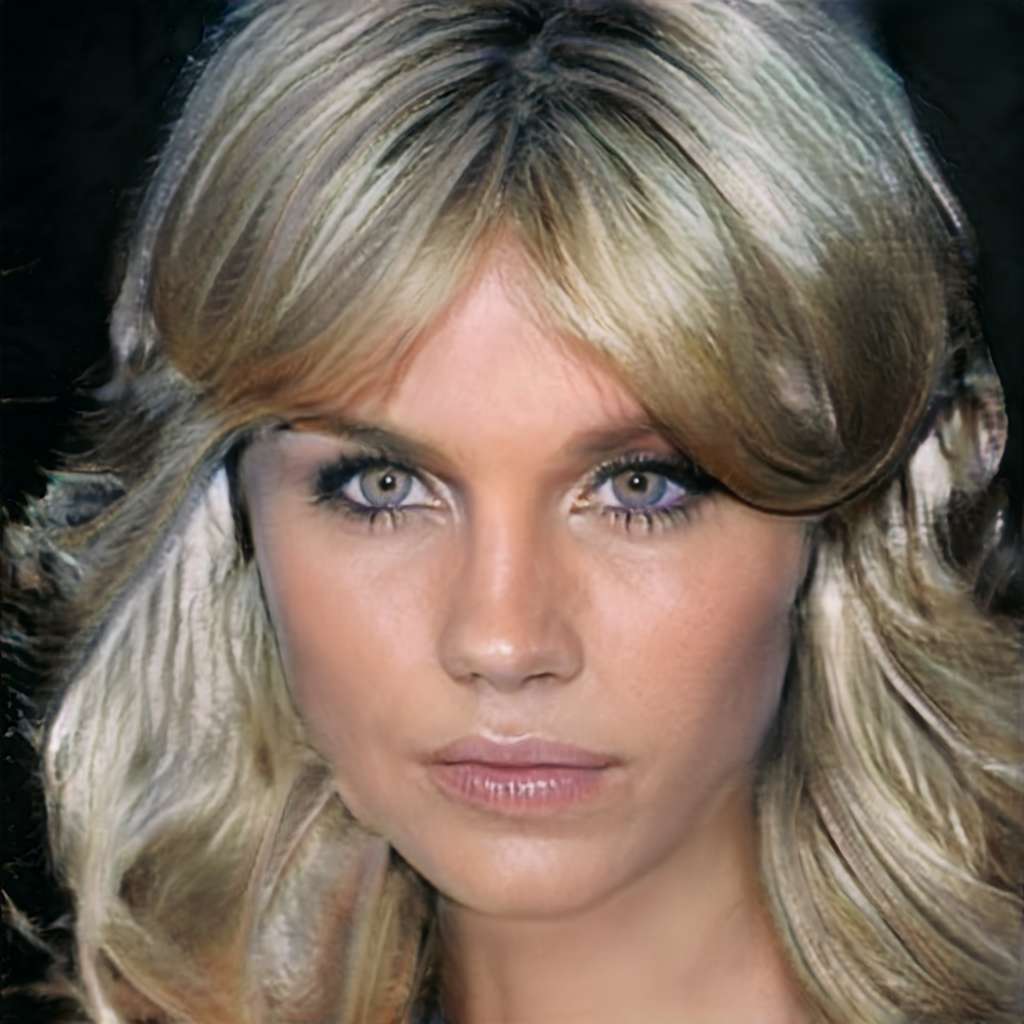} &
\includegraphics[width=\pganw]{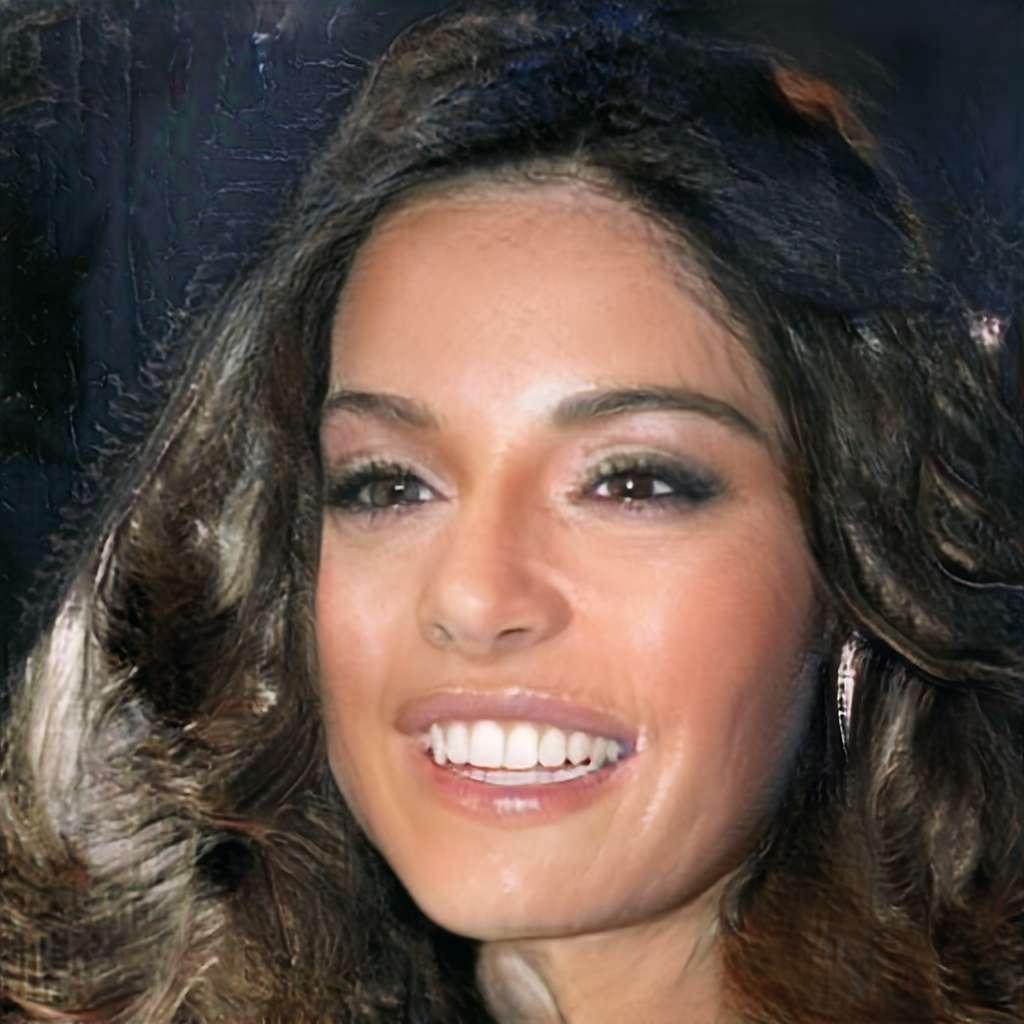} &
\includegraphics[width=\pganw]{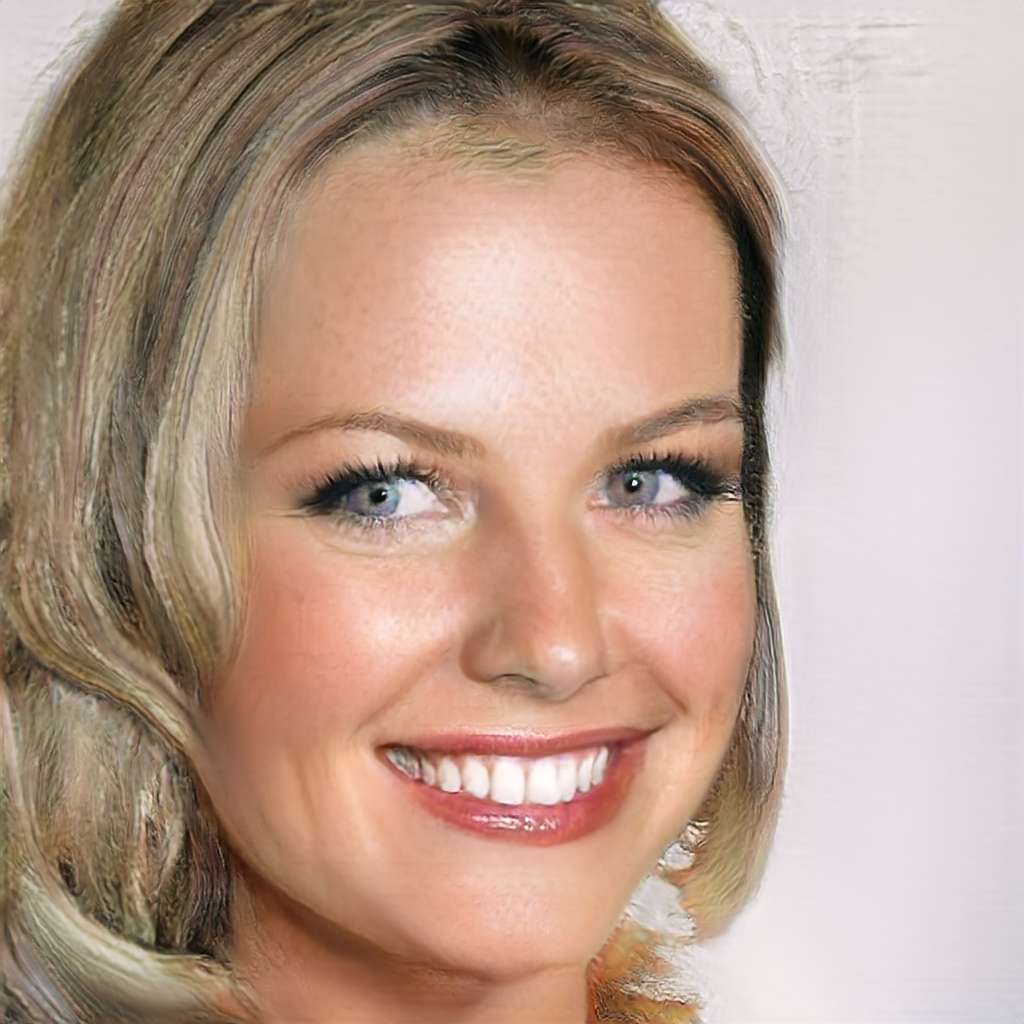} &
\unreal{\includegraphics[width=\pganw]{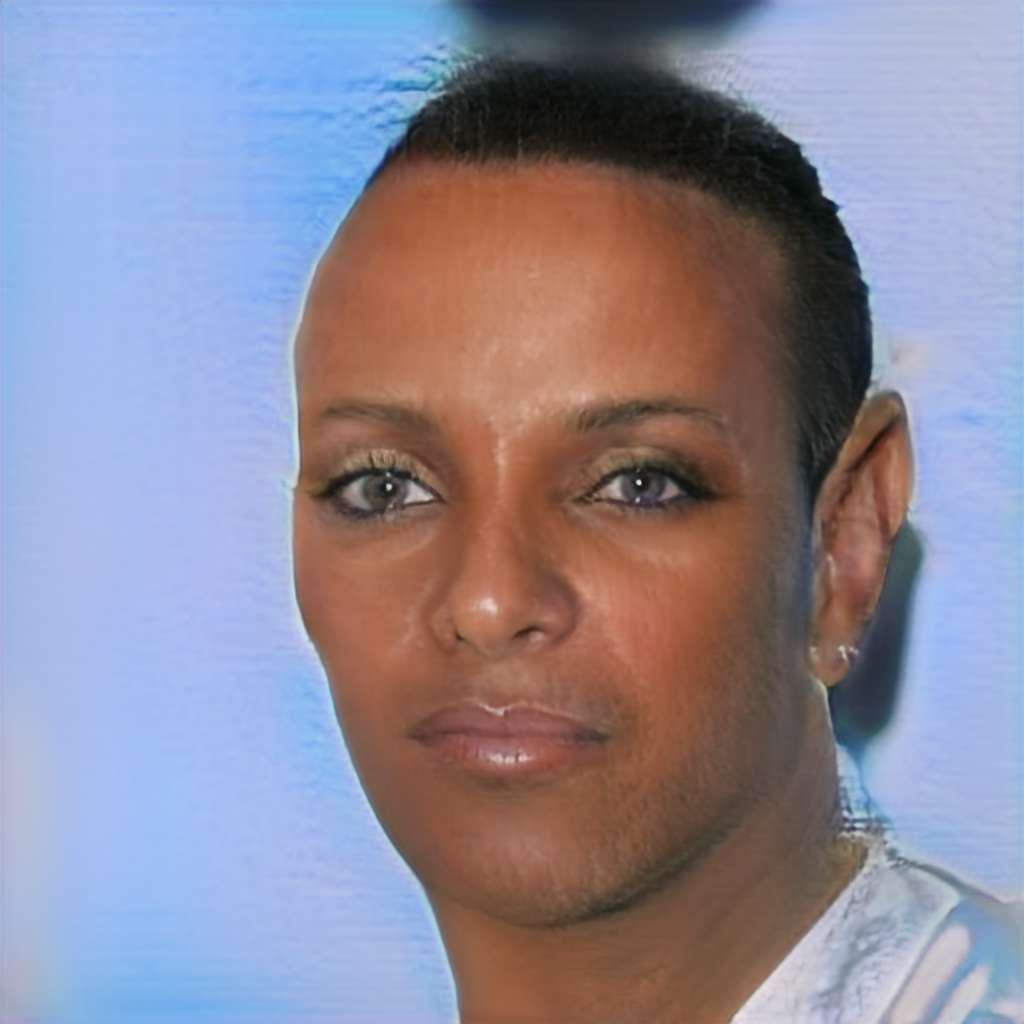}} \\
\end{tabular}
\end{center}
\end{table*}

\FloatBarrier

\section*{Acknowledgements}
We thank Rosanne Liu and Zoubin Ghahramani for useful discussions and comments.

\bibliography{mgan_refs}
\bibliographystyle{icml2019}

\FloatBarrier

\appendix

\section{Supplementary Material}
\label{sec:SI}

In this section we present some of the samples from the various GAN setups in full page figures below.

We also note that the GAN approach to density estimation is complementary to the earlier \emph{density ratio estimation} approach~\citep{Sugiyama2012}\@.
In density ratio estimation, the generator $G$ is fixed, and the density is found by combining Bayes' rule and the learned classifier $D$.
In GANs, the key is learning $G$ well; while in density ratio estimation, the key is learning $D$ well.
The MH-GAN has flavors of both in that it uses both $G$ and $D$ to build $G'$.

\begin{figure*}[htbp]
    \centering
    \includegraphics[width=\exfactor\textwidth]{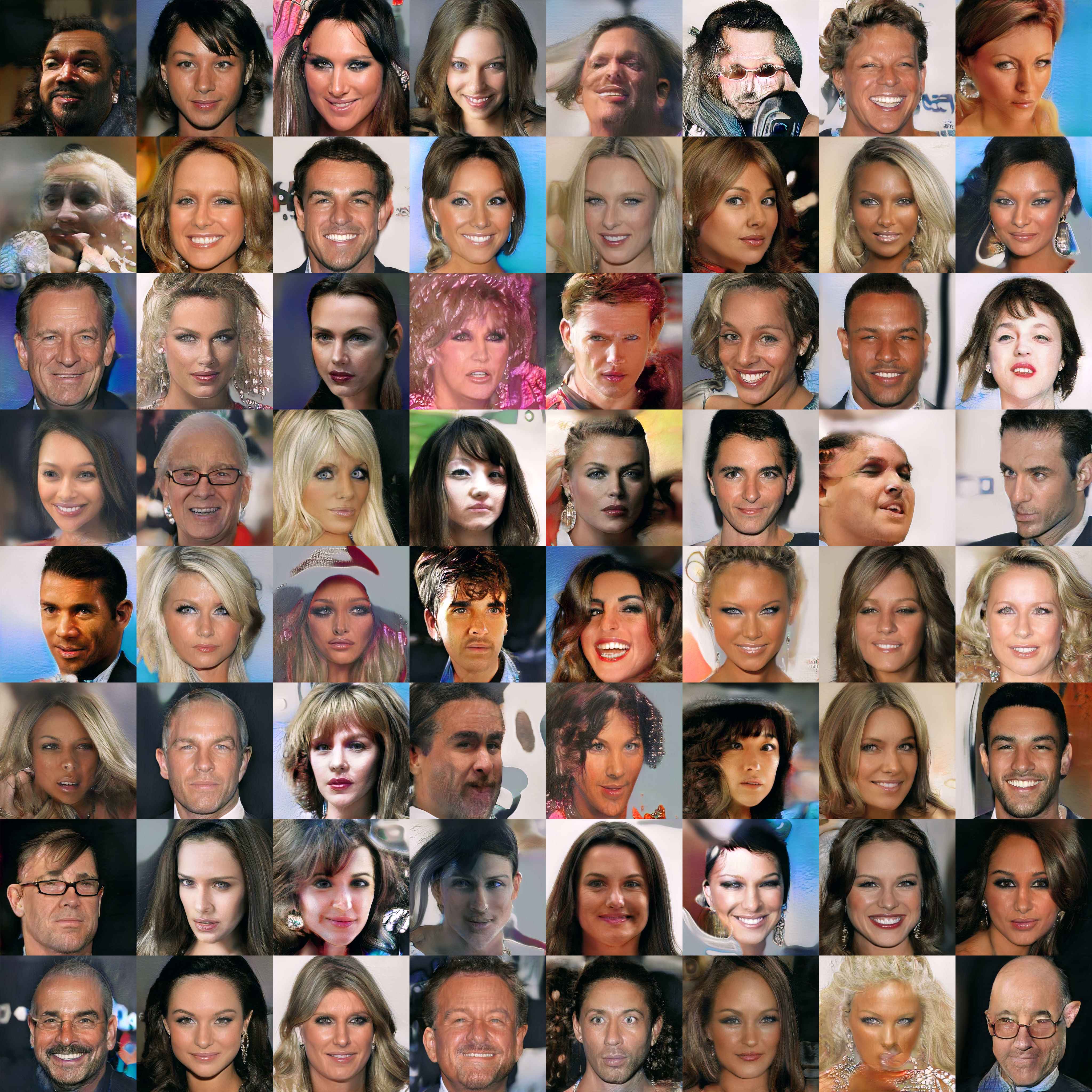}
    \caption{
    Set of 64 random samples from the (base) PGAN\@.
    }
    \label{fig:PGAN samples}
\end{figure*}

\begin{figure*}[htbp]
    \centering
       \includegraphics[width=0.2\textwidth]{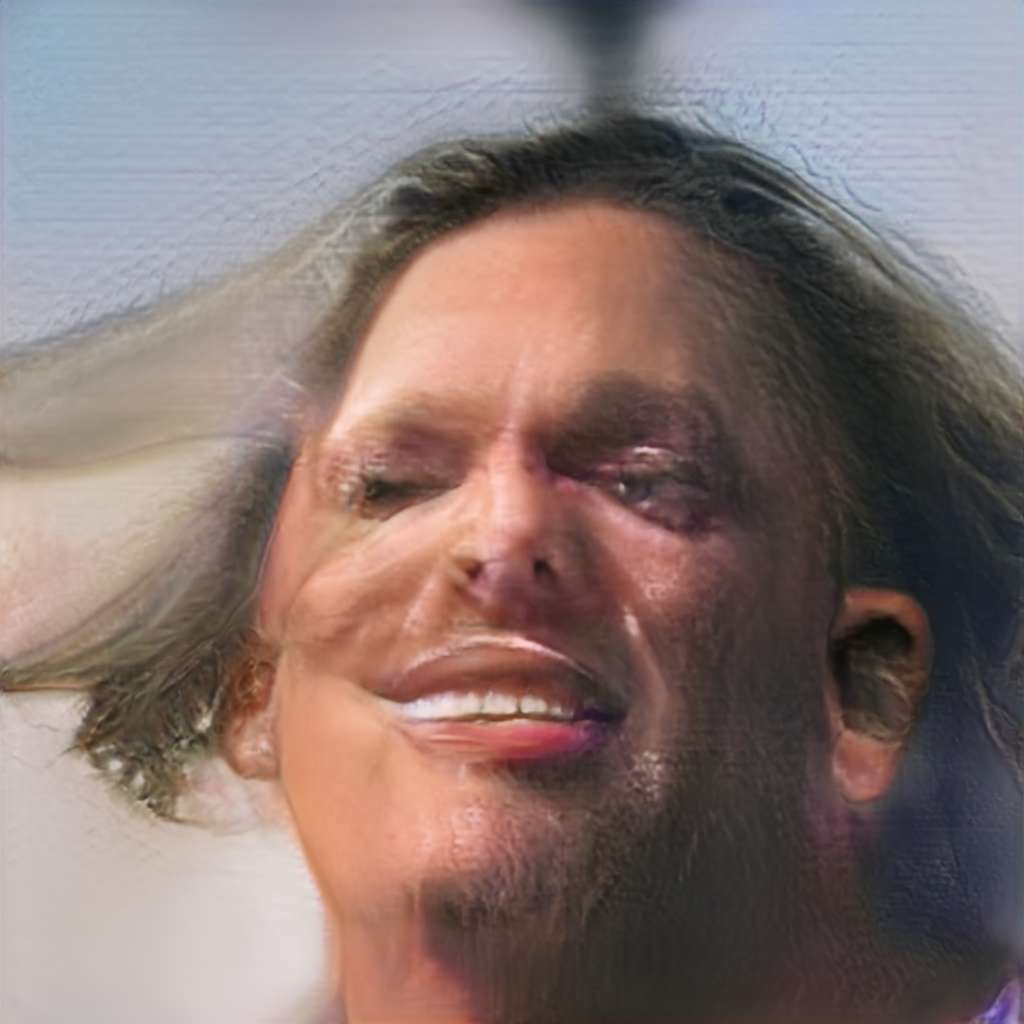}
    \hfill
       \includegraphics[width=0.2\textwidth]{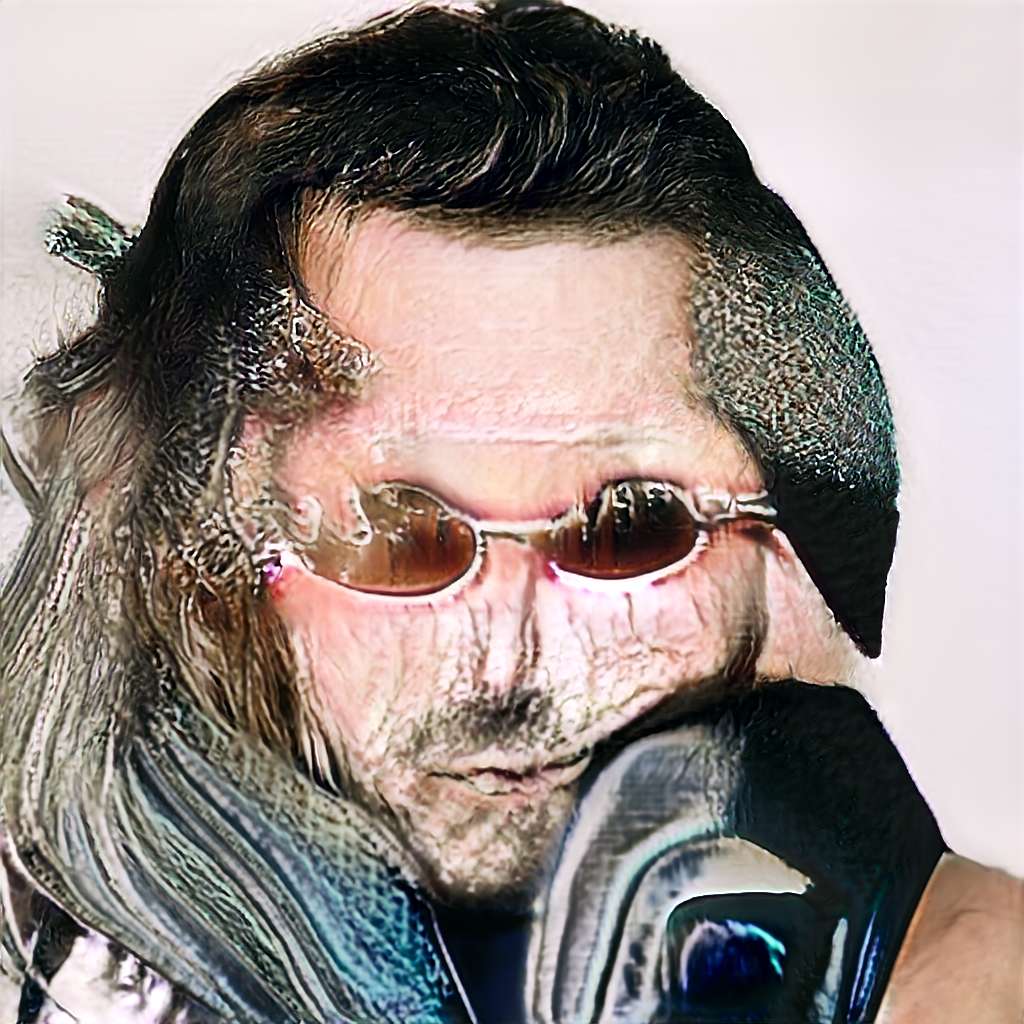}
    \hfill
       \includegraphics[width=0.2\textwidth]{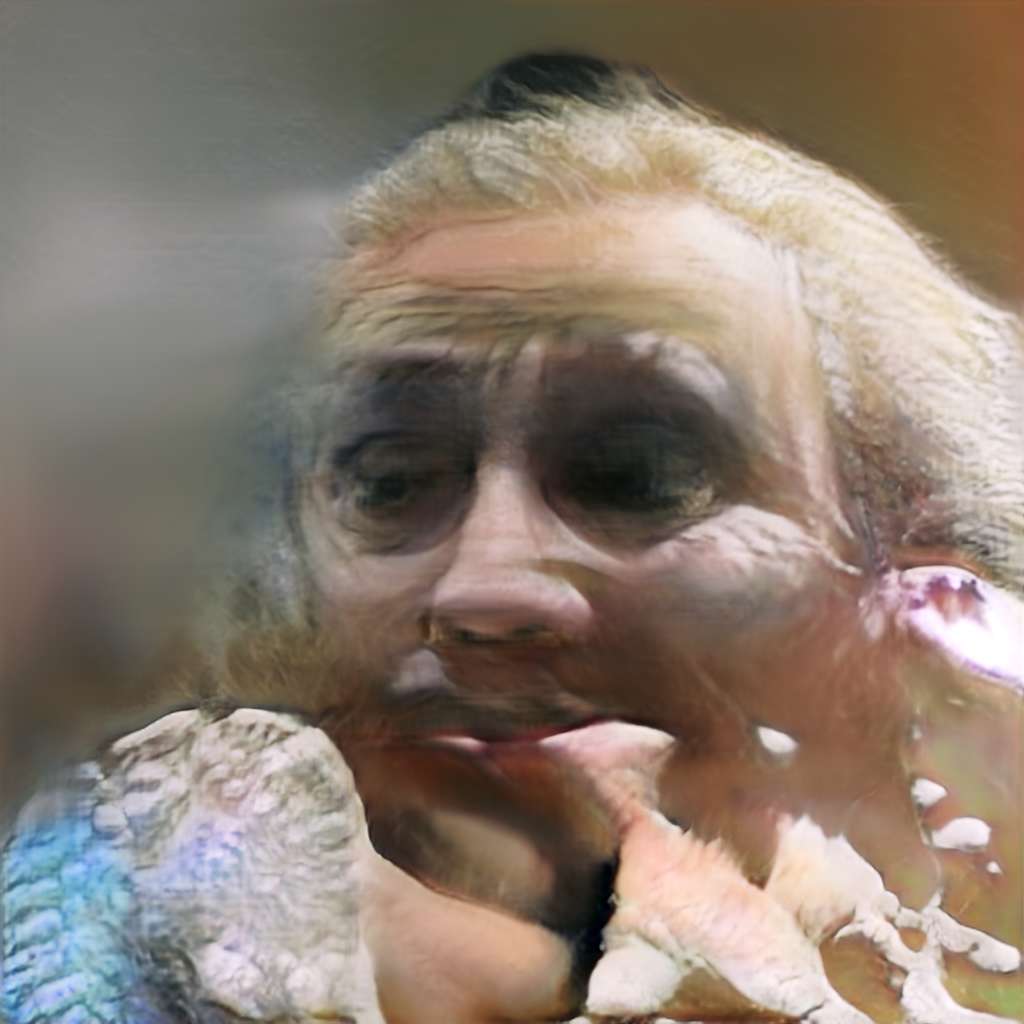}
    \hfill
       \includegraphics[width=0.2\textwidth]{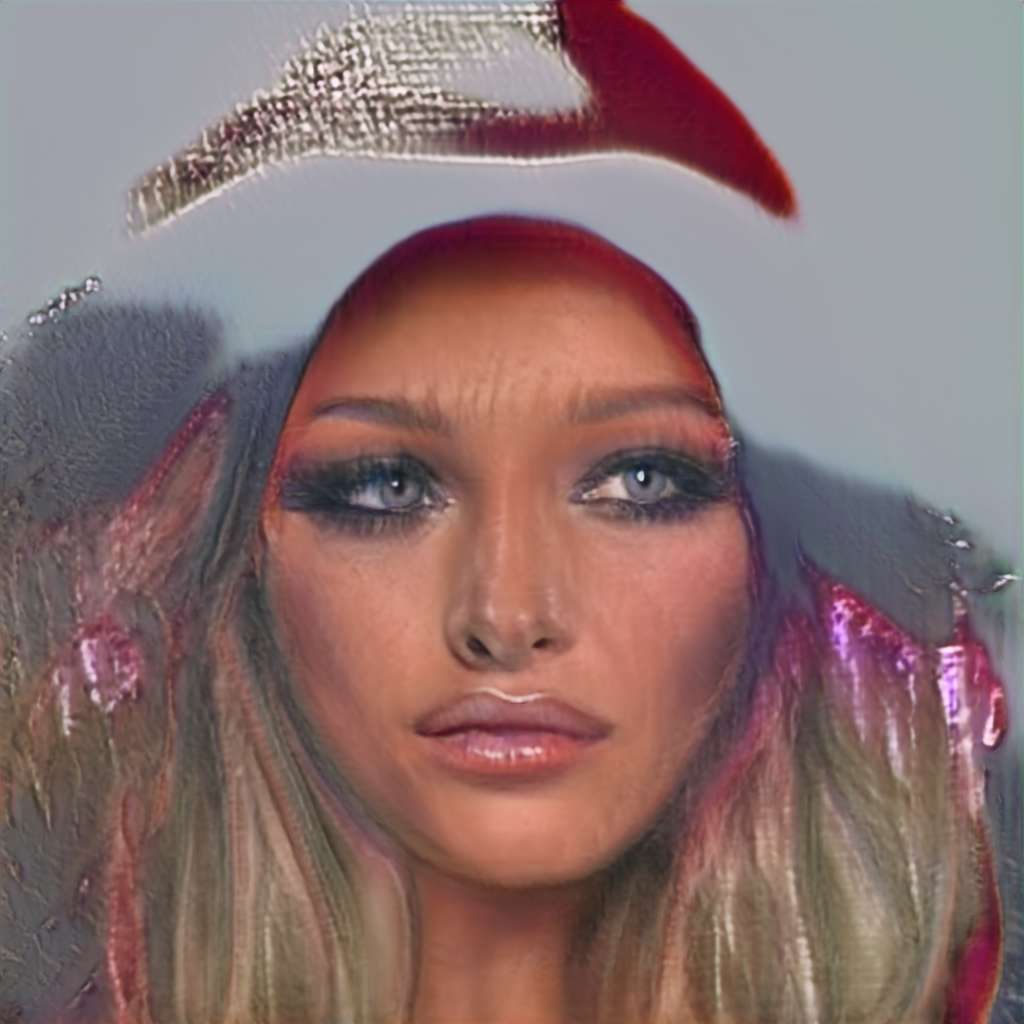}
    \caption{
    Flawed examples from the PGAN\@.
    Such a degree of non-realism is rarer in the DRS samples (Figure~\ref{fig:DRS 64x}) and nearly absent in the MH-GAN samples (Figure~\ref{fig:MHGAN 64x})\@.
    }
\end{figure*}

\begin{figure*}[htbp]
    \centering
    \includegraphics[width=\exfactor\textwidth]{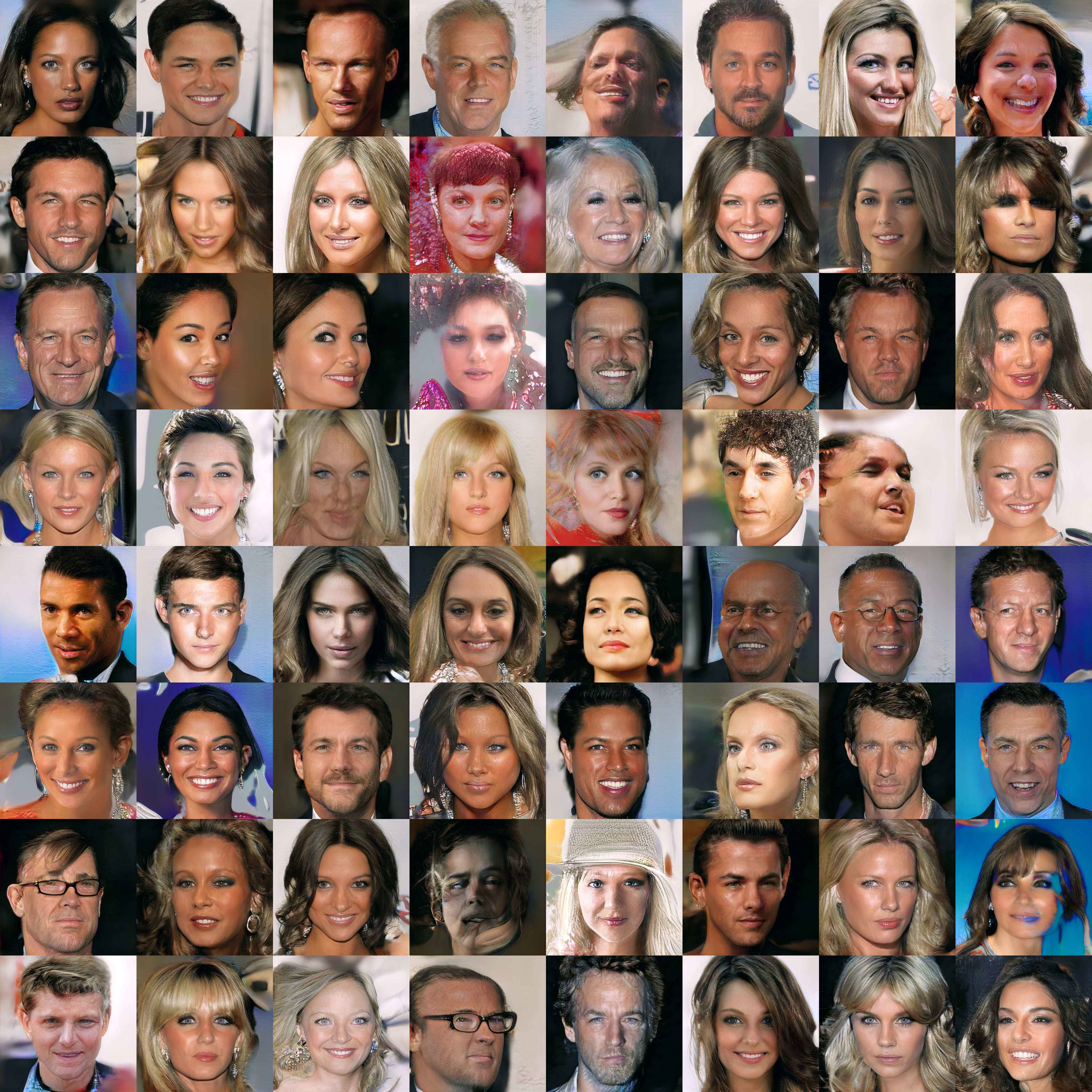}
    \caption{
    Set of 64 random samples from the calibrated DRS setup.
    }
    \label{fig:DRS 64x}
\end{figure*}

\begin{figure*}[htbp]
    \centering
    \includegraphics[width=\exfactor\textwidth]{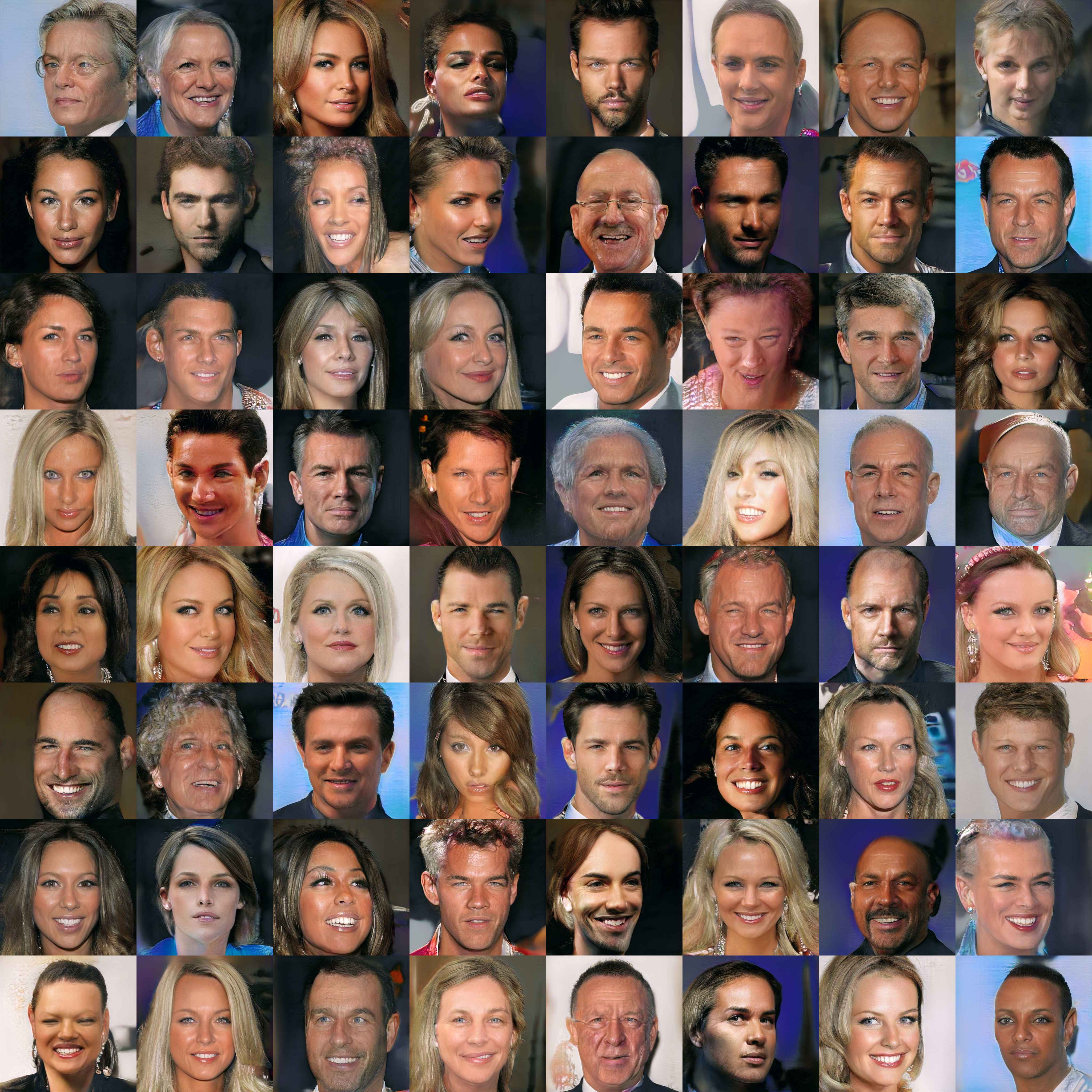}
    \caption{
    Set of 64 random samples from the calibrated MH-GAN\@.
    }
    \label{fig:MHGAN 64x}
\end{figure*}

\begin{figure*}[htbp]
    \centering
    \begin{subfigure}[b]{0.49\textwidth}
       \centering
       \includegraphics[width=\exfactor\textwidth]{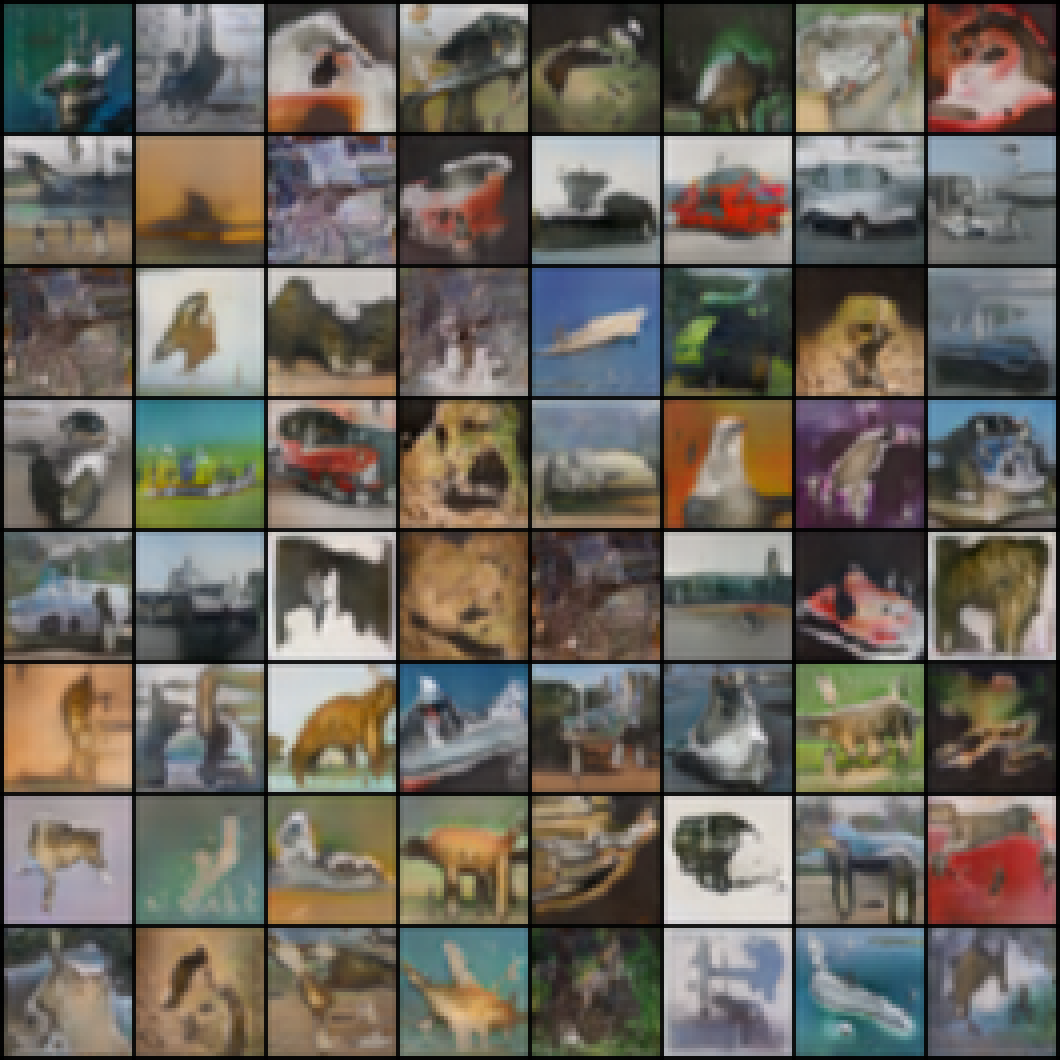}
       \caption{GAN}
    \end{subfigure}
    \begin{subfigure}[b]{0.49\textwidth}
       \centering
       \includegraphics[width=\exfactor\textwidth]{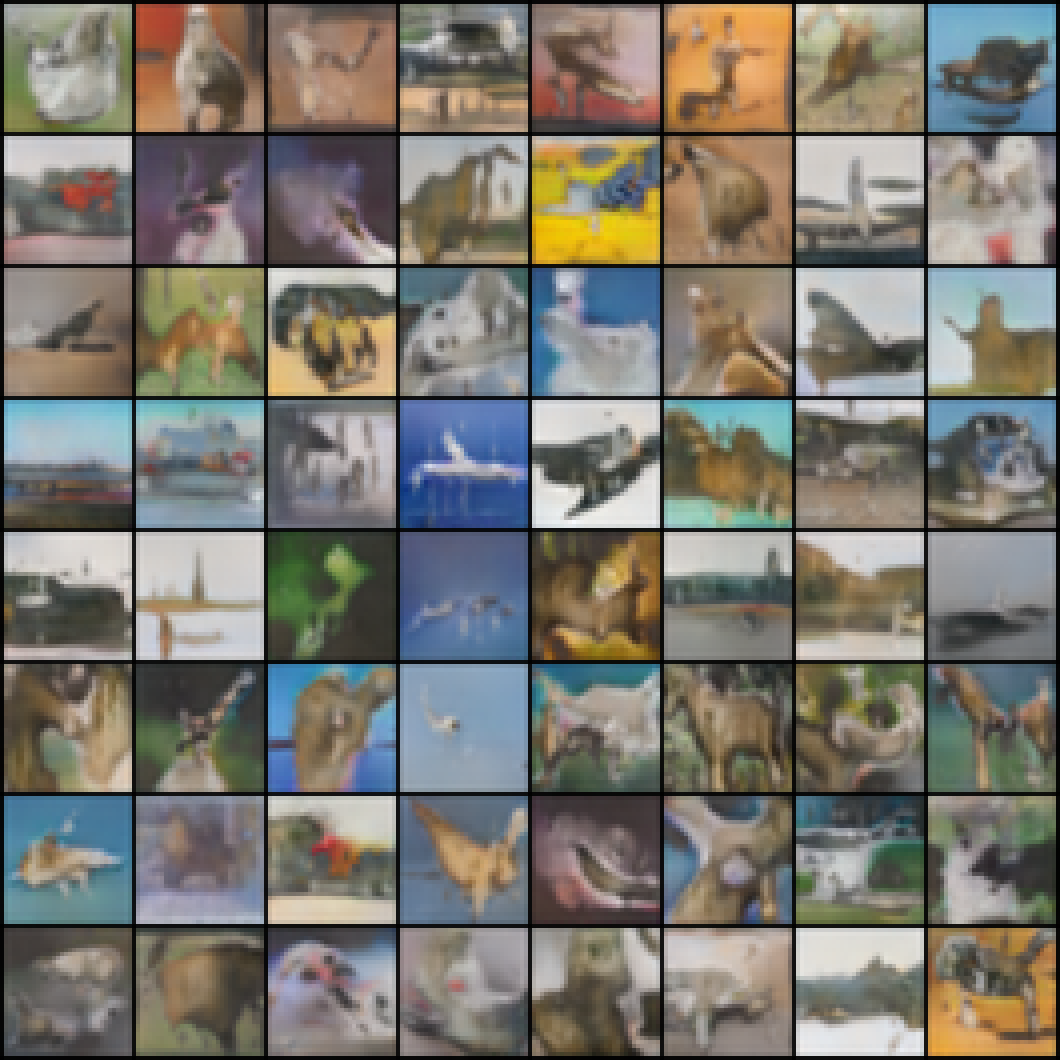}
       \caption{DRS}
    \end{subfigure}
    \begin{subfigure}[b]{0.49\textwidth}
       \centering
       \includegraphics[width=\exfactor\textwidth]{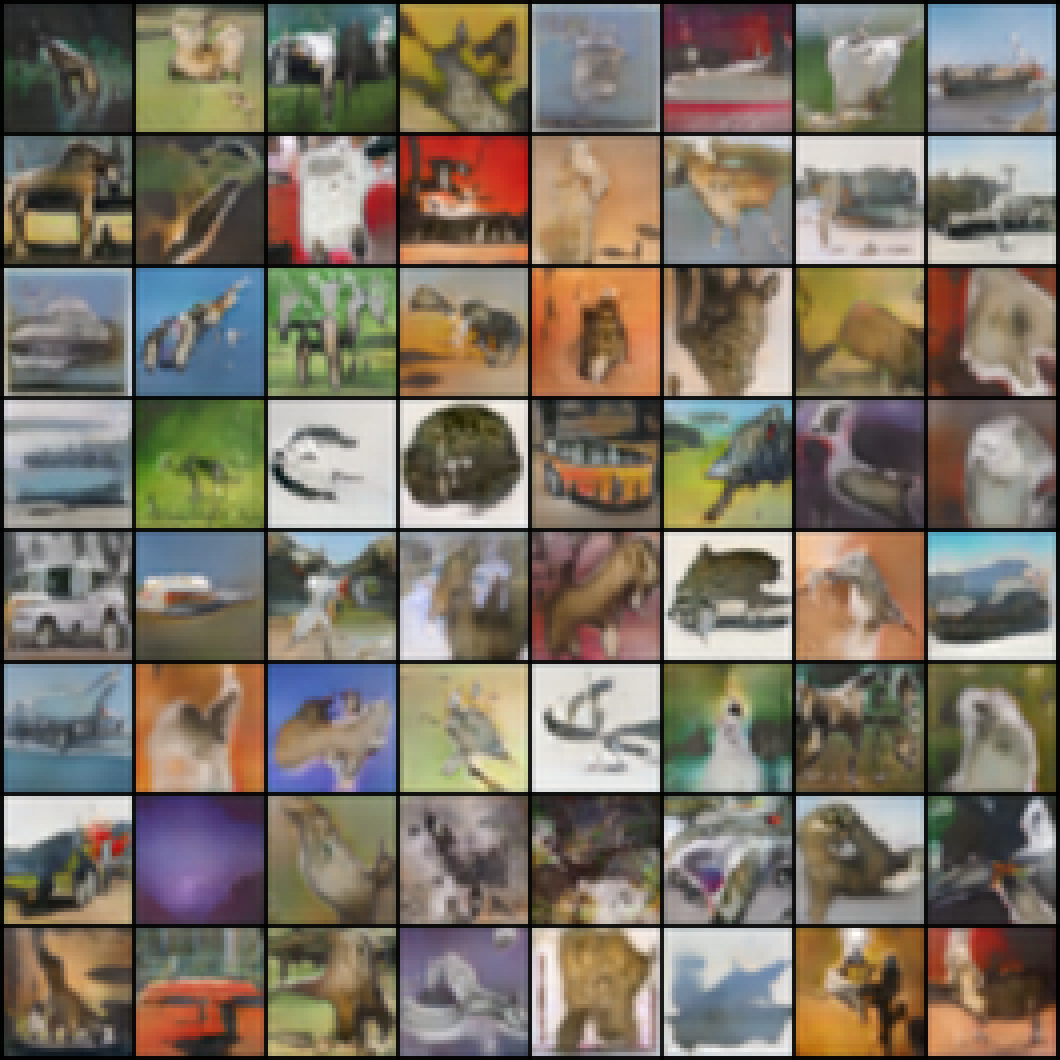}
       \caption{MH-GAN}
    \end{subfigure}
    \begin{subfigure}[b]{0.49\textwidth}
       \centering
       \includegraphics[width=\exfactor\textwidth]{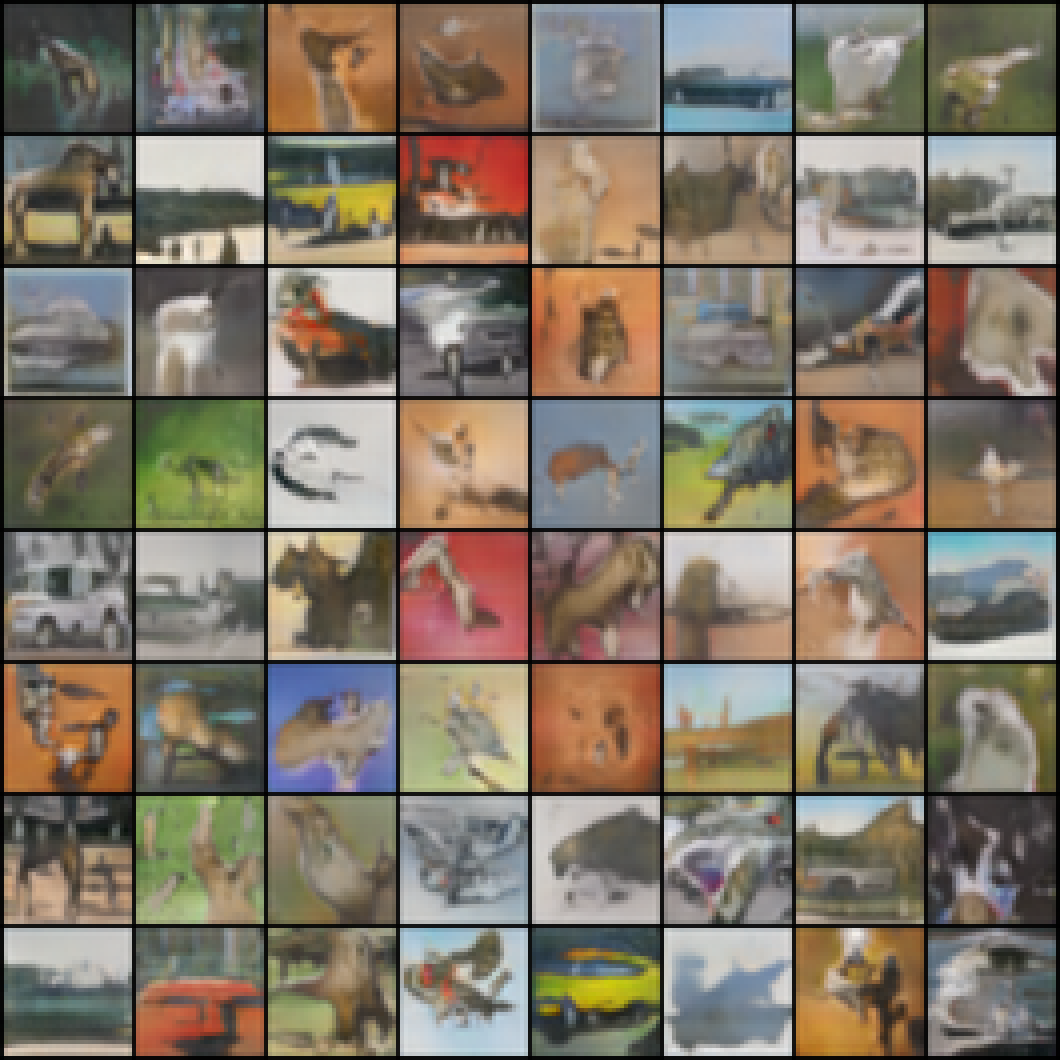}
       \caption{MH-GAN (cal)}
    \end{subfigure}
    \caption{
    Example images on CIFAR-10 for different GAN setups.
    The different selectors (MH-GAN and DRS) are run on the same batch of images.
    Meaning, the same images may appear for both generators.
    The calibrated MH-GAN shows a greater preference for animal-like images with four legs.
    }
    \label{fig:cifar_samples}
\end{figure*}

\begin{figure*}[htbp]
    \centering
    \begin{subfigure}[b]{0.49\textwidth}
       \centering
       \includegraphics[width=\exfactor\textwidth]{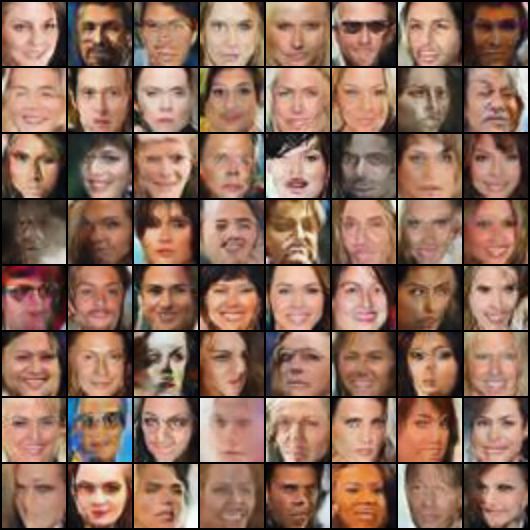}
       \caption{GAN}
    \end{subfigure}
    \begin{subfigure}[b]{0.49\textwidth}
       \centering
       \includegraphics[width=\exfactor\textwidth]{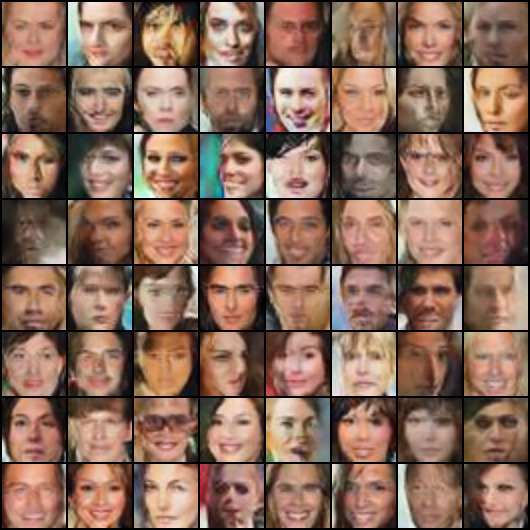}
       \caption{DRS}
    \end{subfigure}
    \begin{subfigure}[b]{0.49\textwidth}
       \centering
       \includegraphics[width=\exfactor\textwidth]{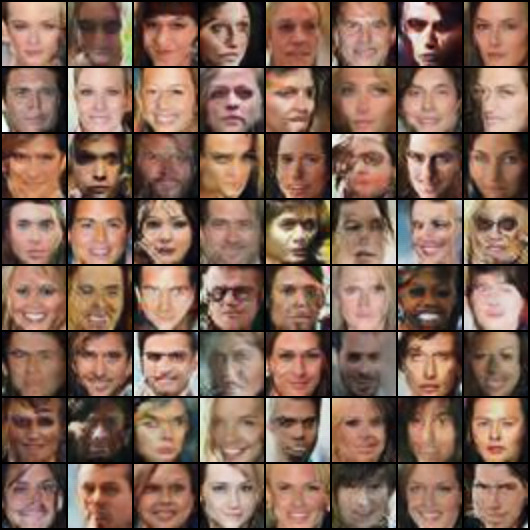}
       \caption{MH-GAN}
    \end{subfigure}
    \begin{subfigure}[b]{0.49\textwidth}
       \centering
       \includegraphics[width=\exfactor\textwidth]{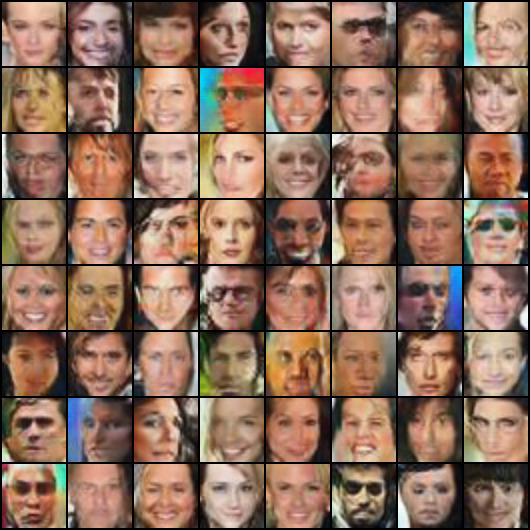}
       \caption{MH-GAN (cal)}
    \end{subfigure}
    \caption{
    Example images on CelebA for different GAN setups.
    Like Figure~\ref{fig:cifar_samples}, the same batch of images goes into each selector.
    }
    \label{fig:celeba_samples}
\end{figure*}

\end{document}